%% file: conference.tex
\documentclass[conference]{IEEEtran}
\IEEEoverridecommandlockouts

\usepackage{cite}
\usepackage{amsmath,amssymb,amsfonts}
\usepackage{algorithmic}
\usepackage{graphicx}
\usepackage{textcomp}
\usepackage[table]{xcolor}
\usepackage{flushend}  
\usepackage{stfloats}  
\usepackage{booktabs}
\usepackage{multirow}
\usepackage{placeins}

\def\BibTeX{{\rm B\kern-.05em{\sc i\kern-.025em b}\kern-.08em
    T\kern-.1667em\lower.7ex\hbox{E}\kern-.125emX}}

\begin{document}

\title{Seeing What Matters, Verifying What Pays:\\
Load-Adaptive Speculative Decoding for Vision-Language Models}

\author{
\IEEEauthorblockN{Hong Nie\textsuperscript{1,\dag}\thanks{\textsuperscript{\dag}This work was achieved during Hong Nie's internship at Unisound AILab under the mentorship of Zhangchi Gao (+1) and Bo Wang (+2).},
Zhangchi Gao\textsuperscript{2},
Yuqian Liang\textsuperscript{1},
Jun Yu\textsuperscript{1,*}\thanks{\textsuperscript{*}Corresponding author: harryjun@ustc.edu.cn},
Bo Wang\textsuperscript{2},
Jianqing Sun\textsuperscript{2}
and Shengping Liu\textsuperscript{2}}
\IEEEauthorblockA{\textsuperscript{1}University of Science and Technology of China
\quad\textsuperscript{2}Unisound AI Technology Co., Ltd.\\
\{gme.niehong, lyq20021130\}@mail.ustc.edu.cn, harryjun@ustc.edu.cn\\
\{gaozhangchi, wangbo, sunjianqing, liushengping\}@unisound.com}
}

\makeatletter
\let\ulsOrigMaketitle\@maketitle
\def\@maketitle{\ulsOrigMaketitle\par\vspace{1mm}%
  \begin{minipage}{\textwidth}
    \centering
    \includegraphics[width=\textwidth]{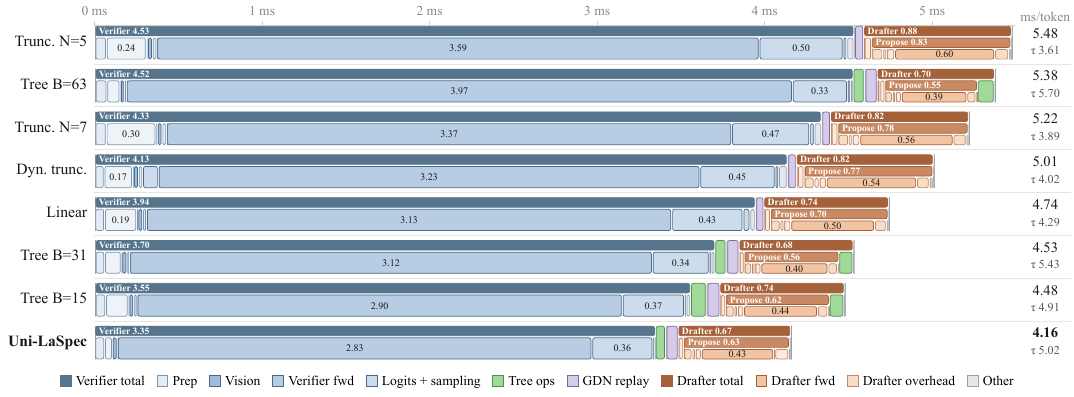}\par\vspace{1mm}%
    \setcounter{figure}{1}
    \parbox{\textwidth}{\footnotesize Fig.~\thefigure.~~Time per output token of one draft--verify round of Qwen3.5-4B
    under each verification strategy with a single concurrent request ($R{=}1$). Each bar shows the round in execution order,
    divided by the strategy's mean acceptance length $\tau$, so a shorter bar
    means higher throughput. A container block (Verifier, Drafter, and the drafter's Propose step)
    spans the segments beneath it. In the verifier, Prep prepares the inputs and attention metadata,
    Vision runs the image encoder, Verifier fwd is the forward pass of the target
    model, and Logits + sampling computes the verifier logits and rejection
    sampling. Tree ops compact the KV cache and recurrent state along the
    accepted path and build the next draft tree (tree rounds only), and GDN
    replay recomputes the state of the verifier's Gated DeltaNet linear-attention
    layers up to the accepted length. In the drafter, Drafter fwd is its single
    parallel forward pass, and Drafter overhead covers input preparation,
    visual-token selection, context KV, length prediction, and candidate
    generation. Other is step bookkeeping outside both models.}
  \end{minipage}\par\vspace{4mm}}
\def\ulsTeaserLabel{\protected@edef\@currentlabel{\thefigure}\label{fig:step-timeline}}
\makeatother

\maketitle
\ulsTeaserLabel

\pagestyle{plain}
\thispagestyle{plain}

\bstctlcite{IEEEtran:BSTcontrol}

\begin{abstract}
Speculative decoding accelerates vision-language model (VLM) inference
without altering the output distribution, yet two factors limit its gains.
Existing drafters either lack the image evidence required for visually
grounded tokens or pay to condition on every visual token, although a short
draft block typically depends on only a few image regions. Moreover, the value
of verifying additional candidates varies with GPU load, so no fixed
verification structure is efficient across serving conditions. We present
\emph{Uni-LaSpec}, a block-parallel speculative decoding framework that
addresses both factors. To improve draft quality, Uni-LaSpec reuses the
verifier's own attention: the query retained from the preceding verification
ranks the cached, int8-quantized visual keys, and a cross-layer vote selects a
few visual tokens that the drafter fuses through gated cross-attention.
Acceptance-aware training further weights each draft position by its
contribution to the accepted prefix. To allocate verification, a marginal-cost
criterion selects the structure per request and per round: draft trees exploit
idle verifier capacity under light load, whereas a dynamic head truncates
linear drafts under heavy load. Across six VLMs from three model families and
17 text and multimodal benchmarks, selecting eight visual tokens retains at
least 98.6\% of the acceptance obtained from the full image at sub-millisecond
selection cost.
With greedy decoding at concurrency one, Uni-LaSpec achieves up to
5.11$\times$ text and 3.24$\times$ multimodal throughput over autoregressive
decoding. Across six load levels and four GPU types, it
outperforms autoregressive decoding in every setting and matches or exceeds
the best fixed verification strategy in 157 of 192 settings; under bursty
arrivals, it reduces mean end-to-end latency by 47.6\% relative to linear
verification.
\end{abstract}


\section{Introduction}
\label{sec:introduction}

Vision-language models (VLMs) now handle document, chart, and multi-image
understanding, yet they still generate one token per forward pass. Each pass
runs the full model over a long, visually dominated context, and its latency
is bound by memory bandwidth rather than by
arithmetic~\cite{patterson2004latency,shazeer2019mqa}.
Speculative decoding removes much of this cost without changing the output:
a lightweight drafter proposes several tokens, and the target model, as the
verifier, checks them in one pass~\cite{xia2022specdec,leviathan2023fast,chen2023accelerating,gante2023assisted}.
Its speedup is determined by two quantities, the mean acceptance length
$\tau$ and the cost of each draft--verify round. Feature-level autoregressive
drafters~\cite{li2024eagle,li2025eagle3} and multi-head
predictors~\cite{cai2024medusa} raise $\tau$, but an autoregressive drafter
spends one drafter step per proposed token, so its drafting cost grows with
the draft length. Block-parallel drafters such as DFlash~\cite{chen2026dflash},
rooted in blockwise and non-autoregressive parallel
decoding~\cite{stern2018blockwise,gu2017nat,qian2020glat,santilli2023parallel},
emit a whole block in one pass and make the drafting cost nearly independent
of the draft length~\cite{xia2024survey}. These advances, however, were
designed and evaluated almost exclusively for text-only language models.

Extending speculative decoding to VLMs exposes a tension between visual
grounding and drafting cost. Text-only drafters never observe the
image and cannot propose tokens that must be read from
it~\cite{gagrani2024speculative}. Vision-aware drafters
receive visual tokens or verifier
features~\cite{lin2025msd,hu2025dream,ganesan2025massv}, but visual tokens
dominate the context of a VLM, and under block drafting, conditioning on one
more visual token costs more than drafting one more
token~\cite{li2026mmsdready}. A second line of work therefore compresses the
visual input of the drafter into a small set of
tokens~\cite{kang2025vispec,huang2025specvlm,wang2025specflash,xie2025hivis,lee2026glance}.
Compression lowers the cost, yet the retained tokens are chosen independently
of the tokens being drafted, so existing drafters still gain little from
visual evidence~\cite{li2026mmsdready}. Yet the evidence a short draft block
needs is local, and it moves as generation proceeds. The first problem is therefore to supply
the drafter, in every round and at negligible cost, with \emph{the visual
evidence on which the current draft actually depends}.

The second problem concerns the cost of each draft--verify round. Under the
lossless constraint~\cite{leviathan2023fast,chen2023accelerating}, one line of
work verifies more candidates per round, expecting a longer accepted prefix to
repay the extra verification; the most common form organizes the candidates
into a tree~\cite{miao2023specinfer,chen2024sequoia,li2024eagle2,ringel2026ddtree}.
These gains are mostly demonstrated at a concurrency of
one~\cite{chen2024sequoia,li2024eagle2,ringel2026ddtree,zhang2026caddtree}.
Another line observes the opposite side of the same trade-off in
block-parallel drafting: the later positions of a block are rarely accepted,
so verifying them wastes computation~\cite{zhang2026flexdraft}. Both respond
to the cost of verification, which depends on GPU load: under light
load, the verifier is memory-bound and additional positions are inexpensive,
so a tree raises $\tau$ at little cost. Under heavy load, it becomes
compute-bound, every additional position slows every request in the batch,
and trimming drafts that will be rejected raises throughput. Because the load
fluctuates from one round to the next, no fixed structure is efficient
throughout. The second problem is therefore \emph{to choose the verification
structure and the number of verified drafts according to the current load}.

Both problems bear on the same objective, the number of accepted tokens per
unit of GPU time. We address them with \emph{Uni-LaSpec}, a block-parallel
speculative decoding framework with two complementary paths. \emph{Path~I}
raises draft quality at almost no drafting cost: the verifier's own
attention, recomputed from its retained query against cached
int8 visual keys, ranks the visual tokens; a cross-layer vote selects the
top-$k$, which the drafter fuses through gated cross-attention; and
acceptance-aware training weights each draft position by its contribution to
the accepted prefix. \emph{Path~II} adapts verification to the load through a
marginal-cost criterion. While the verifier is memory-bound, each request is
verified as a draft tree that exploits the idle capacity; once it becomes
compute-bound, verification switches to linear drafts, and a dynamic head
truncates each draft to the prefix worth verifying. Because every committed token is still checked by the
verifier, neither path changes the output distribution.

The main contributions of this work are as follows:
\begin{itemize}
    \item We equip a block-parallel drafter with an attention-guided visual
    branch that reuses the verifier's attention to select, in every round, the
    few visual tokens on which the current draft depends, with rank-only int8
    scoring and kernel fusion keeping selection below one millisecond.
    \item We propose a load-adaptive verification scheme that derives the
    verification structure and width per request and per round from the
    marginal cost of the verifier, combining draft trees under light load
    with dynamically truncated linear drafts under heavy load in one batch.
    \item We evaluate Uni-LaSpec on six VLMs (Qwen3.5-4B/9B,
    InternVL3.5-4B/8B, and Gemma 4 E2B/E4B) across ten multimodal and seven
    text benchmarks, six load levels, and four GPU types.
    Selecting eight visual tokens retains at least 98.6\% of the acceptance
    obtained from the full image. The load-adaptive policy achieves up to
    5.11$\times$ text and 3.24$\times$ multimodal throughput over
    autoregressive decoding, matches or exceeds the best fixed verification
    strategy in 157 of 192 model--modality--load settings, and reduces the
    mean end-to-end latency under bursty arrivals by 47.6\% relative to linear
    verification.
\end{itemize}

The remainder of this paper is organized as follows.
Section~\ref{sec:related-work} positions the study with respect to prior work.
Section~\ref{sec:method} describes the proposed method.
Section~\ref{sec:experiments} presents the experimental evaluation, and
Section~\ref{sec:conclusion} concludes the paper.

\section{Related Work}
\label{sec:related-work}

\begin{figure*}[!t]
    \centering
    \includegraphics[width=\textwidth]{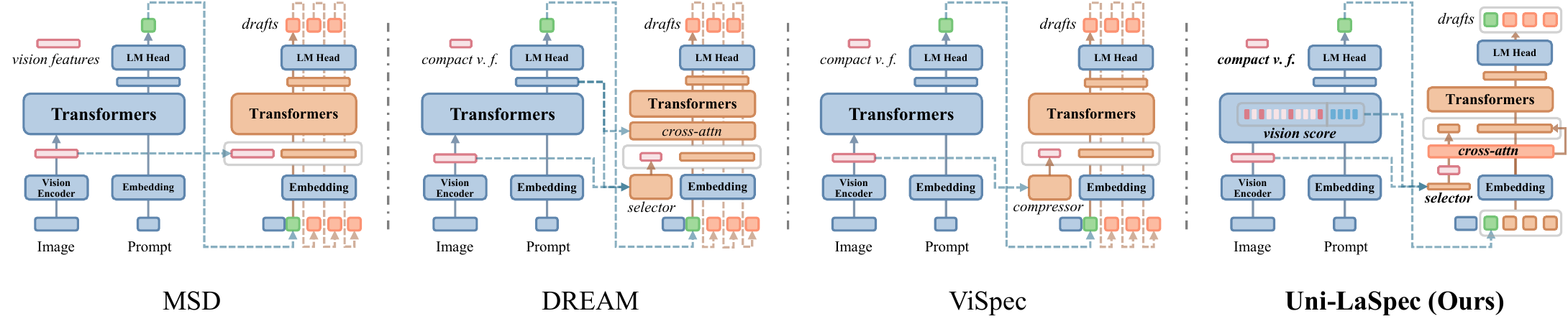}
    \caption{Comparison of visual evidence acquisition mechanisms in MSD~\cite{lin2025msd},
    DREAM~\cite{hu2025dream}, ViSpec~\cite{kang2025vispec}, and Uni-LaSpec.
    In each panel, the verifier is on the left and the drafter is on the right.
    Uni-LaSpec scores the visual tokens with the quantized visual keys cached
    by the verifier, fuses only the selected tokens through gated cross-attention,
    and drafts a masked block in one pass.}
    \label{fig:visual-evidence}
\end{figure*}


\subsection{Multimodal Speculative Decoding}
Multimodal speculative decoding trades visual grounding against drafting cost.
A text-only drafter is cheap but blind to image evidence~\cite{gagrani2024speculative}.
Vision-aware methods reduce this mismatch through a projected verifier vision
encoder and self-distillation (MASSV)~\cite{ganesan2025massv}, staged text and
multimodal training (MSD)~\cite{lin2025msd}, or entropy-selected verifier features
(DREAM)~\cite{hu2025dream}. Lower-cost variants compress the drafter's visual
input (ViSpec and SpecVLM)~\cite{kang2025vispec,huang2025specvlm}, combine
compression with semi-autoregressive prediction
(SpecFLASH)~\cite{wang2025specflash}, or transfer visual semantics from verifier
hidden states without exposing image tokens (HiViS)~\cite{xie2025hivis}. The
central tension remains: additional visual context may improve acceptance yet
erase its benefit through higher draft or prefill cost. MMSpec confirms that
text-oriented methods can degrade on multimodal workloads and that throughput
and latency gains need not coincide~\cite{shen2026mmspec}.

This trade-off is especially pronounced in OCR workloads. Compared with
general visual understanding tasks, OCR typically preserves fine-grained text
at higher image resolutions, causing visual tokens to occupy a larger share of
the context and thus increasing processing cost. Yet the visual evidence needed
for speculative drafting is often local: the current drafting position and the
short block that follows usually depend strongly on only a small region of the
full image.
Although a complete response may draw on different parts of a page, each
draft--verify round often needs only a subset of relevant patches. This
locality provides stronger support for compressing or selecting the visual
information available to the drafter, provided that the retained patches
track the content being generated.

\subsection{Parallel Drafting and Verification}
Standard speculative decoding already evaluates all positions in a drafted
sequence in one verifier pass, but drafter and verifier calls remain
serialized. One line of work extracts more value from each verification:
Block Verification uses a joint block-level acceptance test while preserving
the verifier distribution~\cite{sun2024blockverification}, and Traversal
Verification retains valid subsequences in speculative
trees~\cite{weng2025traversal}. These methods improve candidate utilization
without eliminating the idle time between the two models. Pipeline-parallel methods instead overlap the stages. PEARL overlaps drafting
with pre- and post-verification~\cite{liu2025pearl}; SpecBranch follows several
possible continuations to reduce rollback bubbles~\cite{shen2026specbranch};
and FlexDraft prepares future drafts during current verification and switches
between parallel and sequential execution with batch
size~\cite{zhang2026flexdraft}. ParallelVLM extends this idea to long-video
VLMs by overlapping prefill and decode work while pruning the drafter's visual
input~\cite{kong2026parallelvlm}. Across these methods, additional overlap and
branches shorten the critical path but increase verifier computation and cache
traffic, so their benefit diminishes as the batch becomes compute-bound.

A broader family changes the structure or source of the draft. Candidate
trees trade verification width for acceptance~\cite{miao2023specinfer,chen2024sequoia,li2024eagle2};
dependent draft heads restore token dependencies~\cite{ankner2024hydra,cheng2024redrafter};
self-speculative methods skip verifier layers or exit early~\cite{zhang2023draftverify,yang2023ppd,liu2024kangaroo,kim2023bild};
and retrieval-based methods reuse previously generated
text~\cite{he2023rest,fu2024lookahead}. Hierarchical variants target long-context
or memory-constrained settings~\cite{sun2024triforce,svirschevski2024specexec}.
These designs choose candidates from model confidence or a predefined
structure, rather than from the verification capacity available under the
current load.

\subsection{Speculative Decoding under Concurrent Serving}
Applying speculative decoding to concurrent requests is not a direct extension
of the single-request case. Requests in a continuous batch differ in prompt
length, generation progress, and acceptance pattern, so they advance by
different amounts after every verification round. The system must preserve
their position IDs, attention masks, and KV-cache states while allocating
limited drafting and verification capacity. A more aggressive configuration
may improve average throughput yet worsen time per output token (TPOT), tail
latency, or service-level-objective (SLO) attainment. BASS~\cite{qian2024bass} parallelizes across requests and draft positions,
handling ragged attention either by padding KV tensors or by launching
per-sequence attention kernels; the former wastes computation, while the
latter adds launch overhead. EqSpec/EXSpec~\cite{zhang2025batchspec} further
shows that ragged batches are a correctness problem: requests must realign
their positions, masks, and KV caches after verification. Synchronizing each
request preserves equivalence, whereas grouping equal-length requests reduces
realignment at the risk of head-of-line blocking. Batched speculation is
therefore both an efficiency and a state-management problem.

Verification is also a load-dependent allocation problem. TETRIS assigns
different draft windows under a fixed batch-wide budget, favoring requests
with high cumulative acceptance probability~\cite{wu2025tetris}. MagicDec
shows that the hardware regime changes with workload: short-context,
large-batch verification can be compute-bound, whereas long contexts are often
dominated by KV-cache traffic~\cite{sadhukhan2024magicdec}. AdaSpec adapts
speculation at both request and step granularity under throughput and SLO
constraints~\cite{huang2025adaspec}, while SPECTRE extends adaptive resource
management to multi-model clusters~\cite{xie2026spectre}. Together, these
systems show that a fixed speculative configuration cannot serve all loads;
the verification policy must account for request heterogeneity, acceptance
uncertainty, available capacity, and latency objectives.

\section{Method}
\label{sec:method}

\begin{figure*}[!t]
    \centering
    \includegraphics[width=\textwidth]{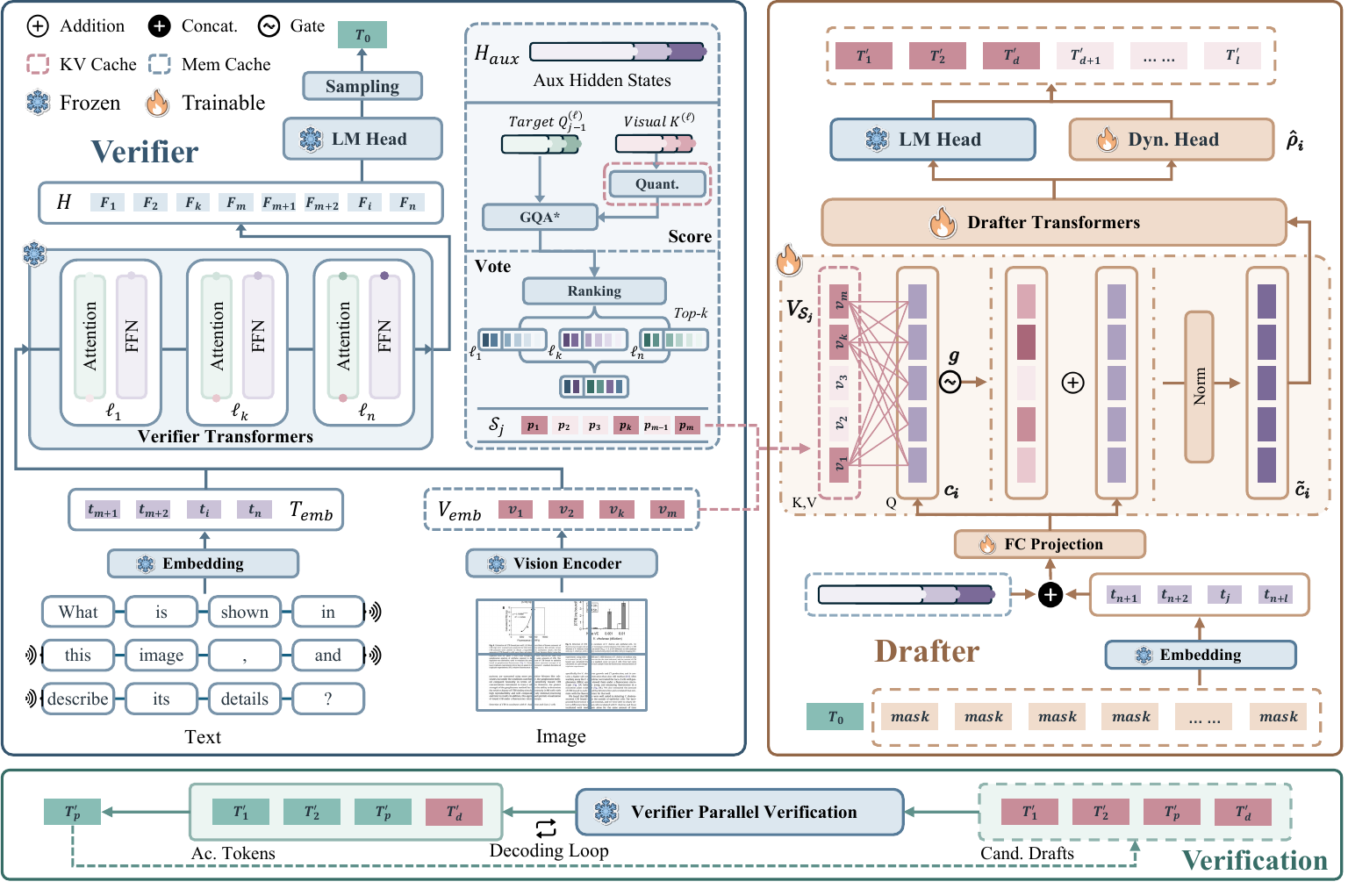}
    \caption{Overview of Uni-LaSpec. Left: the frozen verifier caches
    the auxiliary hidden states $H_{\mathrm{aux}}$ and quantized visual keys. Its
    last retained query, at position $j{-}1$ for the current root $x_j$,
    scores the visual tokens in every selector layer, and a
    cross-layer vote selects the positions $\mathcal{S}_j$. Right: the drafter fuses the
    projected $H_{\mathrm{aux}}$ with the selected visual embeddings through
    gated cross-attention and decodes a masked block in one pass. The LM head
    emits draft tokens, and the dynamic head emits survival probabilities.
    Bottom: the candidates (a dynamically truncated linear draft or a draft tree) are verified in one
    verifier pass, and the accepted tokens start the next round.}
    \label{fig:overview}
\end{figure*}

\subsection{Problem Formulation}
\label{sec:problem}

\textbf{Verifier VLM.}
The frozen VLM $\mathcal{M}_p$ serves as the verifier. Given one or more images
$I$ and a text prompt, its vision encoder
produces $m$ visual embeddings $V=[v_1,\dots,v_m]$ that are interleaved with
the text embeddings to form the multimodal context. Let $x_{1:j}$ be the
committed sequence and $h^{(\ell)}_i$ the hidden state at position $i$ after
layer $\ell$. The verifier defines
\begin{equation}
    p(\cdot\mid x_{1:i})=\mathrm{softmax}\big(W_{\mathrm{lm}}h^{(N)}_i\big),
    \label{eq:verifier}
\end{equation}
and emits one token per forward pass. For visually rich inputs, visual tokens
dominate the context and the prefill is expensive; once it is complete,
however, $V$, the KV cache, and the intermediate hidden states of the verifier
are already in memory and can be reused during decoding at no extra verifier
cost. This asymmetry motivates the whole design: the expensive part of the
model has already computed, for free, most of what a drafter needs.

\textbf{Draft--verify rounds.}
Decoding proceeds in rounds~\cite{leviathan2023fast,chen2023accelerating}. A
round starts from the root token $T_0\triangleq x_j$, the bonus token emitted
by the previous verification. This token has been emitted but has not yet
been processed by the verifier; its cached history ends at $x_{j-1}$.
A drafter $\mathcal{M}_q$ proposes candidate
continuations, the verifier scores them in a single pass, and under greedy
decoding the accepted length is the longest prefix the verifier would itself
have produced,
\begin{equation}
    a=\max\Big\{k:
    T'_i=\mathop{\arg\max}\,p^{\star}_i,\ \forall\, i\le k\Big\},
    \label{eq:accept}
\end{equation}
where $p^{\star}_i=p(\cdot\mid x_{1:j},T'_{1:i-1})$. The round commits
$T'_{1:a}$ and the bonus token $\arg\max p^{\star}_{a+1}$. The stochastic rule
accepts $T'_i$ with probability $\min\{1,p^{\star}_i(T'_i)/q_i(T'_i)\}$ and
resamples from the normalized residual. Either rule leaves the output
distribution identical to that of $\mathcal{M}_p$. This is the property the
method builds on: \emph{the drafter, the evidence it reads, and the number of
candidates verified change only speed, never the text}. We may therefore treat
them as free design variables and optimize them directly.

\textbf{What determines speed.}
With $C_{\mathrm{draft}}$ the cost of producing a draft and
$C_{\mathrm{ver}}(W)$ the cost of one verifier pass over $W$ total
positions, including roots, the per-token latency for a linear draft of
$\gamma$ candidates is
\begin{equation}
    \bar{C}=\frac{\mathbb{E}[C_{\mathrm{draft}}]
    +\mathbb{E}[C_{\mathrm{ver}}(1+\gamma)]}{\tau},\quad
    \tau=\mathbb{E}[a]+1 .
    \label{eq:latency}
\end{equation}
Three forces act on this ratio. Sequential drafters pay
$C_{\mathrm{draft}}$ once per proposed token, which caps how far ahead they can
look. In a VLM, conditioning the drafter on the image is more expensive than
proposing one more token~\cite{li2026mmsdready}, so visual grounding, not draft
length, is the scarce resource. Finally, $C_{\mathrm{ver}}$ is nearly flat
while the verifier is memory-bound and grows once it becomes
compute-bound~\cite{sadhukhan2024magicdec}, so the value of verifying one more
candidate depends on how busy the GPU is.

\textbf{Block drafting.}
Following DFlash~\cite{chen2026dflash}, $\mathcal{M}_q$ is a lightweight
non-causal drafter that shares the token embedding with the verifier and
predicts over a compressed vocabulary (Section~\ref{sec:setup}). It
reads context features fused from a set $\mathcal{A}$ of verifier layers,
\begin{equation}
    c_i=W_c\big[h^{(\ell_1)}_i\,\Vert\,\cdots\,\Vert\,
    h^{(\ell_{|\mathcal{A}|})}_i\big],
    \label{eq:context-feature}
\end{equation}
(Fig.~\ref{fig:overview}, $H_{\mathrm{aux}}$), visual evidence
$V_{\mathcal{S}}$ indexed by a small set $\mathcal{S}$ with
$|\mathcal{S}|\le k\ll m$, and a block that consists of the root followed by
$\Gamma$ mask tokens. One pass fills every mask slot at once and returns
per-position marginals
\begin{equation}
    q_i(\cdot)=q_\theta\big(\cdot\mid c_{<j},V_{\mathcal{S}},T_0\big),
    \quad i=1,\dots,\Gamma .
    \label{eq:draft-marginal}
\end{equation}
Two consequences matter for what follows. First, $C_{\mathrm{draft}}$ is
essentially independent of how many tokens are proposed, so the draft horizon
is no longer the bottleneck. Second, because each $q_i$ is conditioned on the
root rather than on the preceding draft tokens, one pass yields a
\emph{distribution per position} instead of a single linear draft. The extra
information is free, and both halves of our verification policy consume it.

\textbf{Optimization Objective.}
Let $Y$ be the verifier's greedy continuation after $x_{1:j}$ and let the
verified candidates of a request form a prefix-closed set $\mathcal{V}$ with
one of two structures: a linear draft or a draft tree. Dynamic truncation
applies to the linear draft. Writing
$\rho(u)=\Pr(u\text{ is a prefix of }Y)$ for the
survival probability of a candidate, the expected accepted length is additive:
\begin{equation}
    \begin{aligned}
    \mathbb{E}[a\mid\mathcal{V}]
    &=\sum_{u\in\mathcal{V}}\rho(u)\\
    &=\begin{cases}
    \displaystyle\sum_{i=1}^{\gamma}\rho_i,
        &\text{linear},\\[2pt]
    \displaystyle\sum_{d=1}^{D}\sum_{u\in\mathcal{V}_d}\rho(u),
        &\text{tree}.
    \end{cases}
    \end{aligned}
    \label{eq:survival}
\end{equation}
Here $\rho_i$ is the survival probability of the linear draft's prefix of
length $i$, so $\rho_1\ge\cdots\ge\rho_{\gamma}$; $\mathcal{V}_d$ is the set
of draft-tree nodes at depth $d$, and $D$ is the tree's maximum depth.
For $R$ concurrent requests occupying $W=\sum_r(1+|\mathcal{V}^{(r)}|)$
verifier positions, the throughput is
\begin{equation}
    \Theta=\frac{\sum_{r=1}^{R}\big(1+\sum_{u\in\mathcal{V}^{(r)}}
    \rho^{(r)}(u;\theta,\mathcal{S})\big)}
    {C_{\mathrm{draft}}(k)+C_{\mathrm{ver}}(W)} .
    \label{eq:objective}
\end{equation}
Eq.~\eqref{eq:objective} states the design problem compactly. The numerator
is a sum of survival probabilities, so it can be raised either by making each
candidate more likely to survive, which is a property of the drafter and of
the evidence it reads, or by adding candidates. The denominator says how much
an added candidate costs. The two are independent knobs, and we treat them as
two paths.

\subsection{Method Overview}
\label{sec:overview}
Uni-LaSpec has two complementary paths:
\begin{equation}
\begin{aligned}
    \text{Path I:}\quad &\uparrow\rho(u;\theta,\mathcal{S}),\\
    \text{Path II:}\quad &\mathcal{V}^{(r)}_t\in\big\{
    \mathrm{Tree}(B_t),\mathrm{Linear}(\gamma^{(r)}_t)\big\}.
\end{aligned}
    \label{eq:two-paths}
\end{equation}
Path~I (Section~\ref{sec:path1}) raises every $\rho$ at a fixed verification
width, by giving the drafter the visual evidence it is missing and by training
it on the quantity that actually matters, the accepted prefix. Path~II
(Section~\ref{sec:path2}) spends the verifier budget where it is worth most.
Under low load, eligible requests verify a draft tree of alternatives. Under high load, it
uses a linear draft and dynamically truncates each request at the depth where
survival no longer pays for its verification cost, giving
$\gamma^{(r)}_t$. A request-level scheduler combines these structures as
requests arrive and depart (Section~\ref{sec:hybrid-scheduling}).
Fig.~\ref{fig:overview} shows one round of the resulting system.

\subsection{Path I: Improving Draft Quality}
\label{sec:path1}

\subsubsection{Attention-Guided Visual Evidence}
\label{sec:visual}

\begin{figure}[t]
    \centering
    \includegraphics[width=\columnwidth]{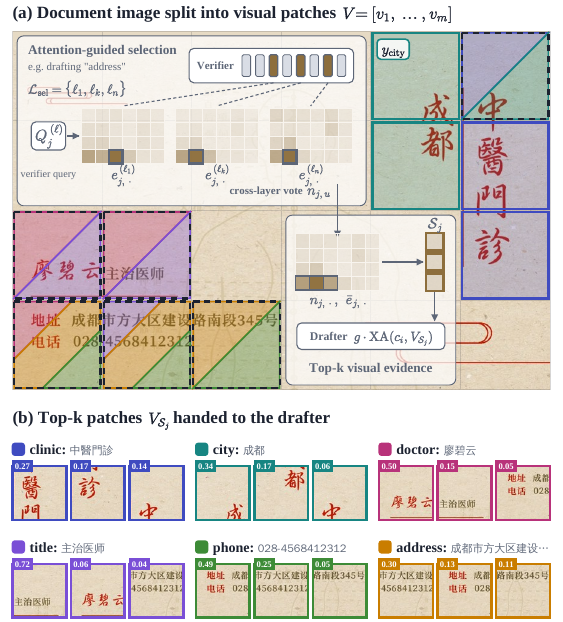}
    \caption{Attention-guided visual evidence selection for schema extraction.
    (a)~Colored regions link output fields to supporting image patches;
    diagonal splits indicate patches shared by multiple fields. The insets
    show verifier attention scoring, cross-layer voting, and fusion of the
    selected visual embeddings into the drafter.
    (b)~Top-$k$ patches selected for each field, with $k{=}3$ shown for
    illustration. Selection
    is refreshed in each drafting round to follow the field being generated.}
    \label{fig:visual-patches}
\end{figure}
\begin{figure*}[t]
    \centering
    \includegraphics[width=\textwidth]{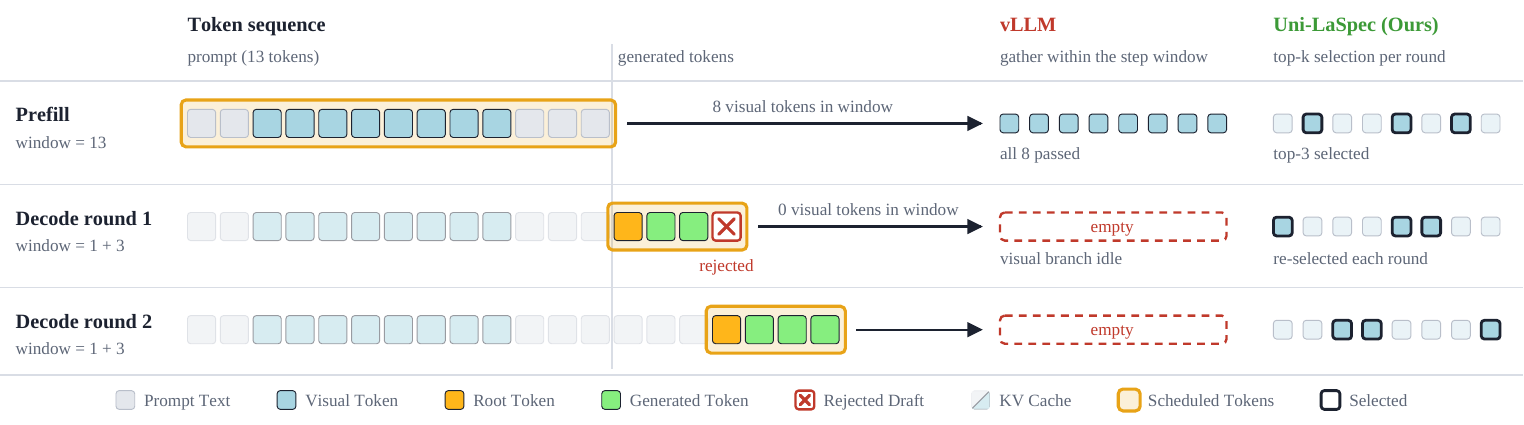}
    \caption{Visual input availability across prefill and decoding.
    In vLLM, step-local gathering restricts visual inputs to the current scheduled
    window. Image embeddings reach the drafter during prefill, whereas
    subsequent decoding windows contain no visual tokens, leaving the visual
    branch inactive. Uni-LaSpec caches image embeddings for each request and
    refreshes the top-$k$ selection using the verifier's last retained query
    in every drafting round, maintaining access to visual evidence throughout
    generation.}
    \label{fig:visual-cache}
\end{figure*}

A text-only drafter often produces low-quality drafts on multimodal inputs,
especially when the next tokens must be read off the image. Passing all $m$
visual tokens $V=[v_1,\dots,v_m]$ to the drafter would restore what it sees
but not what it costs: in a block drafter, conditioning on $V$ dominates
$C_{\mathrm{draft}}$~\cite{li2026mmsdready}. Existing designs therefore narrow
the visual input with a compressor or with a second encoder trained for the
purpose (Fig.~\ref{fig:visual-evidence}), which adds parameters and gives the
drafter another source of drift with respect to the verifier.
DREAM~\cite{hu2025dream} instead derives its visual features from the
verifier's intermediate hidden states, so its visual branch cannot start until
the verifier pass has produced them. This severely delays the visual pipeline:
in every round, visual processing is pushed behind verification instead of
overlapping with it, which hurts the parallelism of drafting.

Our goal is to give the drafter, at as little cost as possible, a set
$\mathcal{S}_j$ of $k\ll m$ patches on which the block rooted at $T_0=x_j$
actually depends. Both requirements point to the same source. The verifier has
already decided where to look: when it produced $x_j$, its query
$Q^{(\ell)}_{j-1}$ at each layer $\ell$ attended over the visual keys
$K^{(\ell)}_{1:m}$, and this attention indicates the regions from which the
next tokens are being read. The keys $K^{(\ell)}_{1:m}$ were computed during
prefill and $Q^{(\ell)}_{j-1}$ during the last verification pass, so reading this
signal needs neither a new network nor a wait for another forward pass. In
principle, the scores thus come for free. In practice, mature serving engines
compute attention with tiled, fused
kernels~\cite{dao2022flashattention,dao2023flashattention2,kwon2023pagedattention}
that never materialize or return the attention scores, so the scores must be
recomputed; the rest of our design makes this recomputation as cheap as
possible.

The selector reads the last valid verifier input row retained after
acceptance, excluding rejected suffixes or branches. That row predicts the
new root $x_j$; it is not a query computed from $x_j$ itself.
To find the relevant patches, $Q^{(\ell)}_{j-1}$ scores the cached keys at a small
set $\mathcal{L}_{\mathrm{sel}}$ of full-attention layers,
\begin{equation}
    e^{(\ell)}_{j,u}\propto\big\langle Q^{(\ell)}_{j-1},\,K^{(\ell)}_{u}\big\rangle,
    \quad u=1,\dots,m,\ \ell\in\mathcal{L}_{\mathrm{sel}}.
    \label{eq:visual-score}
\end{equation}
A single layer is a noisy judge, and different layers often favor different
patches (Fig.~\ref{fig:visual-patches}a, insets). Each layer $\ell$ therefore
nominates its $k_c$ highest-scoring positions
$\mathcal{C}^{(\ell)}_j=\mathrm{TopK}_{k_c}\big(e^{(\ell)}_{j,\cdot}\big)$, and
a cross-layer vote keeps the positions on which several layers agree,
\begin{equation}
    n_{j,u}=\sum_{\ell\in\mathcal{L}_{\mathrm{sel}}}
    \mathbb{1}\big[u\in\mathcal{C}^{(\ell)}_j\big],\quad
    \mathcal{S}_j=\mathrm{TopK}_{k}\big(n_{j,\cdot},\,\bar{e}_{j,\cdot}\big),
    \label{eq:visual-vote}
\end{equation}
where $\bar{e}_{j,u}$ is the mean of $e^{(\ell)}_{j,u}$ over
$\mathcal{L}_{\mathrm{sel}}$ and breaks ties in $n_{j,u}$. Because
$Q^{(\ell)}_{j-1}$ produced the root of the current block, $\mathcal{S}_j$ is
specific to that block, and recomputing it in every round lets the evidence
follow the text across the page: as the output moves from one field to the
next, so does $\mathcal{S}_j$ (Fig.~\ref{fig:visual-patches}b).

Several choices keep this recomputation cheap. Only the ranking of
$e^{(\ell)}_{j,\cdot}$ is used, never its values, so the softmax is skipped.
The visual keys are cached once per request in 8-bit form, with one scale
$s^{(\ell)}_{u,h}$ per position $u$ and key head $h$~\cite{liu2024kivi},
\begin{equation}
    \hat{K}^{(\ell)}_{u,h}=\mathrm{round}\big(K^{(\ell)}_{u,h}/s^{(\ell)}_{u,h}\big),
    \quad
    s^{(\ell)}_{u,h}=\big\|K^{(\ell)}_{u,h}\big\|_{\infty}/127,
    \label{eq:visual-quant}
\end{equation}
which halves both the memory held for $K^{(\ell)}_{1:m}$ and the memory
traffic of every scoring pass relative to 16-bit keys. The scoring also
follows the verifier's grouped-query attention~\cite{ainslie2023gqa}: the $G$
query heads that share key head $h$ are averaged before the dot product, and
the scale, which factors out of the dot product, is applied to the result,
\begin{equation}
\begin{aligned}
    e^{(\ell)}_{j,u}&=\frac{1}{H}\sum_{h=1}^{H}
    s^{(\ell)}_{u,h}\,\big\langle\bar{q}^{(\ell)}_{j-1,h},\,\hat{K}^{(\ell)}_{u,h}\big\rangle,\\
    \bar{q}^{(\ell)}_{j-1,h}&=\frac{1}{G}\sum_{g=1}^{G}Q^{(\ell)}_{j-1,h,g},
\end{aligned}
    \label{eq:visual-grouped}
\end{equation}
where $H$ is the number of key heads. Since the dot product is linear, this
equals the mean of the per-head scores of Eq.~\eqref{eq:visual-score} up to
quantization, but it takes $H$ dot products per position instead of $GH$ and
never replicates the shared keys. Only $|\mathcal{L}_{\mathrm{sel}}|$
full-attention layers take part. The encoder output $V$ is kept in a
request-level cache, because the default serving path exposes the image
embeddings only in the step that schedules them and would leave the visual
branch empty in every decode round (Fig.~\ref{fig:visual-cache});
$\mathcal{S}_j$ indexes this cache directly, so nothing is re-encoded.
Finally, the drafter attends to $V_{\mathcal{S}_j}$, i.e., $k$ tokens instead
of $m$, through a gated cross-attention added to each context feature $c_i$
of Eq.~\eqref{eq:context-feature},
\begin{equation}
    \tilde{c}_i=\mathrm{RMSNorm}\big(c_i+g\cdot\mathrm{XA}(c_i,V_{\mathcal{S}_j})\big),
    \label{eq:visual-fusion}
\end{equation}
whose gate $g$ is initialized to $0$, so the visual increment initially
vanishes and the context follows the baseline $\mathrm{RMSNorm}(c_i)$ path;
training decides how far to open the gate. The
drafter's visual cost therefore scales with $k$ rather than with $m$, while
the verifier still reads all of $V$ and checks every committed token.

This design suits OCR-style tasks, such as document parsing and schema
extraction, particularly well (Fig.~\ref{fig:visual-patches}). There, most
drafted tokens are copied from a small region of the page, so the verifier's
attention for a field concentrates on one or two patches and a small $k$
already covers the evidence the block needs. The tokens that are hardest for a
text-only drafter, such as names, addresses, and digit strings, are also those
that no language prior can predict, and they are exactly the tokens whose
patches $\mathcal{S}_j$ places in front of the drafter. OCR outputs further
follow the reading order of the page, so consecutive blocks read neighboring
patches and $\mathcal{S}_j$ moves smoothly as generation proceeds; in
Fig.~\ref{fig:visual-patches}, the address is read left to right across
adjacent patches, and because the selection is redone for every block, this
movement costs nothing extra. Finally, document images are high-resolution,
so $m$ is large, which is where keeping the drafter's visual cost proportional
to $k$ rather than to $m$ pays off most.

\subsubsection{Acceptance-Aware Training}
\label{sec:training}

A block drafter trained with equal weight on every position optimizes the wrong
quantity. What a round is paid for is the accepted \emph{prefix}: by
Eq.~\eqref{eq:survival} an error at position $i$ does not cost one token, it
destroys every token after it. Positions are therefore not exchangeable, and
the loss should say so.

We therefore weight each position by how much of the block it gates. The
verifier, the embedding, and the LM head stay frozen, and the drafter is
supervised by the verifier distribution on blocks anchored at sampled
positions as in distillation-based drafter
alignment~\cite{zhou2023distillspec},
\begin{equation}
    \mathcal{L}_{\mathrm{tok}}=\sum_{a}\sum_{i=1}^{\Gamma}
    \mathrm{sg}(w_{a,i})\,
    \mathrm{CE}\big(p(\cdot\mid x_{1:a+i-1}),\,q_{a,i}\big),
    \label{eq:token-loss}
\end{equation}
with weights taken from the surrogate of Eq.~\eqref{eq:survival},
\begin{equation}
\begin{aligned}
    \bar{q}_{a,i}&=(1-\alpha)\,e^{-\mathrm{CE}_{a,i}}+\alpha,\\
    w_{a,i}&=\sum_{v=i}^{\Gamma}\prod_{t=1}^{v}\bar{q}_{a,t}
    =\frac{\partial}{\partial\log\bar{q}_{a,i}}
    \sum_{v=1}^{\Gamma}\prod_{t=1}^{v}\bar{q}_{a,t}.
\end{aligned}
    \label{eq:position-weight}
\end{equation}
The weight at position $i$ is the expected number of accepted tokens from
$i$ onward under the surrogate, and equals the sensitivity of the expected
accepted length to that position's accuracy.

Gradients are then spent where they buy tokens: early positions, and
positions that the model is close to getting right, dominate; deep positions
that are already unreachable are not chased. The objective the drafter is
trained on is the numerator of Eq.~\eqref{eq:objective} rather than a proxy for
it.

\begin{figure*}[t]
    \centering
    \includegraphics[width=\textwidth]{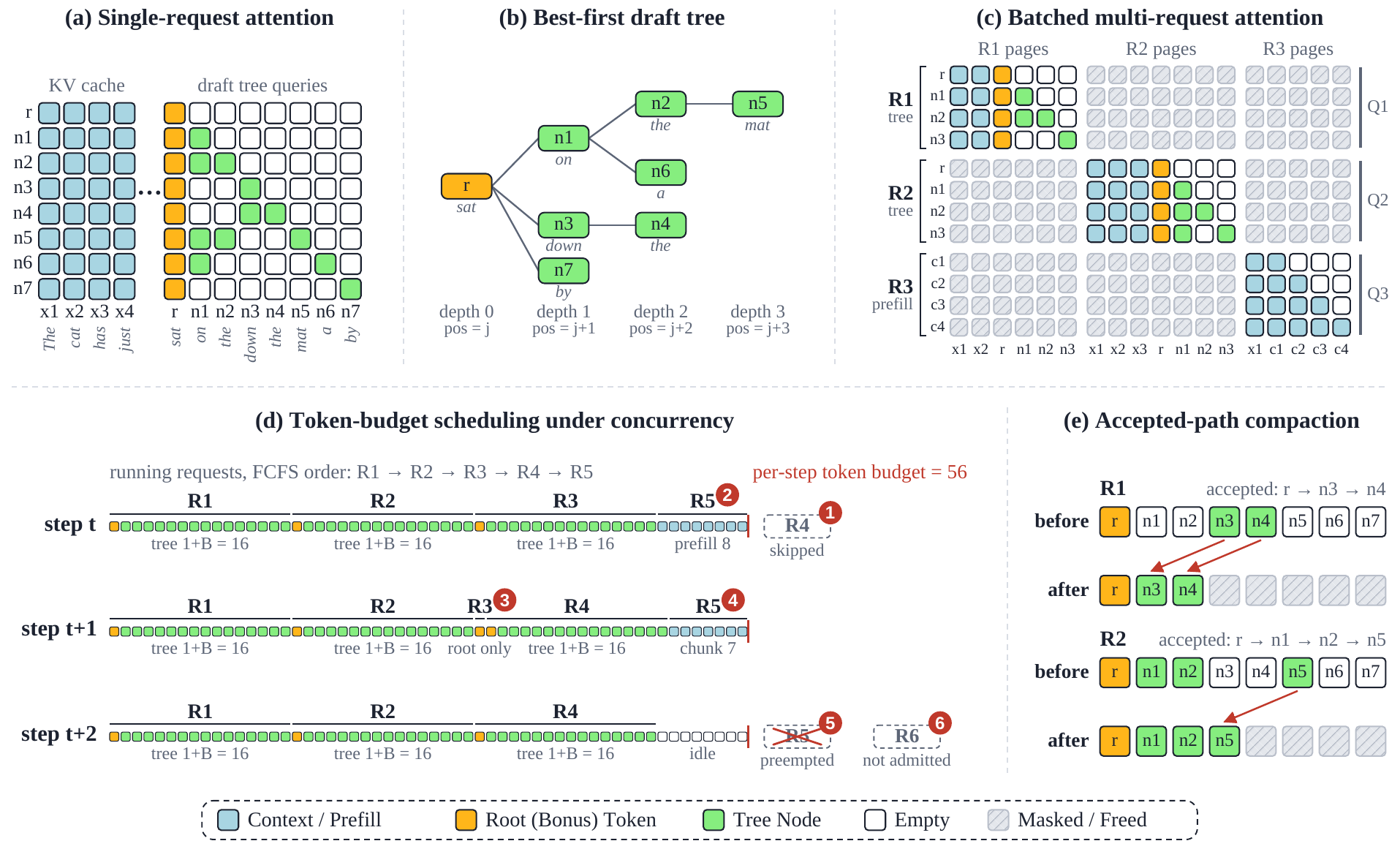}
    \caption{Parallel verification and scheduling of draft trees.
    (a)~Attention mask for a single request. The root is the previous round's
    bonus token; each node attends to the cached prefix $x_{1:j-1}$, its
    ancestors, and itself.
    (b)~Best-first tree construction with nodes indexed in topological order.
    Rotary positions follow tree depth, $\mathrm{pos}(u)=j+|u|$, independently
    of the packed KV-cache indices.
    (c)~Batched execution of two draft trees and a prefill chunk. The logical
    attention mask is block diagonal across requests, preserving each tree's
    ancestor dependencies and the prefill chunk's causal order.
    (d)~Scheduling under a shared token budget. The numbered events illustrate
    deferral of a tree that exceeds the remaining budget~(1), allocation of
    residual capacity to prefill~(2), verification of the root alone near the
    context limit~(3), discarding drafts during continued prefill~(4),
    preemption under KV-cache pressure~(5), and suspension of new
    admissions~(6).
    (e)~Compaction of the accepted path. Retained entries are gathered into
    contiguous slots within each request's storage region, and entries from
    rejected branches are discarded.}
    \label{fig:draft-tree}
\end{figure*}

\subsection{Path II: Load-Adaptive Verification}
\label{sec:path2}

\subsubsection{The Marginal Value of a Candidate}
\label{sec:allocation}

Path~I improves draft quality; Path~II allocates verifier capacity to the
resulting candidates. A fixed verification width ignores the load-dependent
cost of each additional verifier position. When verification is memory-bound,
weight transfers dominate and more positions can be processed with little
incremental latency. At higher occupancy, verification becomes compute-bound,
so expanding the batch increases the latency of every concurrent request.
We therefore set verification width by comparing each candidate's expected
acceptance gain with its incremental verification cost.

Comparing Eq.~\eqref{eq:objective} before and after adding a candidate whose
parent prefix is already retained, with the drafting cost held fixed, gives
the admission criterion. A candidate with survival probability $\pi$ improves
throughput if and only if
\begin{equation}
    \pi>\delta(W)\triangleq
    \frac{R\,\bar{\tau}\,\big[C_{\mathrm{ver}}(W{+}1)-C_{\mathrm{ver}}(W)\big]}
    {C_{\mathrm{draft}}+C_{\mathrm{ver}}(W)},
    \label{eq:marginal-threshold}
\end{equation}
where $\bar{\tau}$ is the mean expected number of committed tokens per request.
The threshold $\delta(W)$ is the minimum survival probability required at
batch width $W$. In the memory-bound region, the marginal verification cost is
small and $\delta(W)$ approaches zero. Beyond the knee of the cost curve, the
threshold rises and candidate expansion becomes increasingly selective. We
estimate the knee offline as the largest width whose verification cost remains
within a factor of $1+\epsilon$ of the cost at the reference width of 16,
\begin{equation}
    W^{\star}=\max\big\{W:
    C_{\mathrm{ver}}(W)\le(1+\epsilon)\,C_{\mathrm{ver}}(16)\big\},
    \label{eq:knee}
\end{equation}
and use it to determine when branching is affordable. Both regimes use the
same drafter: a draft tree covers alternative continuations within the
low-cost region, while dynamic truncation limits each linear draft to
candidates whose expected gain justifies their verification cost. We first
describe these two structures, then explain how the scheduler combines them
under changing load in Section~\ref{sec:hybrid-scheduling}.

\subsubsection{Low Load under Concurrent Serving: Widening with Draft Trees}
\label{sec:draft-tree}

When $\delta\approx0$, verifying a single linear draft is a missed opportunity. The
block drafter has produced a full distribution at every position
(Eq.~\eqref{eq:draft-marginal}), and a linear draft uses only the top-1 entry of each.
The positions where the linear draft is most likely to break are precisely the ones
whose distributions are flat, i.e., where a second candidate would often be the
right one.

Because the marginals are independent given the root, a candidate continuation
is just a choice of rank at each depth, and its survival can be read off the
marginals directly: with $y_{i,r}$ the rank-$r$ token at depth $i$, the node
$u=(r_1,\dots,r_d)$ has plug-in value
\begin{equation}
    \log\hat{\pi}(u)=\sum_{i=1}^{d}\log q_i\big(y_{i,r_i}\big).
    \label{eq:draft-tree-value}
\end{equation}
Value decreases along every path, so the $B$ most valuable prefixes are
automatically prefix-closed: they form a tree, and that tree maximizes the
surrogate expected acceptance length under a budget
of $B$ nodes~\cite{ringel2026ddtree},
\begin{equation}
    \mathcal{T}^{\star}=\mathop{\arg\max}_{|\mathcal{T}|\le B}
    \sum_{u\in\mathcal{T}}\hat{\pi}(u).
    \label{eq:draft-tree}
\end{equation}
Growing it is a best-first expansion: repeatedly take the most valuable
unexplored prefix and offer its sibling and its first child as new candidates
(Fig.~\ref{fig:draft-tree}b). The whole tree is then verified in one verifier pass by
telling attention what the tree means: a node sees its ancestors and itself,
and its position is its depth,
\begin{equation}
    M_{uv}=\mathbb{1}\big[v\preceq u\big],\quad
    \mathrm{pos}(u)=j+|u|,
    \label{eq:tree-mask}
\end{equation}
so every node is scored under its exact prefix rather than under whatever
happens to precede it in the packed sequence (Fig.~\ref{fig:draft-tree}a).
Acceptance is then a walk from the root,
\begin{equation}
    u_{t+1}=\mathrm{child}\big(u_t,\;\mathop{\arg\max}\,
    p(\cdot\mid x_{1:j},u_t)\big),
    \label{eq:tree-walk}
\end{equation}
which stops at the first verifier prediction that no child offers; the path is
committed with one bonus token and the rest of the tree is discarded
(Fig.~\ref{fig:draft-tree}e). Under concurrency the trees of different requests are
verified together but never interact, because a request may only attend to its
own history (Fig.~\ref{fig:draft-tree}c--d).

Branching therefore buys acceptance exactly where the linear draft is weak, and it is paid for
with verifier width that was idle. Since each node is scored under its true
prefix, the walk accepts only tokens the verifier itself would have produced, so
widening changes speed and not the output.

\subsubsection{High Load under Concurrent Serving: Dynamic Truncation of Linear Drafts}
\label{sec:truncation}

\begin{figure}[t]
    \centering
    \includegraphics[width=\columnwidth]{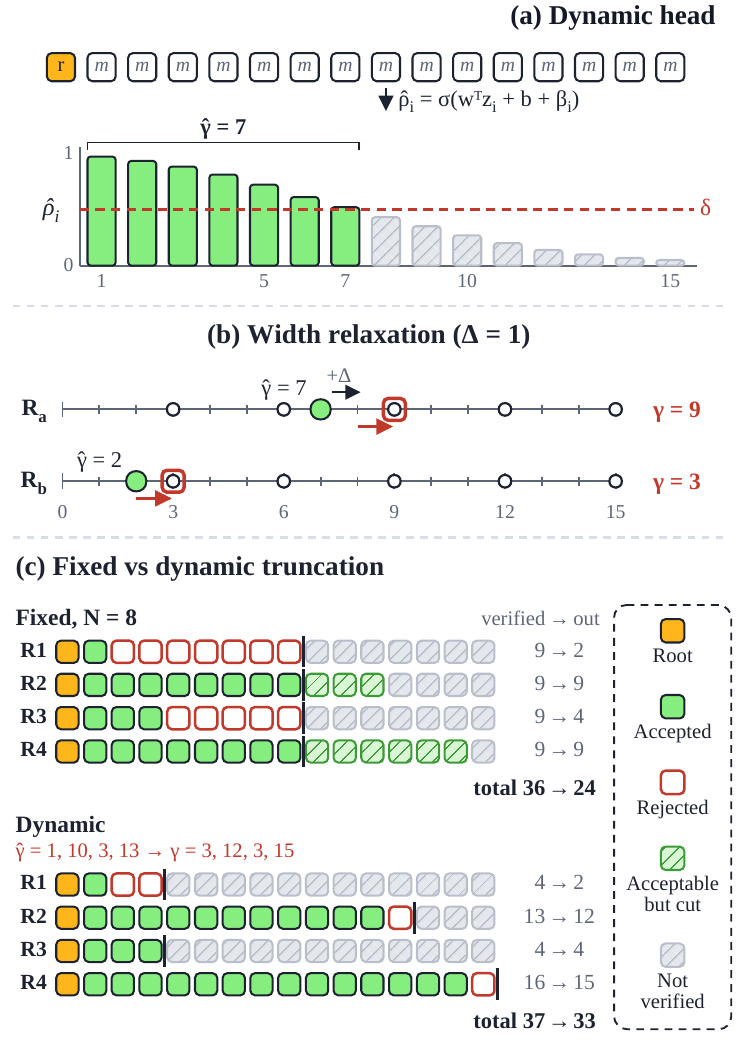}
    \caption{Dynamic truncation of linear drafts using predicted survival
    probabilities. (a)~The dynamic head estimates $\hat{\rho}_i$ for each
    draft position. The longest prefix satisfying $\hat{\rho}_i\ge\delta$
    defines the raw width $\hat{\gamma}$.
    (b)~Width relaxation maps $\hat{\gamma}$ to an admissible draft width
    $\gamma$. Slack $\Delta$ is applied only above the first bucket, followed
    by upward bucket quantization.
    (c)~Comparison with fixed-width truncation across a batch. Adapting widths
    to individual requests reduces verification of rejected candidates and
    retains longer acceptable prefixes.}
    \label{fig:truncation}
\end{figure}

When $\delta>0$ the argument reverses. Verification retains a linear structure,
but survival decays along the block, so its deep positions are mostly verified only to be rejected, and under a
full batch that waste is paid by every request. A single fixed width cannot fix
this, because the right depth is a property of the moment: a request in the
middle of a predictable phrase can be drafted far ahead, while one at a
genuinely uncertain point should barely be drafted at all
(Fig.~\ref{fig:truncation}c).

\begin{figure*}[t]
    \centering
    \includegraphics[width=\textwidth]{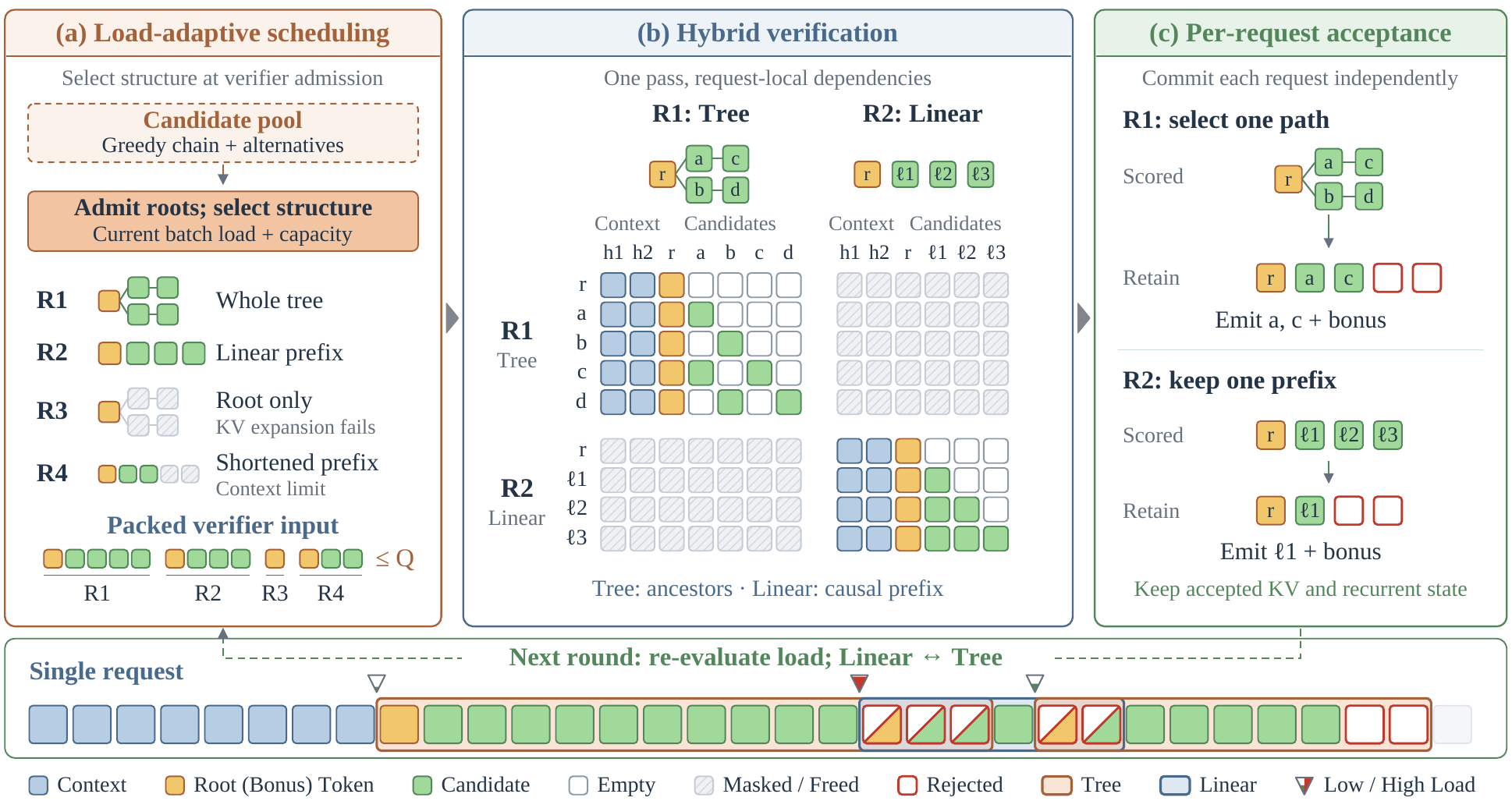}
    \caption{Hybrid verification under variable serving load.
    (a)~At verifier admission, the scheduler selects a draft tree or a linear
    prefix from stored candidates according to current batch load and available
    capacity, subject to the shared token budget $Q$. KV-cache and context
    constraints may reduce a request to its root or a shorter prefix.
    (b)~A single verifier pass processes both structures. The logical attention
    mask isolates requests: tree queries attend to their own cached history,
    ancestors, and themselves, while linear queries follow causal order within
    their request.
    (c)~Acceptance and state retention are resolved independently for each
    request. The accepted tree path or linear prefix is retained with its
    corresponding KV cache and recurrent state. Accepted tokens are committed
    with a bonus token. Structure selection is repeated in the next round as load
    changes.}
    \label{fig:load-adaptive}
\end{figure*}

The drafter already knows how confident it is; we only have to read it out. A
small dynamic head shared by all positions maps each slot state $z_i$ to an estimate of
the survival probability of Eq.~\eqref{eq:survival},
\begin{equation}
    \hat{\rho}_i=\sigma\big(w^{\top}z_i+b+\beta_i\big)
    \quad\text{or}\quad
    \hat{\rho}_i=\prod_{t=1}^{i}\sigma\big(w^{\top}z_t+b+\beta_t\big),
    \label{eq:length-head}
\end{equation}
the second form multiplying conditional survival estimates, which makes the
estimated curve monotone by construction. Sharing the head across
positions is what makes a single checkpoint serve any horizon. The width kept
for a request is then the depth at which survival stops paying for verification
(Fig.~\ref{fig:truncation}a--b):
\begin{equation}
\begin{aligned}
    \hat{\gamma}&=\sum_{i=1}^{\Gamma}\prod_{t=1}^{i}
    \mathbb{1}\big[\hat{\rho}_t\ge\delta\big],\\
    \Delta_{\mathrm{eff}}&=\Delta\,\mathbb{1}[\hat{\gamma}>b_1],\\
    \gamma&=\mathcal{Q}\big(\mathrm{clip}(\hat{\gamma}+\Delta_{\mathrm{eff}},\,
    \gamma_{\min},\,\Gamma)\big),
\end{aligned}
    \label{eq:truncation}
\end{equation}
where the threshold is the marginal cost $\delta(W)$ of
Eq.~\eqref{eq:marginal-threshold}, so the same scalar that describes the load
decides the depth. For $N_b>0$, the bucket edges are
$\mathcal{B}=\{\lceil i\Gamma/N_b\rceil:i=1,\dots,N_b\}$ and
$b_1=\min\mathcal{B}$. The map $\mathcal{Q}$ rounds a positive width up to
the next edge and preserves zero. Predictions in the first bucket skip
slack; for $\Gamma=15$ and $N_b=5$, $\hat{\gamma}=3$ therefore remains 3.
Without bucketing, $\mathcal{Q}$ is the identity and
$\Delta_{\mathrm{eff}}=\Delta$. The configured floor $\gamma_{\min}$ is
applied before rounding in either case.
The relaxation is one-sided relative to $\hat{\gamma}$:
slack $\Delta$ and rounding up to one of a few coarse widths only extend the
predicted prefix. This reduces the risk of truncating acceptable tokens when
$\hat{\gamma}$ is underestimated, while limiting the verifier to a small set
of widths.

The dynamic head is trained jointly with the drafter on labels that are produced by
drafting itself, $y_{a,i}=\prod_{t\le i}\mathbb{1}[\arg\max q_{a,t}=x_{a+t}]$,
i.e., whether the whole prefix up to $i$ matched:
\begin{equation}
    \mathcal{L}_{\mathrm{len}}=-\sum_{(a,i)\in\Omega}
    \big[\omega\,y_{a,i}\log\hat{\rho}_{a,i}
    +(1-y_{a,i})\log(1-\hat{\rho}_{a,i})\big].
    \label{eq:length-loss}
\end{equation}
The two kinds of mistake are not equally expensive: cutting a position that
would have been accepted costs an output token, while keeping a doomed one
costs only a little verifier width. The weight $\omega$ reflects this
asymmetry during training, while the one-sided relaxation provides an
additional safeguard at inference time. The length term is
added to $\mathcal{L}_{\mathrm{tok}}$ once the drafter is good enough for its
own acceptance pattern to be a meaningful label.

Verification then concentrates on positions that are likely to survive, so the batch
spends its width on tokens it will keep. Because the width is per request and
driven by the current load, the same checkpoint covers the whole range from a
single idle request to a saturated batch.

\subsection{Load-Adaptive Hybrid Scheduling}
\label{sec:hybrid-scheduling}

Uni-LaSpec chooses a verification structure \emph{per request and per round},
allowing trees and linear drafts to coexist in one batch. Draft generation
does not fix this choice: one drafter pass supplies the greedy chain,
per-position top-ranked alternatives, and dynamic-head estimates for the
configured load thresholds. These mode-independent candidates are retained
until the request is admitted to a verifier batch. The scheduler then uses
that batch's current load and resource limits to select each request's
verification structure and width
(Fig.~\ref{fig:load-adaptive}). It does not use a request-lifetime mode lock
or the previous round's width to select the next round's structure.

\textbf{Selecting structures at verifier admission.}
The scheduler first admits requests in serving order and reserves their roots
and required history. Let $\mathcal{T}_t$ and $\mathcal{L}_t$ denote admitted
requests assigned draft trees and linear drafts, respectively, and let $W_t^{\mathrm{bg}}$
account for other admitted work, including prefill and roots without ready
candidates. Both structures consume the same per-step token budget $Q$:
\begin{equation}
    W_t=W_t^{\mathrm{bg}}+\sum_{r\in\mathcal{T}_t}(1+B_t)
       +\sum_{r\in\mathcal{L}_t}(1+\tilde{\gamma}_t^{(r)})\le Q,
    \label{eq:hybrid-budget}
\end{equation}
where $\tilde{\gamma}_t^{(r)}\le\gamma_t^{(r)}$ is the linear prefix actually
admitted. A tree budget is eligible only if the complete planned width fits
both $W^{\star}$ and $Q$, and each tree fits its request's remaining context
capacity. For a decode-only all-tree batch, this reduces to
$R_t(1+B_t)\le\min(W^{\star},Q)$, with $R_t$ counting admitted requests,
not all live requests. The planner uses the largest eligible configured
budget; when draft trees coexist with linear drafts, it uses the
smallest configured tree budget. The node budget $B_t$ counts all branches
and is independent of the drafter's depth horizon $\Gamma$.

Linear widths are selected with the current batch's threshold $\delta(W_t)$.
Because truncation changes $W_t$, the planner evaluates the configured
threshold buckets for a self-consistent width; if no fixed point exists, it
uses the most restrictive required threshold among the evaluated profiles.
Linear prefixes may be shortened further to fit the remaining budget.
Once the structures and widths are chosen, the scheduler allocates the additional KV
slots. If an expansion fails, that request keeps its already admitted root
and drops its candidates, rather than preempting another request merely to
verify more candidates (Fig.~\ref{fig:load-adaptive}a). The finalized plan is fixed
for this verifier pass; the cost knee selects affordable branching, while
$Q$ bounds all work actually admitted to the step.

\textbf{Preserving dependencies within the batch.}
The admitted roots and candidates are packed into one verifier input, but
storage order does not define their attention dependencies. Let $r(u)$ be
the request containing token $u$, $\mathcal{H}^{(r)}$ its committed history,
and $v\preceq_r u$ mean that $v$ lies on $u$'s prefix, including $u$ itself.
For full attention, the mixed mask is
\begin{equation}
    M_{uv}=\mathbb{1}[r(u)=r(v)]\,
    \mathbb{1}\big[v\in\mathcal{H}^{(r(u))}\ \lor\ v\preceq_{r(u)}u\big].
    \label{eq:hybrid-mask}
\end{equation}
Within a linear request, this is the usual causal mask; within a tree, it
exposes only ancestors and the query itself. Both can read their own cached
history, while cross-request entries are masked
(Fig.~\ref{fig:load-adaptive}b). Tree positions follow depth, so siblings
share a position even when stored in different slots; linear positions
increase along the chain. Layers with a local attention window additionally
apply that window. Recurrent layers preserve the same dependency: a tree node
advances from its parent's state, and a linear node from its predecessor's.

\textbf{Independent acceptance and continuation.}
One verifier forward pass scores the mixed input. Each tree then contributes
the path selected by Eq.~\eqref{eq:tree-walk}, and each linear draft contributes
its longest accepted prefix. The corresponding cached tokens and recurrent
state are retained per request; rejected branches and suffixes do not become
committed history. Each request emits its accepted tokens and a bonus token,
which roots its next round (Fig.~\ref{fig:load-adaptive}c).
This committed history is shared by both structures. Thus an ongoing request
can switch from Linear to Tree when load falls, or from Tree to Linear when
load rises, at the next verifier-admission boundary. No switch occurs within
a verifier pass, and rejected candidates from the previous
round cannot enter the next round's context.

\subsection{Decoding Round}
\label{sec:optimization}
A round proceeds as follows. The verifier scores the root and the current
candidates, and in doing so leaves behind the context features, the visual keys
and the query that Path~I needs. The selector votes on the visual tokens, and
one drafter pass returns a distribution for every position of the block
together with its survival estimates. The drafter retains mode-independent
candidates without fixing the next verification structure. At the next
verifier admission, the scheduler uses the admitted batch's load and
available capacity to select a draft tree, a dynamically truncated linear
prefix, or the root alone for each request. The verifier evaluates the resulting
batch, which may mix both structures, in a single pass; the accepted prefix and a bonus
token are committed, with the bonus token serving as the root of the next
round. Drafting costs one pass regardless of the horizon, and every component
of the method changes only which candidates are offered, so the
emitted text is that of the verifier.

\section{Experiments}
\label{sec:experiments}

The evaluation follows the two paths of Uni-LaSpec.
Section~\ref{sec:visual-experiments} fixes verification to a full-width linear
draft and measures what attention-guided visual evidence contributes to
drafting and what it costs. Section~\ref{sec:dynamic-experiments} fixes the
drafter and evaluates load-adaptive verification against fixed structures
under uniform load, bursty arrivals, and different GPUs.

\subsection{Experimental Setup}
\label{sec:setup}

\textbf{Verifiers and workloads.}
We evaluate six frozen VLM verifiers: Qwen3.5-4B/9B (Q3.5),
InternVL3.5-4B/8B (IVL3.5), and Gemma~4 E2B/E4B (G4).
The seven text benchmarks cover mathematics, code, and dialogue:
GSM8K~\cite{cobbe2021gsm8k},
MATH-500~\cite{hendrycks2021math,lightman2023math500}, AIME25,
HumanEval~\cite{chen2021humaneval}, MBPP~\cite{austin2021mbpp},
LiveCodeBench~\cite{jain2024livecodebench}, and MT-Bench~\cite{zheng2023mtbench}.
The ten multimodal benchmarks cover documents, charts, visual reasoning,
captioning, and dialogue: OmniDocBench~\cite{ouyang2024omnidocbench},
OCRBench-v2~\cite{fu2024ocrbenchv2}, DocVQA~\cite{mathew2020docvqa},
olmOCR~\cite{poznanski2025olmocr}, CharXiv~\cite{wang2024charxiv},
TextVQA~\cite{singh2019textvqa}, MMMU-Pro~\cite{yue2024mmmupro},
GQA~\cite{hudson2019gqa}, COCO Caption~\cite{chen2015cococaptions},
and MM-MT-Bench~\cite{agrawal2024pixtral}.
The business-model experiment additionally evaluates two verifiers
fine-tuned on in-house business data for schema-guided extraction.

\textbf{Drafters and baselines.}
The visual drafter is trained on one million prompts: 600K sampled uniformly
from the twelve categories of LLaVA-OneVision-1.5~\cite{an2025llavaonevision15}
and 400K from NVIDIA Nemotron~\cite{nathawani2025nemotron} and
CodeAlpaca~\cite{chaudhary2023codealpaca}. Responses are regenerated by the
corresponding verifier. The six-layer drafter uses 32 attention heads for
Q3.5/IVL3.5 and eight for G4, a compressed 32{,}768-token vocabulary, and
eight auxiliary verifier layers. Its default visual selector retains eight
tokens from three full-attention layers, with 128 nominees per layer.
Training uses ten epochs, Muon, a learning rate decayed from $10^{-4}$ to
$10^{-6}$, and the position weighting of Eq.~\eqref{eq:position-weight}
with $\alpha{=}0.5$. The visual experiments compare autoregressive decoding
(AR), EAGLE-3 for the two Qwen3.5 verifiers, native MTP heads, and DFlash
trained on text (T) or multimodal data (M). DFlash uses the same
block-drafter architecture without a direct visual-input branch.
EAGLE-3 and MTP draft autoregressively and use 50{,}000-token and full 248{,}320-token vocabularies,
respectively. The dynamic-verification experiments use checkpoints from an
additional dynamic-head training stage; all verification strategies for a
given verifier share that checkpoint and visual-input policy.

\textbf{Protocol and metrics.}
Each general benchmark supplies at most 100 prompts, outputs are capped at
1024 tokens, and decoding is greedy unless
$T{=}1$ is specified. The business workload contains 100 held-out prompts
and uses a 4096-token cap. The drafter predicts $\Gamma{=}15$ candidate
positions given the root token. Fixed-load sweeps use client concurrency
$R\in\{1,4,8,16,32,64\}$; the selector-cost experiment uses the five loads
shown in its table. We report accepted length
$\tau=\mathbb{E}[a]+1$, including the bonus token, and output throughput
over elapsed benchmark time, including prefill and decoding.
Per-benchmark speedup divides speculative throughput by AR throughput for
the same verifier, sampling temperature, concurrency, and prompts.
For text and multimodal summaries, we first take the unweighted mean
throughput over the benchmarks in that group and then divide by the
corresponding AR mean. Avg.\ $\tau$ additionally averages over the six loads.
The adaptive policy uses tree budgets $\{15,31,63\}$, five linear-length
buckets with slack $\Delta{=}1$, the profiled knee $W^{\star}$ of
Eq.~\eqref{eq:knee}, and the piecewise-constant truncation profile
$\delta(W)$ of Eq.~\eqref{eq:marginal-threshold}.

\textbf{Hardware.}
The general visual-drafting comparisons use RTX 6000 Ada GPUs (48\,GB) for
Q3.5-4B, IVL3.5-4B/8B, and G4-E4B, and RTX A6000 GPUs (48\,GB) for
Q3.5-9B and G4-E2B. The business-model and visual-input-policy
experiments use RTX A6000 GPUs; selector implementation costs and the
auxiliary-feature ablation use RTX 6000 Ada GPUs.
The main verification-strategy comparison uses RTX A6000 GPUs;
Table~\ref{tab:verify-concurrency-devices} repeats the concurrency study on
an RTX 5090 (32\,GB), an A800 (80\,GB), and an H100 (80\,GB).
Image preprocessing follows each verifier's processor.

\subsection{Visual-Aware Drafting}
\label{sec:visual-experiments}
\label{sec:ablation}

All comparisons in this subsection verify the full 15-position block with a
linear draft; dynamic truncation and draft trees are disabled. Differences
among drafters therefore reflect the quality of their proposals alone, while
the measured throughput includes the complete cost of producing them,
including visual selection.

\subsubsection{Acceptance and speedup across verifiers, benchmarks, and loads}
Table~\ref{tab:ablation-visual} reports the accepted length per benchmark at
$R{=}1$. Uni-LaSpec exceeds DFlash~(M), which shares its architecture and
training data but has no direct visual input, on 198 of the 204
verifier--temperature--benchmark entries and on 117 of the 120 multimodal
entries. At $T{=}0$, the multimodal mean of Q3.5-4B rises from 3.80 to 4.07
accepted tokens and that of G4-E2B from 2.97 to 3.27. Across the six verifiers
and both temperatures, the multimodal gain over DFlash~(M) ranges from 4.8\%
to 10.1\%, roughly twice the 2.2--6.5\% gain on text, and the gain over the
text-trained DFlash~(T) ranges from 12.7\% to 31.7\%. The same ordering holds in all three
model families, so the improvement does not depend on the verifier architecture;
Table~\ref{tab:visual-input-policy} isolates the visual branch itself.

The distribution of these gains shows where visual evidence matters. The
eight largest per-benchmark gains at $T{=}0$ all occur on COCO Caption and
GQA, whose answers name objects and attributes visible in the image: IVL3.5-8B
improves from 2.41 to 2.93 accepted tokens on COCO Caption (+21.6\%), and
G4-E4B from 2.71 to 3.25 on GQA (+19.9\%). These are the tokens that a
language prior cannot supply and that the selected patches can. Text
acceptance also improves for every verifier at both temperatures, e.g., from
5.24 to 5.40 for Q3.5-4B, showing that the zero-initialized visual increment
and acceptance-aware training strengthen ordinary continuations rather than
trading them for visual ones. Against EAGLE-3, an autoregressive
feature-level drafter, Uni-LaSpec accepts 1.52--1.73$\times$ as many tokens
on the multimodal group at $T{=}0$.

\begin{table*}[tp]
\caption{\normalfont Mean accepted length $\tau$ per benchmark at concurrency $R{=}1$, reported by
verifier, drafter, and sampling temperature. DFlash (T) and DFlash (M) denote
training on text-only prompts and the full multimodal mixture, respectively.
Benchmark abbreviations: MATH (MATH-500), AIME (AIME25), HEval (HumanEval),
LCB (LiveCodeBench), MT-B (MT-Bench), ODoc (OmniDocBench), OCR2 (OCRBench-v2),
DVQA (DocVQA), olmO (olmOCR), CXiv (CharXiv), TVQA (TextVQA), MMMU (MMMU-Pro),
and MMMT (MM-MT-Bench).}
\label{tab:ablation-visual}
\centering
\input{tables/table1}
\end{table*}

Table~\ref{tab:speedup-c1} shows that this acceptance translates into
measured speed. Uni-LaSpec accelerates every one of the 204 single-request
benchmark runs, with speedups of 1.32--6.45$\times$ over AR, and it is the
fastest of EAGLE-3 and the two DFlash drafters in 190 of them. EAGLE-3
and the native MTP heads draft autoregressively: each proposed position costs
another drafter step, so drafting cost grows with the draft length and
offsets the benefit of a longer accepted prefix. MTP, which also uses the
full verifier vocabulary, accepts 6.11 tokens on GSM8K with Q3.5-4B at
$T{=}0$, against 5.35 for Uni-LaSpec, yet reaches only 1.68$\times$ AR
throughput, against 3.58$\times$. Block-parallel drafting, as in
DFlash, removes this bottleneck by filling all mask positions in one forward
pass at a cost nearly independent of the number of proposed tokens;
Uni-LaSpec inherits this property and performs its visual conditioning
within the same pass. Across the paired benchmarks, it is
1.70--4.02$\times$ faster than MTP and 1.59--3.31$\times$ faster than
EAGLE-3. Uni-LaSpec thus improves both the numerator and the denominator of
Eq.~\eqref{eq:latency}.

\begin{table*}[tp]
\caption{\normalfont Speedup relative to autoregressive decoding at concurrency $R{=}1$, reported
by benchmark, verifier, drafter, and sampling temperature. Speedup is defined
as the ratio of speculative to autoregressive output throughput for the same
verifier, temperature, and prompt set. Benchmark abbreviations follow
Table~\ref{tab:ablation-visual}.}
\label{tab:speedup-c1}
\centering
\input{tables/table2}
\end{table*}

Table~\ref{tab:throughput-concurrency} aggregates throughput by benchmark
group across client concurrency. Uni-LaSpec delivers the highest throughput
among EAGLE-3 and the DFlash drafters in all 144
verifier--temperature--modality--load settings, and it also exceeds the MTP
heads in all 48 settings in which they are available, so the advantage of
parallel over autoregressive drafting persists under concurrency. At $R{=}1$ and $T{=}0$,
it reaches 3.60--5.00$\times$ AR throughput on text and 2.35--2.88$\times$
on multimodal benchmarks; at $T{=}1$, the corresponding ranges are
2.70--4.57$\times$ and 1.87--2.48$\times$. Its margin over DFlash~(M) is
preserved at every load, so its additional acceptance comes without a cost
that grows with concurrency. Since these
drafters all verify a fixed 15-token block, Section~\ref{sec:dynamic-experiments}
shows how load-adaptive verification raises throughput further under load.

\begin{table*}[tp]
\caption{\normalfont Output throughput ratios relative to autoregressive decoding
across client concurrency $R$ and sampling temperature. Each ratio is the
quotient of unweighted mean output throughputs over the same benchmark group,
evaluated for the same verifier, $R$, and temperature. Avg.\ $\tau$ denotes
the unweighted mean accepted length across benchmarks within the group and
all six concurrency levels. A ratio below unity indicates lower throughput
than autoregressive decoding.}
\label{tab:throughput-concurrency}
\centering
\input{tables/table3}
\end{table*}

\subsubsection{Why a text-trained drafter accepts visual-task tokens}
\begin{figure*}[t]
    \centering
    \includegraphics[width=\textwidth]{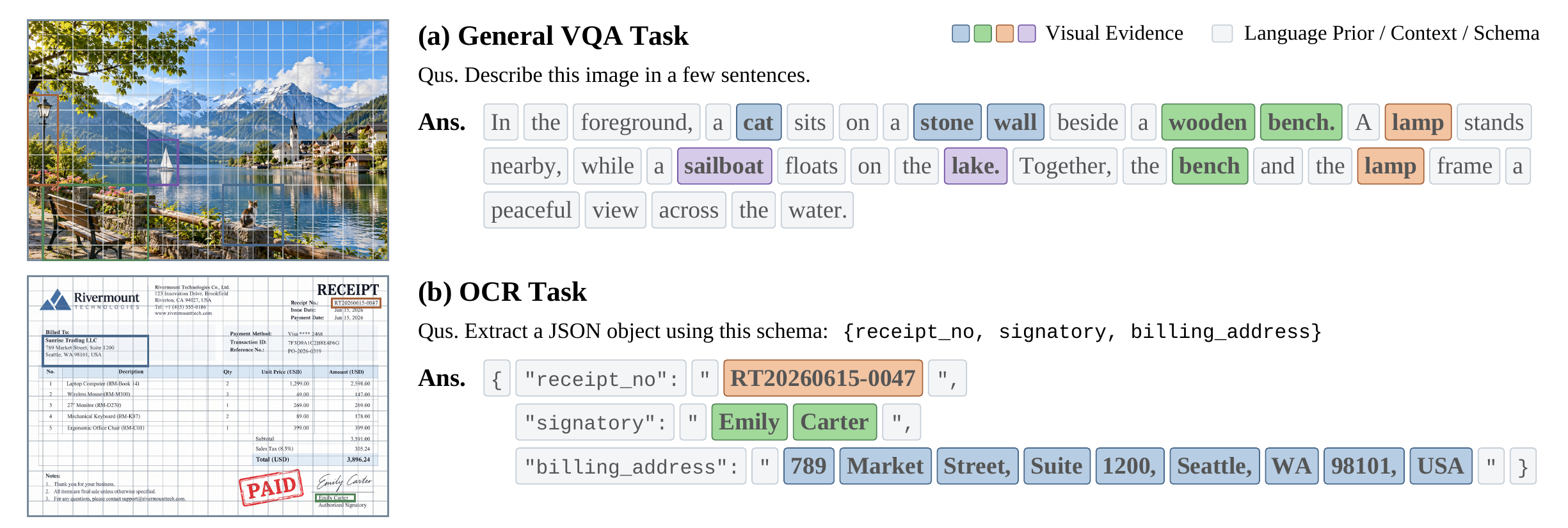}
    \caption{Visual grounding in general VQA and schema-guided OCR.
    (a)~Scene entities and attributes are grounded in image regions, while
    language priors and context support the surrounding description.
    (b)~The schema constrains field names and output structure, while exact
    field values depend on local document regions. The relevant regions change
    as generation proceeds across fields. Colors associate output spans with
    supporting visual regions; gray spans indicate language context or schema
    structure.}
    \label{fig:vqa-ocr-dependence}
\end{figure*}

Table~\ref{tab:ablation-visual} also shows that DFlash~(T), which never sees
an image during training, accepts a substantial number of tokens on the
multimodal benchmarks. Fig.~\ref{fig:vqa-ocr-dependence} explains this at
the token level. A visual response interleaves two kinds of spans.
\emph{Language spans}, such as function words, the syntax of a description,
and the braces, field names, and separators of a schema, are determined by
the question, the requested format, and the generated prefix.
\emph{Grounded spans}, such as object names and attributes in VQA or receipt
numbers, signatories, and addresses in OCR, can only be read from a specific
image region. A block drafter conditioned on verifier hidden states inherits
the multimodal context that the verifier has already processed, so it
predicts language spans well even without visual training.

Acceptance is lost on the grounded spans, and these are exactly what the
visual branch of Uni-LaSpec targets. Moreover, the relevant region shifts from one entity or
field to the next (Fig.~\ref{fig:vqa-ocr-dependence}b), so the evidence a
block requires is local and time-varying. Uni-LaSpec therefore re-selects a
few patches for every block from the verifier's current query, and its
largest gains appear on entity-dense captioning and VQA.

\subsubsection{Vocabulary coverage and domain-specialized extraction}
\begin{figure*}[t]
    \centering
    \includegraphics[width=\textwidth]{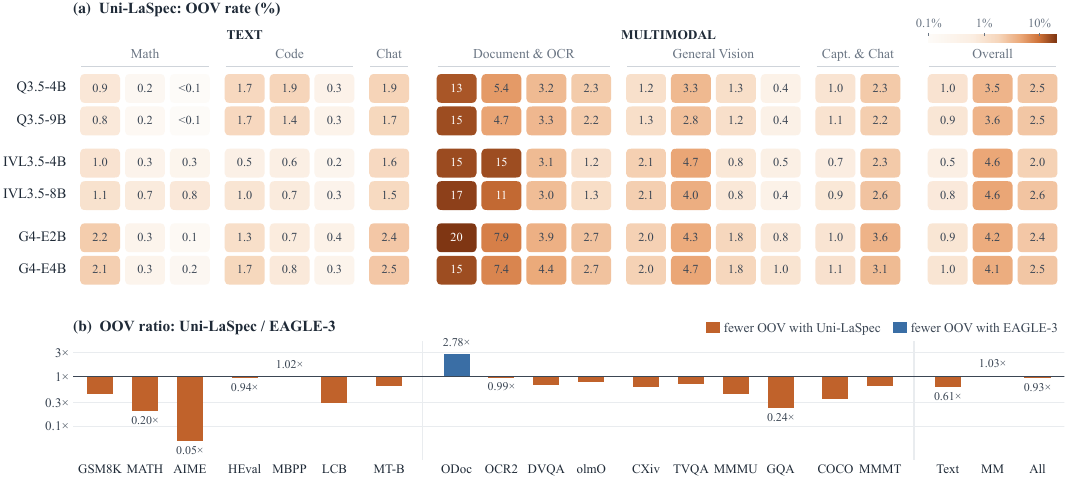}
    \caption{Out-of-vocabulary (OOV) rates of generated outputs, defined as
    the fraction of tokens outside the draft vocabulary.
    (a)~Uni-LaSpec rates by verifier and benchmark; Overall columns use
    token-weighted aggregation.
    (b)~Ratio of the Uni-LaSpec OOV rate to that of EAGLE-3, aggregated using
    the geometric mean over paired verifiers. Ratios below $1\times$ indicate
    lower OOV rates for Uni-LaSpec. Rates are approximate estimates from saved
    outputs re-tokenized with each verifier's tokenizer.
    Benchmark abbreviations follow Table~\ref{tab:ablation-visual}.}
    \label{fig:vocab-oov}
\end{figure*}

Document and OCR benchmarks show the shortest accepted lengths in
Tables~\ref{tab:ablation-visual}--\ref{tab:throughput-concurrency}.
Fig.~\ref{fig:vocab-oov} attributes a large part of this to the draft
vocabulary rather than to visual drafting. The compressed vocabulary keeps
the 32{,}768 most frequent tokens of the drafter's training responses, which
the verifier generates for general multimodal instructions and text prompts.
OCR outputs instead transcribe the language printed in the document image:
proper names, identifiers, formatted numbers, formulas, tabular markup, and
pages in languages and scripts that are rare in the training responses, such
as the Chinese documents in OmniDocBench and OCRBench-v2. Their token
frequencies can therefore differ sharply from those that defined the
vocabulary. The vocabulary covers ordinary outputs well: only 0.5--1.0\% of
generated text tokens and 3.5--4.6\% of multimodal tokens fall outside it.
On OmniDocBench, however, 13.5--19.5\% of output tokens are out of
vocabulary, and on OCRBench-v2 the rate reaches 14.8\%. A token that the
drafter cannot emit terminates the accepted prefix regardless of which
patches were selected, so acceptance on these benchmarks is bounded by
vocabulary coverage rather than by visual evidence. Despite
using a smaller vocabulary than EAGLE-3 (32K versus 50K entries), the draft
vocabulary of Uni-LaSpec yields a lower out-of-vocabulary rate on text
(1.0\% versus 1.6\% for Q3.5-4B) and a comparable rate on multimodal outputs.

Table~\ref{tab:business-concurrency} removes this restriction in a deployment
setting. Two Qwen3.5 verifiers fine-tuned on in-house business data for
schema-guided medical-document extraction are paired with full-vocabulary drafters (248{,}320 tokens) that
fuse three auxiliary verifier layers. Averaged over the six loads, Uni-LaSpec
accepts 9.92 tokens per round on both models, out of at most 16, compared with
9.33--9.37 for DFlash~(M) and 5.55--5.66 for the MTP heads. An autoregressive
drafter conditions each position on its preceding draft tokens and would
ordinarily be expected to predict more accurately than a block drafter.
However, the MTP drafter of Qwen3.5 is a single transformer layer, applied
recurrently with shared weights to every draft position; this limited
capacity, compounded by its own errors along the unrolled chain, caps its
accepted length near 5.6. The block drafter of Uni-LaSpec has six layers and
reads features fused from several verifier layers, so it follows the
structured outputs of this task almost ten tokens at a time. It delivers the
highest throughput at every concurrency on both models: 5.59$\times$ and
5.70$\times$ AR throughput for a single request, and 1.80$\times$ and
1.90$\times$ at $R{=}64$, where the MTP heads reach 0.54$\times$ and
1.18$\times$. Once the vocabulary matches the domain, block drafting with
attention-guided evidence sustains very long accepted blocks on document
extraction, and the visual branch continues to add acceptance on top of the
strongest baseline.

\input{tables/table4}

\subsubsection{How much visual evidence a draft block needs}
Table~\ref{tab:visual-input-policy} varies the visual input of the drafter
while the verifier always processes the complete image. For a fixed
Uni-LaSpec checkpoint, we compare forwarding all visual features with
attention-guided Top-8, Top-32, and Top-64 selections, and with disabling
the visual branch (None). Top-8 votes over three
full-attention layers, Top-32 over five evenly spaced layers, and Top-64 over
the first full-attention layer at or after each auxiliary-feature location,
counting duplicates once.

Selecting only eight visual tokens raises multimodal acceptance over None by
0.14--0.21 tokens (4.2--6.5\%) on all six verifiers, e.g., from 3.86 to 4.06
for Q3.5-4B and from 3.21 to 3.42 for G4-E4B. Top-8 also retains at least
98.6\% of the acceptance obtained by feeding the drafter every visual token,
and enlarging the selection to 32 or 64 tokens changes multimodal acceptance
by at most 0.06 tokens. A 15-token draft block therefore depends on a small,
local set of patches, and the verifier's own attention identifies that set
reliably: a vote over three layers recovers nearly all of the useful evidence,
while the conditioning cost of the drafter scales with eight tokens rather
than with the full image sequence. The experiment thus confirms both premises
of Path~I, namely that the visual dependency of a draft block is sparse and
that verifier attention is the right signal for locating it.

\begin{table*}[tp]
\caption{\normalfont Comparison of visual input policies at concurrency $R{=}32$
and sampling temperature $T{=}0$.
None disables the visual branch; All provides the complete
set of visual features to the drafter. Quant.\ and Softmax correspond to int8
ranking and full softmax scoring, respectively. Text and MM report unweighted
benchmark averages of accepted length $\tau$ for text and multimodal tasks.
$C_{\mathrm{sel}}$ (ms) is the mean wall-clock time per selector invocation
on multimodal inputs; $\kappa$ is the ratio of Softmax to Quant.\ selector time.}
\label{tab:visual-input-policy}
\centering
\input{tables/table5}
\end{table*}

\subsubsection{Making patch selection inexpensive}
Tables~\ref{tab:visual-input-policy} and~\ref{tab:visual-branch-impl}
quantify the engineering choices that keep visual selection off the critical
path. First, selection requires only the ranking of attention scores, not
their normalized values. Replacing full FP32 softmax scoring with int8 ranking
changes Top-8 multimodal acceptance by at most 0.02 tokens on every verifier,
while reducing selector time from 2.44--18.29\,ms to 0.54--0.73\,ms, a
3.45--25.35$\times$ reduction that reaches 19.17$\times$ and 25.35$\times$ on
the two Qwen3.5 verifiers (Table~\ref{tab:visual-input-policy}).

Second, Table~\ref{tab:visual-branch-impl} shows how selection cost behaves
under load and how kernel fusion reduces it further. With full scoring and
separate operator launches, selector time grows with concurrency, from 0.99
to 8.95\,ms for Q3.5-4B and from 1.03 to 15.71\,ms for Q3.5-9B between
$R{=}1$ and $R{=}64$. Quantized ranking removes most of this cost, and fusing
scoring, voting, gathering, and cross-attention into a single execution
additionally removes launch and synchronization overhead. The fused selector
stays below 1\,ms at every measured load (0.38--0.94\,ms), is
2.58--18.04$\times$ faster than unfused full scoring, and leaves the accepted
length unchanged. At $R{=}32$, for instance, selector time falls from 8.61 to
0.76\,ms for Q3.5-4B and from 13.98 to 0.78\,ms for Q3.5-9B. Visual selection
therefore stays below one millisecond at every measured load, which allows the acceptance gains of
Tables~\ref{tab:ablation-visual}--\ref{tab:throughput-concurrency} to appear
in full as throughput.

\begin{table*}[tp]
\caption{\normalfont Latency of visual selection across scoring rules and kernel
implementations at concurrency $R$. $C_{\mathrm{sel}}$ (ms) is the mean
wall-clock time per selector invocation. The time ratio $\eta_{\mathrm{sel}}$
is defined as the unfused Softmax latency divided by the latency of the
evaluated configuration, for the same verifier and $R$.
Reported $\tau$ values are unweighted averages over multimodal benchmarks and concurrency
levels.}
\label{tab:visual-branch-impl}
\centering
\input{tables/table6}
\end{table*}

\subsubsection{Auxiliary-feature ablation}
Table~\ref{tab:auxiliary-top8} varies the number of verifier layers
$|\mathcal{A}|\in\{3,5,8\}$ whose hidden states the drafter fuses, using
Q3.5-4B, the same inference-time Top-8 selector over layers $\{11,19,27\}$,
and fixed-width verification at $R{=}4$. The drafter network remains six
layers deep in every configuration; the eight-feature drafter is the main
Uni-LaSpec checkpoint, and the smaller configurations use their
minimum-validation-loss checkpoints.

All three budgets deliver strong acceleration, 2.56--3.40$\times$ on text and
1.97--2.47$\times$ on multimodal benchmarks, and at each temperature the mean
accepted length varies by at most 0.05 tokens on text and 0.10 tokens on
multimodal inputs. The benefit of visual drafting therefore does not depend on
a large auxiliary-feature budget: three verifier layers already supply the
context that the drafter needs to exploit the selected patches. The
eight-feature default attains the highest acceptance and throughput at
$T{=}0$, 5.38 and 4.08 accepted tokens with 3.40$\times$ and 2.47$\times$
speedups on text and multimodal benchmarks, and the highest multimodal
acceptance at both temperatures, which is why we adopt it.

\input{tables/table7}

\subsection{Load-Adaptive Draft Verification}
\label{sec:dynamic-experiments}

We now hold the drafter fixed and vary only how its proposals are verified.
All strategies for a verifier share the same visual drafter, checkpoint,
prompts, and generation limits. Linear verifies the complete 15-token draft;
fixed truncation retains $N\in\{5,7\}$ tokens; dynamic truncation applies a
survival threshold of $\delta{=}0.3$; and fixed trees verify
$B\in\{15,31,63,127\}$ candidate nodes. Uni-LaSpec selects the structure and
the linear length per request and per round from the admitted work. The
$B{=}127$ control extends branching beyond the candidate budgets of the
adaptive policy.

\subsubsection{Acceptance, single-request speed, and throughput under load}
Table~\ref{tab:verify-tau} shows that branching raises acceptance
monotonically with the tree budget for all six verifiers in both modalities.
For Q3.5-4B, the mean text acceptance grows from 5.3 tokens with Linear to
6.2 with Tree $B{=}15$ and 7.3 with $B{=}127$, and the multimodal acceptance
from 4.1 to 4.7 and 5.6. At $R{=}1$, where verifier capacity is idle,
Uni-LaSpec verifies draft trees and matches the acceptance of Tree $B{=}15$.
The increments, however, shrink as the tree grows: expanding from 16 to 128
verified positions adds about one accepted token for Q3.5-4B. Additional
candidates must therefore be justified by their verification cost, which is
precisely what the marginal-cost rule of Eq.~\eqref{eq:marginal-threshold}
evaluates.

\begin{figure*}[!b]
    \centering
    \includegraphics[width=0.90\textwidth]{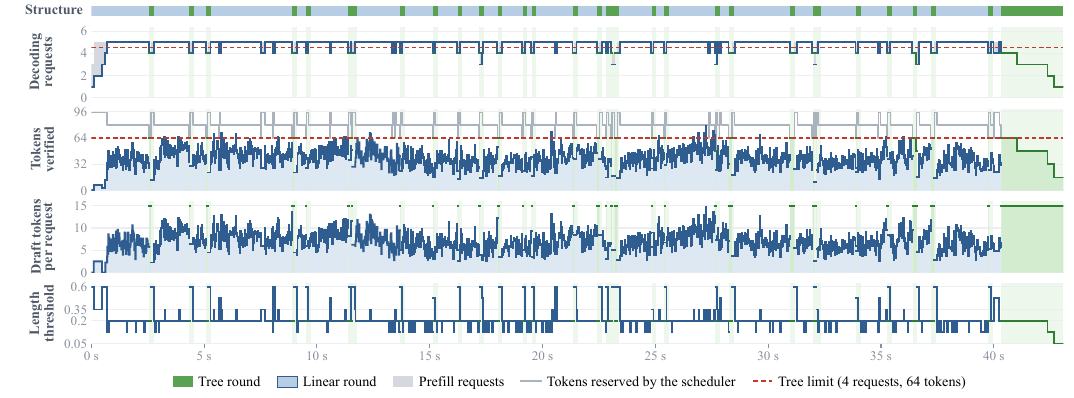}
    \caption{Scheduling decisions and verification workload of Uni-LaSpec.
    From top to bottom: verification structure, decoding request count, tokens
    verified for decoding, mean draft tokens verified per request, and the
    survival threshold $\delta$ for linear truncation. Green and blue indicate
    tree and linear verification, respectively. Gray bars denote requests
    undergoing prefill; the gray trace shows the scheduler's token reservation
    before draft trimming. Red dashed lines mark the request and token limits
    for tree verification under $R_t(1{+}B)\le W^{\star}$.}
    \label{fig:round-schedule}
\end{figure*}

\begin{table*}[tp]
\caption{\normalfont Mean accepted length $\tau$ of verification strategies at
concurrency $R{=}1$ and sampling temperature $T{=}0$. All strategies for a given verifier share the same
drafter. Linear verifies the full draft; Trunc.\ retains a fixed prefix of
$N$ draft tokens; Dyn.\ trunc.\ selects prefix length from predicted survival
probabilities; Tree verifies $B$ candidate nodes; Uni-LaSpec selects the
verification structure according to load. Benchmark abbreviations follow
Table~\ref{tab:ablation-visual}.}
\label{tab:verify-tau}
\centering
\input{tables/table8}
\end{table*}

Table~\ref{tab:verify-speedup-c1} shows that speed and acceptance rank the
strategies differently. Uni-LaSpec attains the highest single-request speedup
on 74 of the 102 verifier--benchmark entries, and it outperforms linear
verification of the same drafts on every entry, by 1.04--1.50$\times$. Large
trees, in contrast, lose speed despite their acceptance: for Q3.5-4B on GSM8K,
Tree $B{=}127$ accepts the most tokens but reaches 2.31$\times$ AR throughput,
against 4.17$\times$ for Uni-LaSpec. The round-level profile in
Fig.~\ref{fig:step-timeline} identifies the cause. Every added node costs
verifier computation as well as tree construction, KV compaction, and GDN
state replay, whereas the deepest nodes of a large tree carry little survival
probability, so the time per output token rises from 4.48\,ms for Tree $B{=}15$ to 5.38\,ms for $B{=}63$.
Uni-LaSpec attains tree-level acceptance at the lowest time per output token
of all strategies, 4.16\,ms, with the smallest verifier and drafter time per
token in the profile.

\begin{table*}[tp]
\caption{\normalfont Speedup of verification strategies relative to autoregressive
decoding at concurrency $R{=}1$ and sampling temperature $T{=}0$. Speedup is defined as the ratio of each
strategy's output throughput to the corresponding autoregressive throughput
for the same verifier and prompt set. Verification strategies are defined in
Table~\ref{tab:verify-tau}.}
\label{tab:verify-speedup-c1}
\centering
\input{tables/table9}
\end{table*}

Table~\ref{tab:verify-concurrency} extends the comparison to competing
requests and exposes how strongly the best fixed structure depends on load.
Trees give the best fixed-strategy throughput in all 12 model--modality
settings at $R{=}1$ and in 9 of 12 at $R{=}4$, whereas fixed or dynamic
truncation is best in every setting from $R{=}16$ onward. Each fixed
structure is thus tuned to one operating point and degrades away from it:
Tree $B{=}15$ falls below AR throughput in 16 of the 72 settings, and Linear
in 10. Uni-LaSpec selects the structure at verifier admission and, unlike the
fixed $\delta{=}0.3$ of dynamic truncation, sets the threshold from the
current verifier width, tracking the best operating point across the range. It improves on AR in all 72
settings, with 1.17--5.11$\times$ on text and 1.01--3.24$\times$ on
multimodal benchmarks, and it matches or exceeds the best fixed strategy in
61 of the 72 settings, strictly exceeding it in 53. Its gains are largest
where fixed structures are least suited: at $R{=}64$, it raises throughput
over linear verification of the same drafts by 1.16--1.49$\times$, and at
$R\ge32$ it is 1.25--2.15$\times$ faster than Tree $B{=}15$.

The average accepted length of Uni-LaSpec across loads is close to that of
Linear, e.g., 5.33 versus 5.36 on text for Q3.5-4B, yet its throughput is
substantially higher. The gain therefore does not come from accepting more
tokens on average, but from spending verification where it pays: additional
tree candidates when verifier positions are inexpensive, and fewer linear
candidates with higher survival probability when they are not. This is the
allocation that maximizes the throughput objective of
Eq.~\eqref{eq:objective}.

\begin{table*}[tp]
\caption{\normalfont Output throughput ratios of verification strategies relative to
autoregressive decoding across client concurrency $R$ at sampling temperature $T{=}0$. Each ratio
is the quotient of unweighted mean output throughputs over the same benchmark
group, evaluated for the same verifier and $R$. Avg.\ $\tau$ denotes the
unweighted mean accepted length across benchmarks within the group and all
six concurrency levels. A ratio below unity indicates lower throughput than
autoregressive decoding. Verification strategies are defined in
Table~\ref{tab:verify-tau}.}
\label{tab:verify-concurrency}
\centering
\input{tables/table10}
\end{table*}

\subsubsection{Structure selection within a running workload}
Fig.~\ref{fig:round-schedule} traces the decisions behind these aggregates for
Q3.5-4B at client concurrency $R{=}5$. Although five requests are in flight,
the number decoding in a given round varies as requests finish and their
replacements enter prefill, so the workload moves between operating points
from one round to the next. With knee $W^{\star}{=}64$, four decoding requests
with $B{=}15$ trees exactly fill the knee, $4(1{+}15){=}64$, whereas five
would require 80 positions. The scheduler responds accordingly. Every round
with at most three decoding requests and no concurrent prefill is verified as
a tree; every round with five decoding requests is verified linearly; and
every round that shares the batch with a prefill chunk is verified linearly,
because the chunk consumes part of the token budget of
Eq.~\eqref{eq:hybrid-budget}. Within linear rounds, the threshold profile sets
the draft length: requests verify 7.1 draft tokens on average rather than the
full 15, and five-request rounds verify 40.7 tokens instead of the 80 that
trees would require. Tree rounds accept 5.70 tokens per request, against 4.20
in linear rounds, so branching is applied whenever the batch leaves room for
it. Since each decision is made at verifier admission for the work actually
executed, no fixed structure can reproduce this behavior.

\subsubsection{Bursty arrivals and end-to-end latency}
\begin{figure*}[t]
    \centering
    \includegraphics[width=\textwidth]{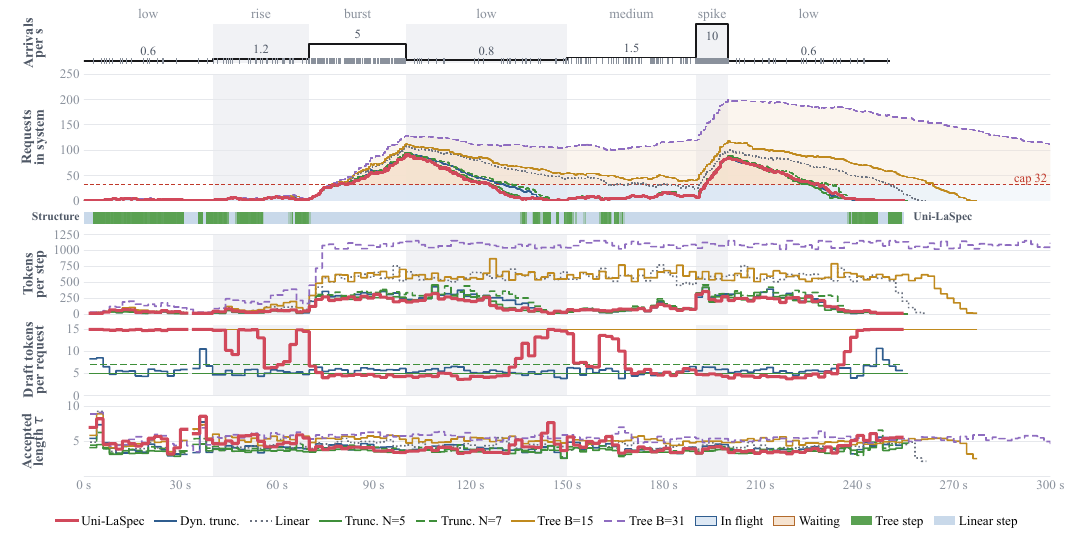}
    \caption{Temporal behavior of verification strategies under bursty arrivals.
    All strategies serve the same arrival trace. From top to bottom: arrival
    rate, outstanding requests, Uni-LaSpec's verification structure, tokens
    verified per step, mean draft tokens verified per decoding request, and
    accepted length $\tau$. Arrival ticks mark individual requests. Pale blue
    and orange shading distinguish requests in flight from those queued at
    the client; the dashed line marks the concurrency cap. The structure strip
    uses green for tree verification and blue for linear verification.}
    \label{fig:bursty-arrivals}
\end{figure*}

Fig.~\ref{fig:bursty-arrivals} replays one open-loop Poisson trace with
low-load intervals, a sustained burst, and a short spike to every strategy
(420 requests over 250\,s on Q3.5-4B, at most 32 in flight); latency
includes queueing at the client.

\begin{table*}[p]
\caption{\normalfont Output throughput ratios of verification strategies relative to
 autoregressive decoding on an RTX 5090 (32\,GB), an A800 (80\,GB), and an H100 (80\,GB),
 across client concurrency $R$ at sampling temperature $T{=}0$. H100 results are shown in the continuation.}
\label{tab:verify-concurrency-devices}
\label{tab:verify-concurrency-5090}
\label{tab:verify-concurrency-a800}
\centering
\input{tables/table11}
\end{table*}

\clearpage
\makeatletter
\global\@colht\textheight
\global\@colroom\textheight
\global\vsize\textheight
\makeatother

\begin{table*}[!t]
\makeatletter
\@makecaption{\tablename~\ref{tab:verify-concurrency-devices} (continued)}{\normalfont
Output throughput ratios and mean accepted lengths on an H100 (80\,GB) for Q3.5-4B/9B and IVL3.5-4B/8B.
Metrics and aggregation follow Table~\ref{tab:verify-concurrency}.}
\makeatother
\centering
\input{tables/table11_h100}
\end{table*}

Uni-LaSpec adapts its structure to the phase of the trace. It verifies trees
during light load, where it attains the lowest latency of all strategies,
3.8\,s in the first low-load phase, and shortens its linear drafts as the
burst builds. Over the whole trace, its mean arrival-to-completion latency is
14.0\,s, against 26.74\,s for Linear, 33.09\,s for Tree $B{=}15$, and
84.13\,s for Tree $B{=}31$, reductions of 47.6\% and 57.7--83.4\%,
respectively. The difference originates in the burst. Fixed trees keep a high
acceptance but spend so much verifier width per request that the queue grows,
and the backlog then compounds during the spike. During the burst, the mean
latency of Uni-LaSpec is 18.1\,s, compared with 31.3\,s for Linear and
34.3--60.7\,s for the trees, and over the trace its time per output token is
38.9\,ms, against 72.0\,ms for Linear and 90.6\,ms for Tree $B{=}15$.
Repeated runs of the same trace confirm the ordering against the strongest
short-draft baselines: Uni-LaSpec averages 13.79\,s over four runs, compared
with 14.64\,s for Trunc.\ $N{=}5$ and 15.25\,s for dynamic truncation,
reductions of 5.8\% and 9.6\%. This experiment uses the threshold profile
$0.25/0.45/0.6/0.6$ at verifier widths $64/128/256/512$. Load-adaptive
verification thus combines the low-load latency of branching with the
high-load efficiency of truncation in a single serving policy.

\subsubsection{Generalization across GPUs}
Table~\ref{tab:verify-concurrency-devices} repeats the comparison on a
consumer RTX 5090 (32\,GB) and data-center A800 and H100 GPUs (80\,GB), each
with its own same-server AR baseline, and transfers the knee and threshold
profile without recalibration. Uni-LaSpec improves on AR in all 120
settings across these three devices, with 1.10--3.70$\times$ on the RTX 5090,
1.06--4.39$\times$ on the A800, and 1.21--4.47$\times$ on the H100.
It matches or exceeds the best fixed strategy in 96 of these
120 settings: 17 of 24 on the RTX 5090, 45 of 48 on the A800, and
34 of 48 on the H100. On the A800, Q3.5-4B, Q3.5-9B, and
IVL3.5-8B attain the highest throughput at every load in both modalities.
Fixed trees fall below AR in 6--12 of the 24 RTX 5090 settings, 14--29 of the
48 A800 settings, and 12--26 of the 48 H100 settings.
Uni-LaSpec raises throughput over linear verification of the same drafts by
18\% on average on the RTX 5090 and A800, and by 11\% on the H100.
The policy therefore transfers across memory capacity and compute class
without per-device tuning, although fixed short drafts retain an advantage
at some individual operating points.

\section{Conclusion}
\label{sec:conclusion}

Uni-LaSpec couples visual evidence selection with load-adaptive verification
for block-parallel VLM drafting. The visual branch reuses cached verifier
features to supply a small set of relevant image tokens, while the scheduler
allocates verifier positions between draft trees and dynamically truncated
linear drafts. Both paths optimize accepted output per unit of verification
cost, with the verifier retaining control over committed tokens.

Across six VLMs and 17 benchmarks, Uni-LaSpec raises acceptance over the
same block drafter without direct visual input on 198 of 204 benchmark
entries, and selecting eight visual tokens retains nearly all of the accepted
length obtained from the complete image while keeping selection below one
millisecond. The verification study shows that acceptance and speed must be
optimized together: larger trees accept more tokens but lose throughput under
load, whereas truncation pays off as concurrency rises. By choosing the
structure per request and per round, the load-adaptive policy improves on
autoregressive decoding in every evaluated setting and matches or exceeds the
best fixed strategy in 157 of the 192 evaluated model--modality--load
settings. Under bursty
arrivals, it reduces mean end-to-end latency by 47.6\% relative to linear
verification.

\bibliographystyle{IEEEtran}
\bibliography{references}

\end{document}

%% file: tables/table1.tex
\footnotesize
\setlength{\tabcolsep}{1.4pt}
\renewcommand{\arraystretch}{1.355}
\setlength{\aboverulesep}{0.2ex}
\setlength{\belowrulesep}{0.3ex}
\begin{tabular}{lllccccccccccccccccc}
\toprule
 &  &  & \multicolumn{7}{c}{Text Benchmarks} & \multicolumn{10}{c}{Multimodal Benchmarks} \\
\cmidrule(lr){4-10} \cmidrule(lr){11-20}
Samp. & Model & Method & \multicolumn{3}{c}{Math} & \multicolumn{3}{c}{Code} & \multicolumn{1}{c}{Chat} & \multicolumn{4}{c}{Document \& OCR} & \multicolumn{4}{c}{General Vision} & \multicolumn{2}{c}{Capt.\ \& Chat} \\
\cmidrule(lr){4-6} \cmidrule(lr){7-9} \cmidrule(lr){10-10} \cmidrule(lr){11-14} \cmidrule(lr){15-18} \cmidrule(lr){19-20}
 &  &  & GSM8K & MATH & AIME & HEval & MBPP & LCB & MT-B & ODoc & OCR2 & DVQA & olmO & CXiv & TVQA & MMMU & GQA & COCO & MMMT \\
\midrule
\multirow{22}{*}{$T{=}0$} & \multirow{5}{*}{Q3.5-4B} & EAGLE-3 & 3.53 & 4.25 & 3.91 & 4.49 & 4.25 & 3.34 & 2.84 & 2.07 & 2.38 & 2.61 & 2.49 & 2.74 & 2.32 & 2.30 & 2.18 & 2.01 & 2.37 \\
 &  & \cellcolor{gray!15}\textcolor{gray!70}{MTP} & \cellcolor{gray!15}\textcolor{gray!70}{6.11} & \cellcolor{gray!15}\textcolor{gray!70}{6.29} & \cellcolor{gray!15}\textcolor{gray!70}{5.66} & \cellcolor{gray!15}\textcolor{gray!70}{6.13} & \cellcolor{gray!15}\textcolor{gray!70}{5.98} & \cellcolor{gray!15}\textcolor{gray!70}{4.73} & \cellcolor{gray!15}\textcolor{gray!70}{4.58} & \cellcolor{gray!15}\textcolor{gray!70}{4.14} & \cellcolor{gray!15}\textcolor{gray!70}{4.02} & \cellcolor{gray!15}\textcolor{gray!70}{5.25} & \cellcolor{gray!15}\textcolor{gray!70}{4.48} & \cellcolor{gray!15}\textcolor{gray!70}{4.65} & \cellcolor{gray!15}\textcolor{gray!70}{4.18} & \cellcolor{gray!15}\textcolor{gray!70}{3.98} & \cellcolor{gray!15}\textcolor{gray!70}{3.72} & \cellcolor{gray!15}\textcolor{gray!70}{3.54} & \cellcolor{gray!15}\textcolor{gray!70}{4.13} \\
 &  & DFlash (T) & 4.92 & 6.41 & 6.52 & 5.27 & 4.99 & 4.48 & 3.48 & 2.81 & 3.24 & 3.40 & 4.29 & 4.66 & 3.09 & 4.01 & 3.00 & 2.66 & 3.74 \\
 &  & DFlash (M) & 5.07 & 6.53 & 6.57 & 5.36 & 5.01 & 4.55 & 3.60 & 3.01 & 3.51 & 3.83 & 4.71 & 4.92 & 3.47 & 4.11 & 3.34 & 3.07 & 4.00 \\
 &  & \cellcolor{orange!15}\textbf{Uni-LaSpec} & \cellcolor{orange!15}\textbf{5.35} & \cellcolor{orange!15}\textbf{6.67} & \cellcolor{orange!15}\textbf{6.69} & \cellcolor{orange!15}\textbf{5.46} & \cellcolor{orange!15}\textbf{5.04} & \cellcolor{orange!15}\textbf{4.78} & \cellcolor{orange!15}\textbf{3.81} & \cellcolor{orange!15}\textbf{3.05} & \cellcolor{orange!15}\textbf{3.78} & \cellcolor{orange!15}\textbf{4.14} & \cellcolor{orange!15}\textbf{4.86} & \cellcolor{orange!15}\textbf{5.20} & \cellcolor{orange!15}\textbf{3.71} & \cellcolor{orange!15}\textbf{4.33} & \cellcolor{orange!15}\textbf{3.66} & \cellcolor{orange!15}\textbf{3.56} & \cellcolor{orange!15}\textbf{4.36} \\
\cmidrule(l){2-20}
 & \multirow{5}{*}{Q3.5-9B} & EAGLE-3 & 3.85 & 4.61 & 4.49 & 4.69 & 4.57 & 3.57 & 2.97 & 2.22 & 2.59 & 2.77 & 2.85 & 3.09 & 2.48 & 2.62 & 2.37 & 2.26 & 2.69 \\
 &  & \cellcolor{gray!15}\textcolor{gray!70}{MTP} & \cellcolor{gray!15}\textcolor{gray!70}{6.19} & \cellcolor{gray!15}\textcolor{gray!70}{6.61} & \cellcolor{gray!15}\textcolor{gray!70}{5.93} & \cellcolor{gray!15}\textcolor{gray!70}{6.32} & \cellcolor{gray!15}\textcolor{gray!70}{6.06} & \cellcolor{gray!15}\textcolor{gray!70}{5.07} & \cellcolor{gray!15}\textcolor{gray!70}{4.69} & \cellcolor{gray!15}\textcolor{gray!70}{4.29} & \cellcolor{gray!15}\textcolor{gray!70}{4.26} & \cellcolor{gray!15}\textcolor{gray!70}{5.60} & \cellcolor{gray!15}\textcolor{gray!70}{4.67} & \cellcolor{gray!15}\textcolor{gray!70}{4.93} & \cellcolor{gray!15}\textcolor{gray!70}{4.27} & \cellcolor{gray!15}\textcolor{gray!70}{4.10} & \cellcolor{gray!15}\textcolor{gray!70}{3.76} & \cellcolor{gray!15}\textcolor{gray!70}{3.60} & \cellcolor{gray!15}\textcolor{gray!70}{4.28} \\
 &  & DFlash (T) & 4.83 & 6.23 & 6.23 & 5.20 & 4.87 & 4.60 & 3.36 & 2.67 & 3.28 & 3.41 & 4.29 & 4.58 & 2.98 & 3.85 & 2.92 & 2.63 & 3.67 \\
 &  & DFlash (M) & 5.00 & 6.26 & 6.20 & 5.22 & \textbf{5.04} & 4.54 & 3.47 & 2.79 & 3.48 & 3.83 & 4.63 & 4.80 & 3.25 & 3.99 & 3.21 & 3.04 & 3.94 \\
 &  & \cellcolor{orange!15}\textbf{Uni-LaSpec} & \cellcolor{orange!15}\textbf{5.22} & \cellcolor{orange!15}\textbf{6.42} & \cellcolor{orange!15}\textbf{6.35} & \cellcolor{orange!15}\textbf{5.29} & \cellcolor{orange!15}5.02 & \cellcolor{orange!15}\textbf{4.65} & \cellcolor{orange!15}\textbf{3.71} & \cellcolor{orange!15}\textbf{2.88} & \cellcolor{orange!15}\textbf{3.78} & \cellcolor{orange!15}\textbf{4.02} & \cellcolor{orange!15}\textbf{4.78} & \cellcolor{orange!15}\textbf{4.99} & \cellcolor{orange!15}\textbf{3.47} & \cellcolor{orange!15}\textbf{4.18} & \cellcolor{orange!15}\textbf{3.54} & \cellcolor{orange!15}\textbf{3.45} & \cellcolor{orange!15}\textbf{4.21} \\
\cmidrule(l){2-20}
 & \multirow{3}{*}{IVL3.5-4B} & DFlash (T) & 5.08 & 7.07 & 4.67 & 3.61 & 4.09 & 3.69 & 3.00 & 2.81 & 2.61 & 2.58 & 4.21 & 4.02 & 2.00 & 4.88 & 2.23 & 1.88 & 3.07 \\
 &  & DFlash (M) & 5.17 & 7.04 & 4.65 & 3.60 & 4.04 & 3.63 & 3.09 & 2.94 & 2.92 & 3.15 & 4.59 & 4.29 & 2.35 & 5.03 & 2.82 & 2.48 & 3.30 \\
 &  & \cellcolor{orange!15}\textbf{Uni-LaSpec} & \cellcolor{orange!15}\textbf{5.88} & \cellcolor{orange!15}\textbf{7.22} & \cellcolor{orange!15}\textbf{4.83} & \cellcolor{orange!15}\textbf{3.79} & \cellcolor{orange!15}\textbf{4.20} & \cellcolor{orange!15}\textbf{3.79} & \cellcolor{orange!15}\textbf{3.46} & \cellcolor{orange!15}\textbf{3.03} & \cellcolor{orange!15}\textbf{3.12} & \cellcolor{orange!15}\textbf{3.56} & \cellcolor{orange!15}\textbf{4.68} & \cellcolor{orange!15}\textbf{4.64} & \cellcolor{orange!15}\textbf{2.70} & \cellcolor{orange!15}\textbf{5.34} & \cellcolor{orange!15}\textbf{3.33} & \cellcolor{orange!15}\textbf{2.98} & \cellcolor{orange!15}\textbf{3.53} \\
\cmidrule(l){2-20}
 & \multirow{3}{*}{IVL3.5-8B} & DFlash (T) & 6.24 & 5.90 & 5.21 & 5.30 & 4.88 & 5.04 & 2.74 & 3.17 & 2.74 & 2.66 & 3.89 & 4.37 & 2.00 & 4.36 & 2.29 & 1.89 & 3.18 \\
 &  & DFlash (M) & 6.53 & 5.91 & 5.17 & 5.25 & 4.88 & 4.95 & 2.89 & 3.27 & 3.06 & 3.15 & 3.93 & 4.64 & 2.39 & 4.50 & 2.92 & 2.41 & 3.42 \\
 &  & \cellcolor{orange!15}\textbf{Uni-LaSpec} & \cellcolor{orange!15}\textbf{7.14} & \cellcolor{orange!15}\textbf{6.21} & \cellcolor{orange!15}\textbf{5.59} & \cellcolor{orange!15}\textbf{5.45} & \cellcolor{orange!15}\textbf{5.13} & \cellcolor{orange!15}\textbf{5.25} & \cellcolor{orange!15}\textbf{3.12} & \cellcolor{orange!15}\textbf{3.36} & \cellcolor{orange!15}\textbf{3.33} & \cellcolor{orange!15}\textbf{3.55} & \cellcolor{orange!15}\textbf{4.21} & \cellcolor{orange!15}\textbf{4.93} & \cellcolor{orange!15}\textbf{2.69} & \cellcolor{orange!15}\textbf{4.65} & \cellcolor{orange!15}\textbf{3.40} & \cellcolor{orange!15}\textbf{2.93} & \cellcolor{orange!15}\textbf{3.74} \\
\cmidrule(l){2-20}
 & \multirow{3}{*}{G4-E2B} & DFlash (T) & 5.01 & 6.67 & 6.29 & \textbf{5.86} & 5.11 & 4.68 & 2.63 & 2.03 & 2.58 & 2.35 & 2.41 & 3.60 & 1.96 & 3.46 & 2.01 & 1.82 & 2.62 \\
 &  & DFlash (M) & 5.46 & 6.52 & \textbf{6.35} & 5.64 & 5.04 & 4.68 & 2.82 & \textbf{2.38} & 3.20 & 3.09 & 2.84 & 4.03 & 2.59 & 3.56 & 2.63 & 2.51 & 2.90 \\
 &  & \cellcolor{orange!15}\textbf{Uni-LaSpec} & \cellcolor{orange!15}\textbf{5.95} & \cellcolor{orange!15}\textbf{6.70} & \cellcolor{orange!15}6.23 & \cellcolor{orange!15}5.72 & \cellcolor{orange!15}\textbf{5.14} & \cellcolor{orange!15}\textbf{4.73} & \cellcolor{orange!15}\textbf{2.97} & \cellcolor{orange!15}2.37 & \cellcolor{orange!15}\textbf{3.58} & \cellcolor{orange!15}\textbf{3.57} & \cellcolor{orange!15}\textbf{3.06} & \cellcolor{orange!15}\textbf{4.34} & \cellcolor{orange!15}\textbf{2.83} & \cellcolor{orange!15}\textbf{3.78} & \cellcolor{orange!15}\textbf{3.07} & \cellcolor{orange!15}\textbf{2.96} & \cellcolor{orange!15}\textbf{3.16} \\
\cmidrule(l){2-20}
 & \multirow{3}{*}{G4-E4B} & DFlash (T) & 5.17 & 6.88 & 6.85 & \textbf{6.37} & 5.53 & 5.00 & 2.77 & 2.13 & 2.69 & 2.50 & 2.81 & 3.79 & 2.15 & 3.93 & 2.15 & 1.87 & 2.75 \\
 &  & DFlash (M) & 5.66 & 7.02 & 6.96 & 6.31 & 5.36 & 4.98 & 2.89 & 2.39 & 3.19 & 3.15 & 3.25 & 4.23 & 2.63 & 4.03 & 2.71 & 2.44 & 3.04 \\
 &  & \cellcolor{orange!15}\textbf{Uni-LaSpec} & \cellcolor{orange!15}\textbf{6.24} & \cellcolor{orange!15}\textbf{7.12} & \cellcolor{orange!15}\textbf{7.10} & \cellcolor{orange!15}6.37 & \cellcolor{orange!15}\textbf{5.58} & \cellcolor{orange!15}\textbf{5.12} & \cellcolor{orange!15}\textbf{3.13} & \cellcolor{orange!15}\textbf{2.48} & \cellcolor{orange!15}\textbf{3.45} & \cellcolor{orange!15}\textbf{3.49} & \cellcolor{orange!15}\textbf{3.42} & \cellcolor{orange!15}\textbf{4.48} & \cellcolor{orange!15}\textbf{3.00} & \cellcolor{orange!15}\textbf{4.19} & \cellcolor{orange!15}\textbf{3.25} & \cellcolor{orange!15}\textbf{2.95} & \cellcolor{orange!15}\textbf{3.26} \\
\midrule
\multirow{22}{*}{$T{=}1$} & \multirow{5}{*}{Q3.5-4B} & EAGLE-3 & 3.11 & 3.60 & 3.22 & 3.68 & 3.38 & 2.77 & 2.37 & 1.95 & 2.16 & 2.45 & 2.29 & 2.40 & 2.04 & 2.03 & 1.96 & 1.83 & 2.16 \\
 &  & \cellcolor{gray!15}\textcolor{gray!70}{MTP} & \cellcolor{gray!15}\textcolor{gray!70}{4.59} & \cellcolor{gray!15}\textcolor{gray!70}{4.88} & \cellcolor{gray!15}\textcolor{gray!70}{4.38} & \cellcolor{gray!15}\textcolor{gray!70}{4.55} & \cellcolor{gray!15}\textcolor{gray!70}{4.28} & \cellcolor{gray!15}\textcolor{gray!70}{3.62} & \cellcolor{gray!15}\textcolor{gray!70}{3.41} & \cellcolor{gray!15}\textcolor{gray!70}{3.59} & \cellcolor{gray!15}\textcolor{gray!70}{3.42} & \cellcolor{gray!15}\textcolor{gray!70}{4.39} & \cellcolor{gray!15}\textcolor{gray!70}{3.84} & \cellcolor{gray!15}\textcolor{gray!70}{3.88} & \cellcolor{gray!15}\textcolor{gray!70}{3.24} & \cellcolor{gray!15}\textcolor{gray!70}{3.33} & \cellcolor{gray!15}\textcolor{gray!70}{3.03} & \cellcolor{gray!15}\textcolor{gray!70}{2.91} & \cellcolor{gray!15}\textcolor{gray!70}{3.44} \\
 &  & DFlash (T) & 4.08 & 4.91 & \textbf{4.58} & 4.18 & 3.81 & 3.41 & 2.85 & 2.55 & 2.86 & 3.14 & 3.66 & 3.80 & 2.63 & 3.18 & 2.59 & 2.31 & 3.15 \\
 &  & DFlash (M) & 4.07 & 4.97 & 4.46 & 4.28 & 3.87 & 3.48 & 2.85 & 2.65 & 3.08 & 3.45 & 3.90 & 3.91 & 2.78 & 3.28 & 2.82 & 2.57 & 3.30 \\
 &  & \cellcolor{orange!15}\textbf{Uni-LaSpec} & \cellcolor{orange!15}\textbf{4.32} & \cellcolor{orange!15}\textbf{5.02} & \cellcolor{orange!15}4.51 & \cellcolor{orange!15}\textbf{4.29} & \cellcolor{orange!15}\textbf{4.05} & \cellcolor{orange!15}\textbf{3.49} & \cellcolor{orange!15}\textbf{2.95} & \cellcolor{orange!15}\textbf{2.75} & \cellcolor{orange!15}\textbf{3.29} & \cellcolor{orange!15}\textbf{3.69} & \cellcolor{orange!15}\textbf{4.09} & \cellcolor{orange!15}\textbf{4.15} & \cellcolor{orange!15}\textbf{2.96} & \cellcolor{orange!15}\textbf{3.43} & \cellcolor{orange!15}\textbf{3.03} & \cellcolor{orange!15}\textbf{2.88} & \cellcolor{orange!15}\textbf{3.52} \\
\cmidrule(l){2-20}
 & \multirow{5}{*}{Q3.5-9B} & EAGLE-3 & 3.37 & 3.98 & 3.63 & 3.97 & 3.82 & 2.98 & 2.58 & 2.11 & 2.40 & 2.61 & 2.67 & 2.77 & 2.28 & 2.38 & 2.19 & 2.06 & 2.46 \\
 &  & \cellcolor{gray!15}\textcolor{gray!70}{MTP} & \cellcolor{gray!15}\textcolor{gray!70}{4.95} & \cellcolor{gray!15}\textcolor{gray!70}{5.20} & \cellcolor{gray!15}\textcolor{gray!70}{4.69} & \cellcolor{gray!15}\textcolor{gray!70}{5.05} & \cellcolor{gray!15}\textcolor{gray!70}{4.85} & \cellcolor{gray!15}\textcolor{gray!70}{3.85} & \cellcolor{gray!15}\textcolor{gray!70}{3.68} & \cellcolor{gray!15}\textcolor{gray!70}{3.83} & \cellcolor{gray!15}\textcolor{gray!70}{3.59} & \cellcolor{gray!15}\textcolor{gray!70}{4.63} & \cellcolor{gray!15}\textcolor{gray!70}{4.12} & \cellcolor{gray!15}\textcolor{gray!70}{4.12} & \cellcolor{gray!15}\textcolor{gray!70}{3.43} & \cellcolor{gray!15}\textcolor{gray!70}{3.47} & \cellcolor{gray!15}\textcolor{gray!70}{3.21} & \cellcolor{gray!15}\textcolor{gray!70}{3.08} & \cellcolor{gray!15}\textcolor{gray!70}{3.66} \\
 &  & DFlash (T) & 4.03 & 4.88 & 4.53 & \textbf{4.29} & 3.98 & 3.51 & 2.83 & 2.50 & 2.95 & 3.14 & 3.70 & 3.80 & 2.62 & 3.26 & 2.66 & 2.34 & 3.16 \\
 &  & DFlash (M) & 4.00 & 4.96 & 4.51 & 4.26 & 3.98 & 3.44 & 2.94 & 2.62 & 3.10 & 3.58 & 4.02 & 3.98 & 2.77 & 3.38 & 2.81 & 2.66 & 3.48 \\
 &  & \cellcolor{orange!15}\textbf{Uni-LaSpec} & \cellcolor{orange!15}\textbf{4.32} & \cellcolor{orange!15}\textbf{5.08} & \cellcolor{orange!15}\textbf{4.69} & \cellcolor{orange!15}4.27 & \cellcolor{orange!15}\textbf{4.14} & \cellcolor{orange!15}\textbf{3.55} & \cellcolor{orange!15}\textbf{3.08} & \cellcolor{orange!15}\textbf{2.66} & \cellcolor{orange!15}\textbf{3.36} & \cellcolor{orange!15}\textbf{3.68} & \cellcolor{orange!15}\textbf{4.10} & \cellcolor{orange!15}\textbf{4.04} & \cellcolor{orange!15}\textbf{2.97} & \cellcolor{orange!15}\textbf{3.51} & \cellcolor{orange!15}\textbf{3.08} & \cellcolor{orange!15}\textbf{2.94} & \cellcolor{orange!15}\textbf{3.60} \\
\cmidrule(l){2-20}
 & \multirow{3}{*}{IVL3.5-4B} & DFlash (T) & 4.00 & 4.75 & 3.59 & 2.89 & 3.24 & 2.81 & 2.58 & \textbf{1.85} & 2.49 & 2.27 & 2.49 & 3.73 & 1.93 & 3.66 & 1.95 & 1.73 & 2.53 \\
 &  & DFlash (M) & 4.09 & 4.83 & 3.67 & 2.86 & 3.21 & 2.82 & 2.59 & 1.77 & 2.71 & 2.51 & 2.51 & 3.55 & 2.02 & 3.62 & 2.32 & 2.05 & 2.85 \\
 &  & \cellcolor{orange!15}\textbf{Uni-LaSpec} & \cellcolor{orange!15}\textbf{4.62} & \cellcolor{orange!15}\textbf{5.22} & \cellcolor{orange!15}\textbf{3.75} & \cellcolor{orange!15}\textbf{2.95} & \cellcolor{orange!15}\textbf{3.36} & \cellcolor{orange!15}\textbf{2.89} & \cellcolor{orange!15}\textbf{2.85} & \cellcolor{orange!15}1.80 & \cellcolor{orange!15}\textbf{2.98} & \cellcolor{orange!15}\textbf{2.80} & \cellcolor{orange!15}\textbf{2.60} & \cellcolor{orange!15}\textbf{3.87} & \cellcolor{orange!15}\textbf{2.28} & \cellcolor{orange!15}\textbf{3.97} & \cellcolor{orange!15}\textbf{2.53} & \cellcolor{orange!15}\textbf{2.29} & \cellcolor{orange!15}\textbf{3.01} \\
\cmidrule(l){2-20}
 & \multirow{3}{*}{IVL3.5-8B} & DFlash (T) & 5.07 & 4.75 & 4.15 & 4.10 & 3.65 & 3.25 & 2.40 & 1.71 & 2.18 & 2.53 & 2.38 & 3.56 & 1.97 & \textbf{3.66} & 2.03 & 1.75 & 2.83 \\
 &  & DFlash (M) & 5.11 & 4.63 & 4.26 & 3.97 & 3.70 & 3.21 & 2.47 & 1.77 & \textbf{2.70} & 2.69 & 2.33 & 3.84 & 2.18 & 3.44 & 2.40 & 2.07 & 2.89 \\
 &  & \cellcolor{orange!15}\textbf{Uni-LaSpec} & \cellcolor{orange!15}\textbf{5.52} & \cellcolor{orange!15}\textbf{4.96} & \cellcolor{orange!15}\textbf{4.50} & \cellcolor{orange!15}\textbf{4.16} & \cellcolor{orange!15}\textbf{3.83} & \cellcolor{orange!15}\textbf{3.27} & \cellcolor{orange!15}\textbf{2.58} & \cellcolor{orange!15}\textbf{1.84} & \cellcolor{orange!15}2.61 & \cellcolor{orange!15}\textbf{3.20} & \cellcolor{orange!15}\textbf{2.47} & \cellcolor{orange!15}\textbf{4.04} & \cellcolor{orange!15}\textbf{2.39} & \cellcolor{orange!15}3.64 & \cellcolor{orange!15}\textbf{2.71} & \cellcolor{orange!15}\textbf{2.31} & \cellcolor{orange!15}\textbf{3.39} \\
\cmidrule(l){2-20}
 & \multirow{3}{*}{G4-E2B} & DFlash (T) & 4.86 & 6.09 & 5.68 & \textbf{5.73} & 4.84 & 4.26 & 2.60 & 1.85 & 2.30 & 2.30 & 2.36 & 3.36 & 1.96 & 3.19 & 1.98 & 1.78 & 2.48 \\
 &  & DFlash (M) & 5.28 & 6.05 & 5.66 & 5.66 & 4.86 & 4.24 & 2.72 & \textbf{2.17} & 2.76 & 3.02 & 2.78 & 3.74 & 2.49 & 3.26 & 2.60 & 2.41 & 2.81 \\
 &  & \cellcolor{orange!15}\textbf{Uni-LaSpec} & \cellcolor{orange!15}\textbf{5.56} & \cellcolor{orange!15}\textbf{6.22} & \cellcolor{orange!15}\textbf{5.76} & \cellcolor{orange!15}5.56 & \cellcolor{orange!15}\textbf{4.96} & \cellcolor{orange!15}\textbf{4.37} & \cellcolor{orange!15}\textbf{2.81} & \cellcolor{orange!15}2.15 & \cellcolor{orange!15}\textbf{3.00} & \cellcolor{orange!15}\textbf{3.24} & \cellcolor{orange!15}\textbf{3.01} & \cellcolor{orange!15}\textbf{3.95} & \cellcolor{orange!15}\textbf{2.66} & \cellcolor{orange!15}\textbf{3.42} & \cellcolor{orange!15}\textbf{2.90} & \cellcolor{orange!15}\textbf{2.74} & \cellcolor{orange!15}\textbf{2.93} \\
\cmidrule(l){2-20}
 & \multirow{3}{*}{G4-E4B} & DFlash (T) & 4.86 & 6.39 & 5.79 & 5.87 & 4.99 & 4.26 & 2.66 & 1.97 & 2.43 & 2.45 & 2.65 & 3.47 & 2.04 & 3.36 & 2.07 & 1.81 & 2.54 \\
 &  & DFlash (M) & 5.29 & 6.25 & 6.05 & 5.79 & 4.92 & 4.33 & 2.74 & 2.19 & 2.94 & 3.09 & 3.02 & 3.66 & 2.38 & 3.53 & 2.47 & 2.28 & 2.77 \\
 &  & \cellcolor{orange!15}\textbf{Uni-LaSpec} & \cellcolor{orange!15}\textbf{5.73} & \cellcolor{orange!15}\textbf{6.62} & \cellcolor{orange!15}\textbf{6.10} & \cellcolor{orange!15}\textbf{5.94} & \cellcolor{orange!15}\textbf{5.09} & \cellcolor{orange!15}\textbf{4.49} & \cellcolor{orange!15}\textbf{2.92} & \cellcolor{orange!15}\textbf{2.37} & \cellcolor{orange!15}\textbf{3.20} & \cellcolor{orange!15}\textbf{3.30} & \cellcolor{orange!15}\textbf{3.24} & \cellcolor{orange!15}\textbf{4.09} & \cellcolor{orange!15}\textbf{2.65} & \cellcolor{orange!15}\textbf{3.58} & \cellcolor{orange!15}\textbf{2.85} & \cellcolor{orange!15}\textbf{2.65} & \cellcolor{orange!15}\textbf{2.95} \\
\bottomrule
\end{tabular}

%% file: tables/table2.tex
\footnotesize
\setlength{\tabcolsep}{1.0pt}
\renewcommand{\arraystretch}{1.395}
\setlength{\aboverulesep}{0.2ex}
\setlength{\belowrulesep}{0.3ex}
\begin{tabular}{lllccccccccccccccccc}
\toprule
 &  &  & \multicolumn{7}{c}{Text Benchmarks} & \multicolumn{10}{c}{Multimodal Benchmarks} \\
\cmidrule(lr){4-10} \cmidrule(lr){11-20}
Samp. & Model & Method & \multicolumn{3}{c}{Math} & \multicolumn{3}{c}{Code} & \multicolumn{1}{c}{Chat} & \multicolumn{4}{c}{Document \& OCR} & \multicolumn{4}{c}{General Vision} & \multicolumn{2}{c}{Capt.\ \& Chat} \\
\cmidrule(lr){4-6} \cmidrule(lr){7-9} \cmidrule(lr){10-10} \cmidrule(lr){11-14} \cmidrule(lr){15-18} \cmidrule(lr){19-20}
 &  &  & GSM8K & MATH & AIME & HEval & MBPP & LCB & MT-B & ODoc & OCR2 & DVQA & olmO & CXiv & TVQA & MMMU & GQA & COCO & MMMT \\
\midrule
\multirow{22}{*}{$T{=}0$} & \multirow{5}{*}{Q3.5-4B} & EAGLE-3 & 1.33$\times$ & 1.58$\times$ & 1.45$\times$ & 1.64$\times$ & 1.59$\times$ & 1.26$\times$ & 1.06$\times$ & 0.79$\times$ & 0.93$\times$ & 0.99$\times$ & 0.94$\times$ & 1.05$\times$ & 0.86$\times$ & 0.88$\times$ & 0.82$\times$ & 0.75$\times$ & 0.92$\times$ \\
 &  & \cellcolor{gray!15}\textcolor{gray!70}{MTP} & \cellcolor{gray!15}\textcolor{gray!70}{1.68$\times$} & \cellcolor{gray!15}\textcolor{gray!70}{1.71$\times$} & \cellcolor{gray!15}\textcolor{gray!70}{1.58$\times$} & \cellcolor{gray!15}\textcolor{gray!70}{1.68$\times$} & \cellcolor{gray!15}\textcolor{gray!70}{1.64$\times$} & \cellcolor{gray!15}\textcolor{gray!70}{1.30$\times$} & \cellcolor{gray!15}\textcolor{gray!70}{1.26$\times$} & \cellcolor{gray!15}\textcolor{gray!70}{1.14$\times$} & \cellcolor{gray!15}\textcolor{gray!70}{1.11$\times$} & \cellcolor{gray!15}\textcolor{gray!70}{1.37$\times$} & \cellcolor{gray!15}\textcolor{gray!70}{1.23$\times$} & \cellcolor{gray!15}\textcolor{gray!70}{1.28$\times$} & \cellcolor{gray!15}\textcolor{gray!70}{1.11$\times$} & \cellcolor{gray!15}\textcolor{gray!70}{1.09$\times$} & \cellcolor{gray!15}\textcolor{gray!70}{1.01$\times$} & \cellcolor{gray!15}\textcolor{gray!70}{0.97$\times$} & \cellcolor{gray!15}\textcolor{gray!70}{1.13$\times$} \\
\bottomrule
\end{tabular}

%% file: tables/table3.tex
\footnotesize
\setlength{\tabcolsep}{2.2pt}
\renewcommand{\arraystretch}{1.307}
\begin{tabular}{lllcccccccccccccc}
\toprule
\multirow{2}{*}[-2.7pt]{Samp.} & \multirow{2}{*}[-2.7pt]{Model} & \multirow{2}{*}[-2.7pt]{Method} & \multicolumn{7}{c}{Text Benchmarks} & \multicolumn{7}{c}{Multimodal Benchmarks} \\
\cmidrule(lr){4-10} \cmidrule(lr){11-17}
 & & & $R{=}1$ & $R{=}4$ & $R{=}8$ & $R{=}16$ & $R{=}32$ & $R{=}64$ & Avg.\ $\tau$ & $R{=}1$ & $R{=}4$ & $R{=}8$ & $R{=}16$ & $R{=}32$ & $R{=}64$ & Avg.\ $\tau$ \\
\midrule
\multirow{22}{*}{$T{=}0$} & \multirow{5}{*}{Q3.5-4B} & EAGLE-3 & 1.41$\times$ & 1.54$\times$ & 1.53$\times$ & 1.33$\times$ & 0.95$\times$ & 0.76$\times$ & 3.82 & 0.89$\times$ & 0.95$\times$ & 0.98$\times$ & 0.84$\times$ & 0.63$\times$ & 0.51$\times$ & 2.34 \\
 &  & \cellcolor{gray!15}\textcolor{gray!70}{MTP} & \cellcolor{gray!15}\textcolor{gray!70}{1.55$\times$} & \cellcolor{gray!15}\textcolor{gray!70}{1.59$\times$} & \cellcolor{gray!15}\textcolor{gray!70}{1.55$\times$} & \cellcolor{gray!15}\textcolor{gray!70}{1.34$\times$} & \cellcolor{gray!15}\textcolor{gray!70}{1.08$\times$} & \cellcolor{gray!15}\textcolor{gray!70}{0.90$\times$} & \cellcolor{gray!15}\textcolor{gray!70}{5.61} & \cellcolor{gray!15}\textcolor{gray!70}{1.14$\times$} & \cellcolor{gray!15}\textcolor{gray!70}{1.18$\times$} & \cellcolor{gray!15}\textcolor{gray!70}{1.12$\times$} & \cellcolor{gray!15}\textcolor{gray!70}{1.00$\times$} & \cellcolor{gray!15}\textcolor{gray!70}{0.83$\times$} & \cellcolor{gray!15}\textcolor{gray!70}{0.71$\times$} & \cellcolor{gray!15}\textcolor{gray!70}{4.20} \\
 &  & DFlash (T) & 3.41$\times$ & 3.25$\times$ & 2.86$\times$ & 1.97$\times$ & 1.32$\times$ & 1.00$\times$ & 5.16 & 2.19$\times$ & 2.15$\times$ & 1.88$\times$ & 1.31$\times$ & 0.94$\times$ & 0.72$\times$ & 3.49 \\
 &  & DFlash (M) & 3.40$\times$ & 3.28$\times$ & 2.92$\times$ & 2.04$\times$ & 1.38$\times$ & 1.03$\times$ & 5.23 & 2.39$\times$ & 2.32$\times$ & 2.03$\times$ & 1.44$\times$ & 1.03$\times$ & 0.78$\times$ & 3.79 \\
 &  & \cellcolor{orange!15}\textbf{Uni-LaSpec} & \cellcolor{orange!15}\textbf{3.60}$\times$ & \cellcolor{orange!15}\textbf{3.42}$\times$ & \cellcolor{orange!15}\textbf{2.95}$\times$ & \cellcolor{orange!15}\textbf{2.11}$\times$ & \cellcolor{orange!15}\textbf{1.41}$\times$ & \cellcolor{orange!15}\textbf{1.06}$\times$ & \cellcolor{orange!15}\textbf{5.38} & \cellcolor{orange!15}\textbf{2.60}$\times$ & \cellcolor{orange!15}\textbf{2.48}$\times$ & \cellcolor{orange!15}\textbf{2.12}$\times$ & \cellcolor{orange!15}\textbf{1.49}$\times$ & \cellcolor{orange!15}\textbf{1.08}$\times$ & \cellcolor{orange!15}\textbf{0.82}$\times$ & \cellcolor{orange!15}\textbf{4.07} \\
\cmidrule(l){2-17}
 & \multirow{5}{*}{Q3.5-9B} & EAGLE-3 & 2.07$\times$ & 1.92$\times$ & 1.74$\times$ & 1.31$\times$ & 1.02$\times$ & 0.79$\times$ & 4.09 & 1.30$\times$ & 1.20$\times$ & 1.10$\times$ & 0.86$\times$ & 0.69$\times$ & 0.55$\times$ & 2.60 \\
 &  & \cellcolor{gray!15}\textcolor{gray!70}{MTP} & \cellcolor{gray!15}\textcolor{gray!70}{1.57$\times$} & \cellcolor{gray!15}\textcolor{gray!70}{1.68$\times$} & \cellcolor{gray!15}\textcolor{gray!70}{1.59$\times$} & \cellcolor{gray!15}\textcolor{gray!70}{1.32$\times$} & \cellcolor{gray!15}\textcolor{gray!70}{1.10$\times$} & \cellcolor{gray!15}\textcolor{gray!70}{0.91$\times$} & \cellcolor{gray!15}\textcolor{gray!70}{5.87} & \cellcolor{gray!15}\textcolor{gray!70}{1.27$\times$} & \cellcolor{gray!15}\textcolor{gray!70}{1.21$\times$} & \cellcolor{gray!15}\textcolor{gray!70}{1.15$\times$} & \cellcolor{gray!15}\textcolor{gray!70}{0.98$\times$} & \cellcolor{gray!15}\textcolor{gray!70}{0.84$\times$} & \cellcolor{gray!15}\textcolor{gray!70}{0.69$\times$} & \cellcolor{gray!15}\textcolor{gray!70}{4.35} \\
 &  & DFlash (T) & 3.82$\times$ & 3.28$\times$ & 2.78$\times$ & 1.83$\times$ & 1.31$\times$ & 0.98$\times$ & 5.03 & 2.56$\times$ & 2.14$\times$ & 1.83$\times$ & 1.26$\times$ & 0.94$\times$ & 0.69$\times$ & 3.42 \\
 &  & DFlash (M) & 3.85$\times$ & 3.31$\times$ & 2.86$\times$ & 1.87$\times$ & 1.35$\times$ & 0.99$\times$ & 5.10 & 2.74$\times$ & 2.29$\times$ & 1.96$\times$ & 1.34$\times$ & 1.00$\times$ & 0.74$\times$ & 3.69 \\
 &  & \cellcolor{orange!15}\textbf{Uni-LaSpec} & \cellcolor{orange!15}\textbf{3.95}$\times$ & \cellcolor{orange!15}\textbf{3.43}$\times$ & \cellcolor{orange!15}\textbf{2.93}$\times$ & \cellcolor{orange!15}\textbf{1.91}$\times$ & \cellcolor{orange!15}\textbf{1.38}$\times$ & \cellcolor{orange!15}\textbf{1.02}$\times$ & \cellcolor{orange!15}\textbf{5.24} & \cellcolor{orange!15}\textbf{2.88}$\times$ & \cellcolor{orange!15}\textbf{2.41}$\times$ & \cellcolor{orange!15}\textbf{2.04}$\times$ & \cellcolor{orange!15}\textbf{1.40}$\times$ & \cellcolor{orange!15}\textbf{1.04}$\times$ & \cellcolor{orange!15}\textbf{0.77}$\times$ & \cellcolor{orange!15}\textbf{3.93} \\
\cmidrule(l){2-17}
 & \multirow{3}{*}{IVL3.5-4B} & DFlash (T) & 3.41$\times$ & 3.05$\times$ & 2.54$\times$ & 1.60$\times$ & 1.20$\times$ & 0.91$\times$ & 4.47 & 2.08$\times$ & 1.89$\times$ & 1.57$\times$ & 1.17$\times$ & 1.04$\times$ & 0.88$\times$ & 3.06 \\
 &  & DFlash (M) & 3.43$\times$ & 2.99$\times$ & 2.58$\times$ & 1.62$\times$ & 1.23$\times$ & 0.93$\times$ & 4.46 & 2.24$\times$ & 2.02$\times$ & 1.66$\times$ & 1.26$\times$ & 1.09$\times$ & 0.90$\times$ & 3.42 \\
 &  & \cellcolor{orange!15}\textbf{Uni-LaSpec} & \cellcolor{orange!15}\textbf{3.63}$\times$ & \cellcolor{orange!15}\textbf{3.17}$\times$ & \cellcolor{orange!15}\textbf{2.66}$\times$ & \cellcolor{orange!15}\textbf{1.66}$\times$ & \cellcolor{orange!15}\textbf{1.24}$\times$ & \cellcolor{orange!15}\textbf{0.95}$\times$ & \cellcolor{orange!15}\textbf{4.67} & \cellcolor{orange!15}\textbf{2.35}$\times$ & \cellcolor{orange!15}\textbf{2.10}$\times$ & \cellcolor{orange!15}\textbf{1.70}$\times$ & \cellcolor{orange!15}\textbf{1.27}$\times$ & \cellcolor{orange!15}\textbf{1.12}$\times$ & \cellcolor{orange!15}\textbf{0.93}$\times$ & \cellcolor{orange!15}\textbf{3.70} \\
\cmidrule(l){2-17}
 & \multirow{3}{*}{IVL3.5-8B} & DFlash (T) & 3.81$\times$ & 3.54$\times$ & 2.87$\times$ & 1.82$\times$ & 1.24$\times$ & 0.98$\times$ & 5.05 & 2.18$\times$ & 1.94$\times$ & 1.59$\times$ & 1.14$\times$ & 0.94$\times$ & 0.78$\times$ & 3.06 \\
 &  & DFlash (M) & 3.84$\times$ & 3.54$\times$ & 2.84$\times$ & 1.80$\times$ & 1.24$\times$ & 0.96$\times$ & 5.06 & 2.34$\times$ & 2.04$\times$ & 1.65$\times$ & 1.16$\times$ & 0.97$\times$ & 0.80$\times$ & 3.36 \\
 &  & \cellcolor{orange!15}\textbf{Uni-LaSpec} & \cellcolor{orange!15}\textbf{4.08}$\times$ & \cellcolor{orange!15}\textbf{3.74}$\times$ & \cellcolor{orange!15}\textbf{3.10}$\times$ & \cellcolor{orange!15}\textbf{1.96}$\times$ & \cellcolor{orange!15}\textbf{1.32}$\times$ & \cellcolor{orange!15}\textbf{1.04}$\times$ & \cellcolor{orange!15}\textbf{5.38} & \cellcolor{orange!15}\textbf{2.47}$\times$ & \cellcolor{orange!15}\textbf{2.17}$\times$ & \cellcolor{orange!15}\textbf{1.75}$\times$ & \cellcolor{orange!15}\textbf{1.28}$\times$ & \cellcolor{orange!15}\textbf{1.03}$\times$ & \cellcolor{orange!15}\textbf{0.85}$\times$ & \cellcolor{orange!15}\textbf{3.69} \\
\cmidrule(l){2-17}
 & \multirow{3}{*}{G4-E2B} & DFlash (T) & 4.84$\times$ & 3.66$\times$ & 2.76$\times$ & 2.57$\times$ & 1.80$\times$ & 1.03$\times$ & 5.16 & 2.26$\times$ & 2.05$\times$ & 1.61$\times$ & 1.22$\times$ & 0.96$\times$ & 0.77$\times$ & 2.49 \\
 &  & DFlash (M) & 4.88$\times$ & 3.72$\times$ & 2.77$\times$ & 2.67$\times$ & 1.83$\times$ & 1.04$\times$ & 5.21 & 2.60$\times$ & 2.40$\times$ & 1.85$\times$ & 1.39$\times$ & 1.07$\times$ & 0.86$\times$ & 2.98 \\
 &  & \cellcolor{orange!15}\textbf{Uni-LaSpec} & \cellcolor{orange!15}\textbf{5.00}$\times$ & \cellcolor{orange!15}\textbf{3.77}$\times$ & \cellcolor{orange!15}\textbf{2.83}$\times$ & \cellcolor{orange!15}\textbf{2.71}$\times$ & \cellcolor{orange!15}\textbf{1.94}$\times$ & \cellcolor{orange!15}\textbf{1.07}$\times$ & \cellcolor{orange!15}\textbf{5.36} & \cellcolor{orange!15}\textbf{2.70}$\times$ & \cellcolor{orange!15}\textbf{2.41}$\times$ & \cellcolor{orange!15}\textbf{1.90}$\times$ & \cellcolor{orange!15}\textbf{1.41}$\times$ & \cellcolor{orange!15}\textbf{1.11}$\times$ & \cellcolor{orange!15}\textbf{0.89}$\times$ & \cellcolor{orange!15}\textbf{3.26} \\
\cmidrule(l){2-17}
 & \multirow{3}{*}{G4-E4B} & DFlash (T) & 4.43$\times$ & 3.93$\times$ & 3.20$\times$ & 2.01$\times$ & 1.32$\times$ & 0.89$\times$ & 5.53 & 2.05$\times$ & 1.77$\times$ & 1.55$\times$ & 1.13$\times$ & 0.90$\times$ & 0.69$\times$ & 2.68 \\
 &  & DFlash (M) & 4.48$\times$ & 3.99$\times$ & 3.25$\times$ & 2.11$\times$ & 1.30$\times$ & 0.86$\times$ & 5.61 & 2.28$\times$ & 1.98$\times$ & 1.72$\times$ & 1.26$\times$ & 0.99$\times$ & 0.77$\times$ & 3.12 \\
 &  & \cellcolor{orange!15}\textbf{Uni-LaSpec} & \cellcolor{orange!15}\textbf{4.65}$\times$ & \cellcolor{orange!15}\textbf{4.09}$\times$ & \cellcolor{orange!15}\textbf{3.38}$\times$ & \cellcolor{orange!15}\textbf{2.21}$\times$ & \cellcolor{orange!15}\textbf{1.38}$\times$ & \cellcolor{orange!15}\textbf{0.90}$\times$ & \cellcolor{orange!15}\textbf{5.82} & \cellcolor{orange!15}\textbf{2.43}$\times$ & \cellcolor{orange!15}\textbf{2.08}$\times$ & \cellcolor{orange!15}\textbf{1.81}$\times$ & \cellcolor{orange!15}\textbf{1.34}$\times$ & \cellcolor{orange!15}\textbf{1.07}$\times$ & \cellcolor{orange!15}\textbf{0.83}$\times$ & \cellcolor{orange!15}\textbf{3.41} \\
\midrule
\multirow{22}{*}{$T{=}1$} & \multirow{5}{*}{Q3.5-4B} & EAGLE-3 & 1.10$\times$ & 1.30$\times$ & 1.31$\times$ & 1.08$\times$ & 0.75$\times$ & 0.61$\times$ & 3.18 & 0.76$\times$ & 0.87$\times$ & 0.89$\times$ & 0.74$\times$ & 0.57$\times$ & 0.45$\times$ & 2.12 \\
 &  & \cellcolor{gray!15}\textcolor{gray!70}{MTP} & \cellcolor{gray!15}\textcolor{gray!70}{1.17$\times$} & \cellcolor{gray!15}\textcolor{gray!70}{1.23$\times$} & \cellcolor{gray!15}\textcolor{gray!70}{1.20$\times$} & \cellcolor{gray!15}\textcolor{gray!70}{1.00$\times$} & \cellcolor{gray!15}\textcolor{gray!70}{0.77$\times$} & \cellcolor{gray!15}\textcolor{gray!70}{0.66$\times$} & \cellcolor{gray!15}\textcolor{gray!70}{4.25} & \cellcolor{gray!15}\textcolor{gray!70}{0.96$\times$} & \cellcolor{gray!15}\textcolor{gray!70}{0.98$\times$} & \cellcolor{gray!15}\textcolor{gray!70}{0.97$\times$} & \cellcolor{gray!15}\textcolor{gray!70}{0.85$\times$} & \cellcolor{gray!15}\textcolor{gray!70}{0.71$\times$} & \cellcolor{gray!15}\textcolor{gray!70}{0.57$\times$} & \cellcolor{gray!15}\textcolor{gray!70}{3.50} \\
 &  & DFlash (T) & 2.59$\times$ & 2.41$\times$ & 2.11$\times$ & 1.44$\times$ & 0.93$\times$ & 0.75$\times$ & 3.96 & 1.95$\times$ & 1.76$\times$ & 1.56$\times$ & 1.09$\times$ & 0.79$\times$ & 0.58$\times$ & 2.98 \\
 &  & DFlash (M) & 2.62$\times$ & 2.63$\times$ & 2.29$\times$ & 1.48$\times$ & 0.97$\times$ & 0.75$\times$ & 3.99 & 2.04$\times$ & 1.96$\times$ & 1.73$\times$ & 1.17$\times$ & 0.84$\times$ & 0.62$\times$ & 3.17 \\
 &  & \cellcolor{orange!15}\textbf{Uni-LaSpec} & \cellcolor{orange!15}\textbf{2.70}$\times$ & \cellcolor{orange!15}\textbf{2.64}$\times$ & \cellcolor{orange!15}\textbf{2.34}$\times$ & \cellcolor{orange!15}\textbf{1.55}$\times$ & \cellcolor{orange!15}\textbf{0.99}$\times$ & \cellcolor{orange!15}\textbf{0.78}$\times$ & \cellcolor{orange!15}\textbf{4.08} & \cellcolor{orange!15}\textbf{2.16}$\times$ & \cellcolor{orange!15}\textbf{2.06}$\times$ & \cellcolor{orange!15}\textbf{1.81}$\times$ & \cellcolor{orange!15}\textbf{1.26}$\times$ & \cellcolor{orange!15}\textbf{0.90}$\times$ & \cellcolor{orange!15}\textbf{0.67}$\times$ & \cellcolor{orange!15}\textbf{3.37} \\
\cmidrule(l){2-17}
 & \multirow{5}{*}{Q3.5-9B} & EAGLE-3 & 1.75$\times$ & 1.63$\times$ & 1.46$\times$ & 1.09$\times$ & 0.85$\times$ & 0.66$\times$ & 3.47 & 1.20$\times$ & 1.11$\times$ & 1.01$\times$ & 0.78$\times$ & 0.62$\times$ & 0.49$\times$ & 2.38 \\
 &  & \cellcolor{gray!15}\textcolor{gray!70}{MTP} & \cellcolor{gray!15}\textcolor{gray!70}{1.35$\times$} & \cellcolor{gray!15}\textcolor{gray!70}{1.31$\times$} & \cellcolor{gray!15}\textcolor{gray!70}{1.23$\times$} & \cellcolor{gray!15}\textcolor{gray!70}{1.02$\times$} & \cellcolor{gray!15}\textcolor{gray!70}{0.87$\times$} & \cellcolor{gray!15}\textcolor{gray!70}{0.70$\times$} & \cellcolor{gray!15}\textcolor{gray!70}{4.58} & \cellcolor{gray!15}\textcolor{gray!70}{1.08$\times$} & \cellcolor{gray!15}\textcolor{gray!70}{1.05$\times$} & \cellcolor{gray!15}\textcolor{gray!70}{0.99$\times$} & \cellcolor{gray!15}\textcolor{gray!70}{0.84$\times$} & \cellcolor{gray!15}\textcolor{gray!70}{0.73$\times$} & \cellcolor{gray!15}\textcolor{gray!70}{0.61$\times$} & \cellcolor{gray!15}\textcolor{gray!70}{3.71} \\
 &  & DFlash (T) & 3.00$\times$ & 2.62$\times$ & 2.20$\times$ & 1.47$\times$ & 1.06$\times$ & 0.78$\times$ & 4.01 & 2.21$\times$ & 1.90$\times$ & 1.63$\times$ & 1.11$\times$ & 0.83$\times$ & 0.61$\times$ & 3.01 \\
 &  & DFlash (M) & 3.01$\times$ & 2.62$\times$ & 2.21$\times$ & 1.45$\times$ & 1.07$\times$ & \textbf{0.79}$\times$ & 4.05 & 2.33$\times$ & 2.00$\times$ & 1.70$\times$ & 1.16$\times$ & 0.86$\times$ & 0.64$\times$ & 3.22 \\
 &  & \cellcolor{orange!15}\textbf{Uni-LaSpec} & \cellcolor{orange!15}\textbf{3.11}$\times$ & \cellcolor{orange!15}\textbf{2.69}$\times$ & \cellcolor{orange!15}\textbf{2.26}$\times$ & \cellcolor{orange!15}\textbf{1.49}$\times$ & \cellcolor{orange!15}\textbf{1.08}$\times$ & \cellcolor{orange!15}\textbf{0.79}$\times$ & \cellcolor{orange!15}\textbf{4.14} & \cellcolor{orange!15}\textbf{2.48}$\times$ & \cellcolor{orange!15}\textbf{2.07}$\times$ & \cellcolor{orange!15}\textbf{1.78}$\times$ & \cellcolor{orange!15}\textbf{1.20}$\times$ & \cellcolor{orange!15}\textbf{0.89}$\times$ & \cellcolor{orange!15}\textbf{0.66}$\times$ & \cellcolor{orange!15}\textbf{3.38} \\
\cmidrule(l){2-17}
 & \multirow{3}{*}{IVL3.5-4B} & DFlash (T) & 2.58$\times$ & 2.35$\times$ & 1.93$\times$ & 1.30$\times$ & 0.90$\times$ & 0.69$\times$ & 3.43 & 1.71$\times$ & 1.58$\times$ & 1.27$\times$ & 0.95$\times$ & 0.81$\times$ & 0.69$\times$ & 2.44 \\
 &  & DFlash (M) & 2.63$\times$ & 2.37$\times$ & 1.94$\times$ & 1.32$\times$ & 0.90$\times$ & 0.71$\times$ & 3.45 & 1.79$\times$ & 1.66$\times$ & 1.32$\times$ & 1.00$\times$ & \textbf{0.86}$\times$ & 0.73$\times$ & 2.63 \\
 &  & \cellcolor{orange!15}\textbf{Uni-LaSpec} & \cellcolor{orange!15}\textbf{2.78}$\times$ & \cellcolor{orange!15}\textbf{2.52}$\times$ & \cellcolor{orange!15}\textbf{2.04}$\times$ & \cellcolor{orange!15}\textbf{1.35}$\times$ & \cellcolor{orange!15}\textbf{0.94}$\times$ & \cellcolor{orange!15}\textbf{0.72}$\times$ & \cellcolor{orange!15}\textbf{3.62} & \cellcolor{orange!15}\textbf{1.87}$\times$ & \cellcolor{orange!15}\textbf{1.70}$\times$ & \cellcolor{orange!15}\textbf{1.34}$\times$ & \cellcolor{orange!15}\textbf{1.04}$\times$ & \cellcolor{orange!15}\textbf{0.86}$\times$ & \cellcolor{orange!15}\textbf{0.74}$\times$ & \cellcolor{orange!15}\textbf{2.80} \\
\cmidrule(l){2-17}
 & \multirow{3}{*}{IVL3.5-8B} & DFlash (T) & 2.97$\times$ & 2.76$\times$ & 2.31$\times$ & 1.51$\times$ & 1.00$\times$ & 0.75$\times$ & 3.92 & 1.76$\times$ & 1.62$\times$ & 1.35$\times$ & 0.98$\times$ & 0.77$\times$ & 0.63$\times$ & 2.45 \\
 &  & DFlash (M) & 2.96$\times$ & 2.77$\times$ & 2.28$\times$ & 1.47$\times$ & 0.99$\times$ & \textbf{0.76}$\times$ & 3.94 & 1.87$\times$ & 1.66$\times$ & 1.36$\times$ & 0.97$\times$ & 0.82$\times$ & 0.67$\times$ & 2.65 \\
 &  & \cellcolor{orange!15}\textbf{Uni-LaSpec} & \cellcolor{orange!15}\textbf{3.12}$\times$ & \cellcolor{orange!15}\textbf{2.83}$\times$ & \cellcolor{orange!15}\textbf{2.36}$\times$ & \cellcolor{orange!15}\textbf{1.54}$\times$ & \cellcolor{orange!15}\textbf{1.02}$\times$ & \cellcolor{orange!15}\textbf{0.76}$\times$ & \cellcolor{orange!15}\textbf{4.08} & \cellcolor{orange!15}\textbf{1.98}$\times$ & \cellcolor{orange!15}\textbf{1.71}$\times$ & \cellcolor{orange!15}\textbf{1.41}$\times$ & \cellcolor{orange!15}\textbf{1.01}$\times$ & \cellcolor{orange!15}\textbf{0.84}$\times$ & \cellcolor{orange!15}\textbf{0.68}$\times$ & \cellcolor{orange!15}\textbf{2.83} \\
\cmidrule(l){2-17}
 & \multirow{3}{*}{G4-E2B} & DFlash (T) & 4.40$\times$ & 3.87$\times$ & 3.07$\times$ & 2.16$\times$ & 1.45$\times$ & 1.04$\times$ & 4.88 & 2.12$\times$ & 1.67$\times$ & 1.37$\times$ & 1.00$\times$ & 0.84$\times$ & 0.67$\times$ & 2.35 \\
 &  & DFlash (M) & 4.47$\times$ & 3.78$\times$ & 3.07$\times$ & 2.16$\times$ & 1.46$\times$ & 1.03$\times$ & 4.91 & 2.43$\times$ & 1.84$\times$ & 1.52$\times$ & 1.11$\times$ & 0.92$\times$ & 0.74$\times$ & 2.79 \\
 &  & \cellcolor{orange!15}\textbf{Uni-LaSpec} & \cellcolor{orange!15}\textbf{4.57}$\times$ & \cellcolor{orange!15}\textbf{3.96}$\times$ & \cellcolor{orange!15}\textbf{3.17}$\times$ & \cellcolor{orange!15}\textbf{2.24}$\times$ & \cellcolor{orange!15}\textbf{1.48}$\times$ & \cellcolor{orange!15}\textbf{1.05}$\times$ & \cellcolor{orange!15}\textbf{5.03} & \cellcolor{orange!15}\textbf{2.46}$\times$ & \cellcolor{orange!15}\textbf{1.95}$\times$ & \cellcolor{orange!15}\textbf{1.57}$\times$ & \cellcolor{orange!15}\textbf{1.17}$\times$ & \cellcolor{orange!15}\textbf{0.97}$\times$ & \cellcolor{orange!15}\textbf{0.78}$\times$ & \cellcolor{orange!15}\textbf{3.01} \\
\cmidrule(l){2-17}
 & \multirow{3}{*}{G4-E4B} & DFlash (T) & 3.96$\times$ & 3.54$\times$ & 3.06$\times$ & 1.96$\times$ & 1.30$\times$ & 0.85$\times$ & 5.01 & 1.93$\times$ & 1.70$\times$ & 1.50$\times$ & 1.08$\times$ & 0.84$\times$ & 0.67$\times$ & 2.49 \\
 &  & DFlash (M) & 3.98$\times$ & 3.62$\times$ & 3.08$\times$ & 1.91$\times$ & 1.24$\times$ & 0.84$\times$ & 5.09 & 2.14$\times$ & 1.88$\times$ & 1.65$\times$ & 1.15$\times$ & 0.90$\times$ & 0.72$\times$ & 2.84 \\
 &  & \cellcolor{orange!15}\textbf{Uni-LaSpec} & \cellcolor{orange!15}\textbf{4.20}$\times$ & \cellcolor{orange!15}\textbf{3.70}$\times$ & \cellcolor{orange!15}\textbf{3.13}$\times$ & \cellcolor{orange!15}\textbf{1.99}$\times$ & \cellcolor{orange!15}\textbf{1.31}$\times$ & \cellcolor{orange!15}\textbf{0.86}$\times$ & \cellcolor{orange!15}\textbf{5.27} & \cellcolor{orange!15}\textbf{2.28}$\times$ & \cellcolor{orange!15}\textbf{1.92}$\times$ & \cellcolor{orange!15}\textbf{1.67}$\times$ & \cellcolor{orange!15}\textbf{1.22}$\times$ & \cellcolor{orange!15}\textbf{0.94}$\times$ & \cellcolor{orange!15}\textbf{0.77}$\times$ & \cellcolor{orange!15}\textbf{3.06} \\
\bottomrule
\end{tabular}

%% file: tables/table4.tex
\begin{table}[t]
\caption{\normalfont Throughput scaling of business models fine-tuned on in-house data for medical-document extraction at sampling temperature $T{=}0$. Concurrency columns report the ratio of speculative to autoregressive output throughput for the same model and $R$. Avg.\ $\tau$ denotes the unweighted mean accepted length across the six concurrency levels, with all requests included. Q3.5-2B-SFT at $R{=}4$ pools 5 repetitions of 100 prompts each; throughput is total output tokens divided by total benchmark duration, and acceptance pools the draft and accepted-token counters.}
\label{tab:business-concurrency}
\centering
\footnotesize
\setlength{\tabcolsep}{1.5pt}
\renewcommand{\arraystretch}{1.50}
\begin{tabular}{@{}llccccccc@{}}
\toprule
Model & Method & $R{=}1$ & $R{=}4$ & $R{=}8$ & $R{=}16$ & $R{=}32$ & $R{=}64$ & Avg.\ $\tau$ \\
\midrule
\multirow{3}{*}{\shortstack[l]{Q3.5-2B\\-SFT}} & \cellcolor{gray!15}\textcolor{gray!70}{MTP} & \cellcolor{gray!15}\textcolor{gray!70}{1.49$\times$} & \cellcolor{gray!15}\textcolor{gray!70}{1.18$\times$} & \cellcolor{gray!15}\textcolor{gray!70}{1.02$\times$} & \cellcolor{gray!15}\textcolor{gray!70}{1.06$\times$} & \cellcolor{gray!15}\textcolor{gray!70}{0.69$\times$} & \cellcolor{gray!15}\textcolor{gray!70}{0.54$\times$} & \cellcolor{gray!15}\textcolor{gray!70}{5.553} \\
 & DFlash (M) & 5.45$\times$ & 3.88$\times$ & 2.70$\times$ & 2.73$\times$ & 1.83$\times$ & 1.47$\times$ & 9.328 \\
 & \cellcolor{orange!15}\textbf{Uni-LaSpec} & \cellcolor{orange!15}\textbf{5.59}$\times$ & \cellcolor{orange!15}\textbf{4.11}$\times$ & \cellcolor{orange!15}\textbf{3.15}$\times$ & \cellcolor{orange!15}\textbf{3.02}$\times$ & \cellcolor{orange!15}\textbf{1.85}$\times$ & \cellcolor{orange!15}\textbf{1.80}$\times$ & \cellcolor{orange!15}\textbf{9.923} \\
\midrule
\multirow{3}{*}{\shortstack[l]{Q3.5-4B\\-SFT}} & \cellcolor{gray!15}\textcolor{gray!70}{MTP} & \cellcolor{gray!15}\textcolor{gray!70}{1.55$\times$} & \cellcolor{gray!15}\textcolor{gray!70}{1.50$\times$} & \cellcolor{gray!15}\textcolor{gray!70}{1.41$\times$} & \cellcolor{gray!15}\textcolor{gray!70}{1.39$\times$} & \cellcolor{gray!15}\textcolor{gray!70}{1.28$\times$} & \cellcolor{gray!15}\textcolor{gray!70}{1.18$\times$} & \cellcolor{gray!15}\textcolor{gray!70}{5.664} \\
 & DFlash (M) & 5.57$\times$ & 3.83$\times$ & 3.16$\times$ & 2.54$\times$ & 2.12$\times$ & 1.85$\times$ & 9.369 \\
 & \cellcolor{orange!15}\textbf{Uni-LaSpec} & \cellcolor{orange!15}\textbf{5.70}$\times$ & \cellcolor{orange!15}\textbf{3.86}$\times$ & \cellcolor{orange!15}\textbf{3.18}$\times$ & \cellcolor{orange!15}\textbf{2.60}$\times$ & \cellcolor{orange!15}\textbf{2.18}$\times$ & \cellcolor{orange!15}\textbf{1.90}$\times$ & \cellcolor{orange!15}\textbf{9.920} \\
\bottomrule
\end{tabular}
\end{table}

%% file: tables/table5.tex
\footnotesize
\definecolor{pairsame}{RGB}{234,244,229}
\definecolor{pairlow}{RGB}{253,241,234}
\definecolor{pairhigh}{RGB}{250,222,205}
\definecolor{textfill}{RGB}{242,242,242}
\setlength{\tabcolsep}{3.8pt}
\renewcommand{\arraystretch}{1.50}
\begin{tabular}{@{}lcccccccccccccccccc}
\toprule
\multirow{3}{*}[-5.4pt]{Verifier} & \multirow{3}{*}[-5.4pt]{Modality} & \multirow{3}{*}[-5.4pt]{None} & \multicolumn{5}{c}{Top-8} & \multicolumn{5}{c}{Top-32} & \multicolumn{5}{c}{Top-64} & \multirow{3}{*}[-5.4pt]{All} \\
\cmidrule(lr){4-8} \cmidrule(lr){9-13} \cmidrule(lr){14-18}
 & & & \multicolumn{2}{c}{Quant.} & \multicolumn{2}{c}{Softmax} & \multirow{2}{*}[-2.7pt]{$\kappa$} & \multicolumn{2}{c}{Quant.} & \multicolumn{2}{c}{Softmax} & \multirow{2}{*}[-2.7pt]{$\kappa$} & \multicolumn{2}{c}{Quant.} & \multicolumn{2}{c}{Softmax} & \multirow{2}{*}[-2.7pt]{$\kappa$} & \\
\cmidrule(lr){4-5} \cmidrule(lr){6-7} \cmidrule(lr){9-10} \cmidrule(lr){11-12} \cmidrule(lr){14-15} \cmidrule(lr){16-17}
 & & & $\tau$ & $C_{\mathrm{sel}}$ & $\tau$ & $C_{\mathrm{sel}}$ &  & $\tau$ & $C_{\mathrm{sel}}$ & $\tau$ & $C_{\mathrm{sel}}$ &  & $\tau$ & $C_{\mathrm{sel}}$ & $\tau$ & $C_{\mathrm{sel}}$ &  & \\
\midrule
\multirow{2}{*}{Q3.5-4B} & Text & \cellcolor{textfill}\textcolor{gray}{5.34} & \multicolumn{5}{c}{\cellcolor{textfill}\textcolor{gray}{5.34}} & \multicolumn{5}{c}{\cellcolor{textfill}\textcolor{gray}{5.34}} & \multicolumn{5}{c}{\cellcolor{textfill}\textcolor{gray}{\textbf{5.38}}} & \cellcolor{textfill}\textcolor{gray}{5.35} \\
 & MM & 3.86 & \cellcolor{pairsame}4.06 & 0.595 & \cellcolor{pairsame}4.06 & 11.411 & 19.17$\times$ & \cellcolor{pairlow}4.07 & 0.846 & \cellcolor{pairhigh}\textbf{4.08} & 18.104 & 21.40$\times$ & \cellcolor{pairsame}4.07 & 1.156 & \cellcolor{pairsame}4.07 & 20.973 & 18.14$\times$ & 4.04 \\
\midrule
\multirow{2}{*}{Q3.5-9B} & Text & \cellcolor{textfill}\textcolor{gray}{\textbf{5.26}} & \multicolumn{5}{c}{\cellcolor{textfill}\textcolor{gray}{\textbf{5.26}}} & \multicolumn{5}{c}{\cellcolor{textfill}\textcolor{gray}{\textbf{5.26}}} & \multicolumn{5}{c}{\cellcolor{textfill}\textcolor{gray}{5.21}} & \cellcolor{textfill}\textcolor{gray}{5.25} \\
 & MM & 3.79 & \cellcolor{pairhigh}\textbf{3.95} & 0.722 & \cellcolor{pairlow}3.93 & 18.291 & 25.35$\times$ & \cellcolor{pairsame}3.93 & 1.229 & \cellcolor{pairsame}3.93 & 25.362 & 20.63$\times$ & \cellcolor{pairhigh}\textbf{3.95} & 1.549 & \cellcolor{pairlow}3.94 & 30.479 & 19.67$\times$ & 3.94 \\
\midrule
\multirow{2}{*}{IVL3.5-4B} & Text & \cellcolor{textfill}\textcolor{gray}{4.64} & \multicolumn{5}{c}{\cellcolor{textfill}\textcolor{gray}{\textbf{4.73}}} & \multicolumn{5}{c}{\cellcolor{textfill}\textcolor{gray}{4.72}} & \multicolumn{5}{c}{\cellcolor{textfill}\textcolor{gray}{4.71}} & \cellcolor{textfill}\textcolor{gray}{4.66} \\
 & MM & 3.52 & \cellcolor{pairhigh}3.69 & 0.538 & \cellcolor{pairlow}3.67 & 2.437 & 4.53$\times$ & \cellcolor{pairhigh}3.69 & 0.729 & \cellcolor{pairlow}3.67 & 3.135 & 4.30$\times$ & \cellcolor{pairsame}3.70 & 0.899 & \cellcolor{pairsame}3.70 & 4.067 & 4.52$\times$ & \textbf{3.72} \\
\midrule
\multirow{2}{*}{IVL3.5-8B} & Text & \cellcolor{textfill}\textcolor{gray}{5.37} & \multicolumn{5}{c}{\cellcolor{textfill}\textcolor{gray}{\textbf{5.43}}} & \multicolumn{5}{c}{\cellcolor{textfill}\textcolor{gray}{5.41}} & \multicolumn{5}{c}{\cellcolor{textfill}\textcolor{gray}{5.40}} & \cellcolor{textfill}\textcolor{gray}{5.40} \\
 & MM & 3.48 & \cellcolor{pairlow}3.66 & 0.726 & \cellcolor{pairhigh}3.67 & 2.504 & 3.45$\times$ & \cellcolor{pairhigh}\textbf{3.72} & 1.120 & \cellcolor{pairlow}3.68 & 3.438 & 3.07$\times$ & \cellcolor{pairhigh}\textbf{3.72} & 1.312 & \cellcolor{pairlow}3.66 & 4.440 & 3.38$\times$ & 3.71 \\
\midrule
\multirow{2}{*}{G4-E2B} & Text & \cellcolor{textfill}\textcolor{gray}{5.29} & \multicolumn{5}{c}{\cellcolor{textfill}\textcolor{gray}{5.35}} & \multicolumn{5}{c}{\cellcolor{textfill}\textcolor{gray}{5.35}} & \multicolumn{5}{c}{\cellcolor{textfill}\textcolor{gray}{5.32}} & \cellcolor{textfill}\textcolor{gray}{\textbf{5.36}} \\
 & MM & 3.08 & \cellcolor{pairlow}3.22 & 0.552 & \cellcolor{pairhigh}3.24 & 4.573 & 8.28$\times$ & \cellcolor{pairlow}3.24 & 0.699 & \cellcolor{pairhigh}\textbf{3.27} & 5.911 & 8.46$\times$ & \cellcolor{pairlow}3.22 & 0.784 & \cellcolor{pairhigh}3.24 & 9.477 & 12.09$\times$ & 3.25 \\
\midrule
\multirow{2}{*}{G4-E4B} & Text & \cellcolor{textfill}\textcolor{gray}{5.84} & \multicolumn{5}{c}{\cellcolor{textfill}\textcolor{gray}{5.81}} & \multicolumn{5}{c}{\cellcolor{textfill}\textcolor{gray}{\textbf{5.86}}} & \multicolumn{5}{c}{\cellcolor{textfill}\textcolor{gray}{5.80}} & \cellcolor{textfill}\textcolor{gray}{5.82} \\
 & MM & 3.21 & \cellcolor{pairsame}3.42 & 0.643 & \cellcolor{pairsame}3.42 & 4.951 & 7.69$\times$ & \cellcolor{pairhigh}3.43 & 0.890 & \cellcolor{pairlow}3.42 & 6.425 & 7.22$\times$ & \cellcolor{pairlow}3.42 & 1.093 & \cellcolor{pairhigh}\textbf{3.44} & 12.288 & 11.24$\times$ & 3.43 \\
\bottomrule
\end{tabular}

%% file: tables/table6.tex
\footnotesize
\setlength{\tabcolsep}{4.5pt}
\renewcommand{\arraystretch}{1.50}
\begin{tabular}{@{}lllccccccccccc@{}}
\toprule
\multirow{2}{*}[-2.7pt]{Verifier} & \multirow{2}{*}[-2.7pt]{Scoring} & \multirow{2}{*}[-2.7pt]{Kernels} & \multirow{2}{*}[-2.7pt]{$\tau$} & \multicolumn{2}{c}{$R{=}1$} & \multicolumn{2}{c}{$R{=}4$} & \multicolumn{2}{c}{$R{=}16$} & \multicolumn{2}{c}{$R{=}32$} & \multicolumn{2}{c}{$R{=}64$} \\
\cmidrule(lr){5-6} \cmidrule(lr){7-8} \cmidrule(lr){9-10} \cmidrule(lr){11-12} \cmidrule(lr){13-14}
 & & & & $C_{\mathrm{sel}}$ & $\eta_{\mathrm{sel}}$ & $C_{\mathrm{sel}}$ & $\eta_{\mathrm{sel}}$ & $C_{\mathrm{sel}}$ & $\eta_{\mathrm{sel}}$ & $C_{\mathrm{sel}}$ & $\eta_{\mathrm{sel}}$ & $C_{\mathrm{sel}}$ & $\eta_{\mathrm{sel}}$ \\
\midrule
\multirow{3}{*}{Q3.5-4B} & Softmax & Unfused & 4.07 & 0.99 & 1.00$\times$ & 2.12 & 1.00$\times$ & 4.16 & 1.00$\times$ & 8.61 & 1.00$\times$ & 8.95 & 1.00$\times$ \\
 & Quant. & Unfused & 4.08 & 0.74 & 1.34$\times$ & 0.75 & 2.83$\times$ & 0.98 & 4.25$\times$ & 1.22 & 7.07$\times$ & 1.41 & 6.36$\times$ \\
 & Quant. & Fused & 4.07 & \textbf{0.38} & 2.58$\times$ & \textbf{0.44} & 4.83$\times$ & \textbf{0.62} & 6.73$\times$ & \textbf{0.76} & 11.40$\times$ & \textbf{0.94} & 9.54$\times$ \\
\midrule
\multirow{3}{*}{Q3.5-9B} & Softmax & Unfused & 3.92 & 1.03 & 1.00$\times$ & 2.07 & 1.00$\times$ & 5.17 & 1.00$\times$ & 13.98 & 1.00$\times$ & 15.71 & 1.00$\times$ \\
 & Quant. & Unfused & 3.92 & 0.70 & 1.48$\times$ & 0.75 & 2.76$\times$ & 1.15 & 4.48$\times$ & 1.47 & 9.49$\times$ & 1.55 & 10.15$\times$ \\
 & Quant. & Fused & 3.92 & \textbf{0.38} & 2.70$\times$ & \textbf{0.41} & 5.03$\times$ & \textbf{0.63} & 8.18$\times$ & \textbf{0.78} & 17.95$\times$ & \textbf{0.87} & 18.04$\times$ \\
\bottomrule
\end{tabular}

%% file: tables/table7.tex
\begin{table}[t]
\caption{\normalfont Comparison of auxiliary-feature configurations under Top-8
visual selection at concurrency $R{=}4$. $|\mathcal{A}|$ denotes the number of
verifier layers providing auxiliary features. $\tau$ and TPS report
unweighted benchmark means of accepted length and output throughput
(tokens/s), respectively. $S$ is the ratio of TPS to the corresponding
autoregressive throughput at the same sampling temperature. Results are
aggregated separately for text and multimodal benchmarks. Acceptance
length applies only to speculative methods.}
\label{tab:auxiliary-top8}
\centering
\footnotesize
\setlength{\tabcolsep}{2pt}
\renewcommand{\arraystretch}{1.50}
\begin{tabular*}{\columnwidth}{@{}llc@{\extracolsep{\fill}}cccccc@{}}
\toprule
\multirow{2}{*}[-2.7pt]{Samp.} & \multirow{2}{*}[-2.7pt]{Method} & & \multicolumn{3}{c}{Text} & \multicolumn{3}{c}{Multimodal} \\
\cmidrule(lr){4-6}\cmidrule(lr){7-9}
 & & $|\mathcal{A}|$ & $\tau$ & TPS & $S$ & $\tau$ & TPS & $S$ \\
\midrule
\multirow{4}{*}{$T{=}0$} & AR & -- & -- & 321.1 & 1.00$\times$ & -- & 306.3 & 1.00$\times$ \\
 & \textbf{Uni-LaSpec} & 3 & 5.35 & 1047.5 & 3.26$\times$ & 4.00 & 728.8 & 2.38$\times$ \\
 & \textbf{Uni-LaSpec} & 5 & 5.33 & 1037.2 & 3.23$\times$ & 3.98 & 730.9 & 2.39$\times$ \\
 & \textbf{Uni-LaSpec} & 8 & \textbf{5.38} & \textbf{1090.2} & \textbf{3.40}$\times$ & \textbf{4.08} & \textbf{756.3} & \textbf{2.47}$\times$ \\
\midrule
\multirow{4}{*}{$T{=}1$} & AR & -- & -- & 317.2 & 1.00$\times$ & -- & 308.0 & 1.00$\times$ \\
 & \textbf{Uni-LaSpec} & 3 & 4.08 & 830.2 & 2.62$\times$ & 3.33 & \textbf{628.4} & \textbf{2.04}$\times$ \\
 & \textbf{Uni-LaSpec} & 5 & \textbf{4.12} & \textbf{842.6} & \textbf{2.66}$\times$ & 3.33 & 624.5 & 2.03$\times$ \\
 & \textbf{Uni-LaSpec} & 8 & 4.07 & 810.9 & 2.56$\times$ & \textbf{3.35} & 606.7 & 1.97$\times$ \\
\bottomrule
\end{tabular*}
\end{table}

%% file: tables/table8.tex
\footnotesize
\definecolor{famlin}{RGB}{249,245,252}\definecolor{famtrunc}{RGB}{254,251,229}\definecolor{famtree}{RGB}{240,249,242}\definecolor{famuni}{RGB}{254,244,233}
\setlength{\tabcolsep}{1.5pt}
\renewcommand{\arraystretch}{1.162}
\setlength{\aboverulesep}{0.2ex}
\setlength{\belowrulesep}{0.3ex}
\begin{tabular}{llccccccccccccccccc}
\toprule
 &  & \multicolumn{7}{c}{Text Benchmarks} & \multicolumn{10}{c}{Multimodal Benchmarks} \\
\cmidrule(lr){3-9} \cmidrule(lr){10-19}
Model & Strategy & \multicolumn{3}{c}{Math} & \multicolumn{3}{c}{Code} & \multicolumn{1}{c}{Chat} & \multicolumn{4}{c}{Document \& OCR} & \multicolumn{4}{c}{General Vision} & \multicolumn{2}{c}{Capt.\ \& Chat} \\
\cmidrule(lr){3-5} \cmidrule(lr){6-8} \cmidrule(lr){9-9} \cmidrule(lr){10-13} \cmidrule(lr){14-17} \cmidrule(lr){18-19}
 &  & GSM8K & MATH & AIME & HEval & MBPP & LCB & MT-B & ODoc & OCR2 & DVQA & olmO & CXiv & TVQA & MMMU & GQA & COCO & MMMT \\
\midrule
\multirow{9}{*}{Q3.5-4B} & \cellcolor{famlin}Linear & \cellcolor{famlin}5.31 & \cellcolor{famlin}6.65 & \cellcolor{famlin}6.55 & \cellcolor{famlin}5.46 & \cellcolor{famlin}4.99 & \cellcolor{famlin}4.64 & \cellcolor{famlin}3.81 & \cellcolor{famlin}3.03 & \cellcolor{famlin}3.81 & \cellcolor{famlin}4.12 & \cellcolor{famlin}4.92 & \cellcolor{famlin}5.25 & \cellcolor{famlin}3.66 & \cellcolor{famlin}4.26 & \cellcolor{famlin}3.70 & \cellcolor{famlin}3.56 & \cellcolor{famlin}4.30 \\
 & \cellcolor{famtrunc}Trunc.\ $N{=}5$ & \cellcolor{famtrunc}4.02 & \cellcolor{famtrunc}4.53 & \cellcolor{famtrunc}4.54 & \cellcolor{famtrunc}4.09 & \cellcolor{famtrunc}3.87 & \cellcolor{famtrunc}3.78 & \cellcolor{famtrunc}3.27 & \cellcolor{famtrunc}2.78 & \cellcolor{famtrunc}3.26 & \cellcolor{famtrunc}3.51 & \cellcolor{famtrunc}3.93 & \cellcolor{famtrunc}3.98 & \cellcolor{famtrunc}3.28 & \cellcolor{famtrunc}3.56 & \cellcolor{famtrunc}3.28 & \cellcolor{famtrunc}3.22 & \cellcolor{famtrunc}3.56 \\
\bottomrule
\end{tabular}

%% file: tables/table9.tex
\footnotesize
\definecolor{famlin}{RGB}{249,245,252}\definecolor{famtrunc}{RGB}{254,251,229}\definecolor{famtree}{RGB}{240,249,242}\definecolor{famuni}{RGB}{254,244,233}
\setlength{\tabcolsep}{1.0pt}
\renewcommand{\arraystretch}{1.165}
\setlength{\aboverulesep}{0.2ex}
\setlength{\belowrulesep}{0.3ex}
\begin{tabular}{llccccccccccccccccc}
\toprule
 &  & \multicolumn{7}{c}{Text Benchmarks} & \multicolumn{10}{c}{Multimodal Benchmarks} \\
\cmidrule(lr){3-9} \cmidrule(lr){10-19}
Model & Strategy & \multicolumn{3}{c}{Math} & \multicolumn{3}{c}{Code} & \multicolumn{1}{c}{Chat} & \multicolumn{4}{c}{Document \& OCR} & \multicolumn{4}{c}{General Vision} & \multicolumn{2}{c}{Capt.\ \& Chat} \\
\cmidrule(lr){3-5} \cmidrule(lr){6-8} \cmidrule(lr){9-9} \cmidrule(lr){10-13} \cmidrule(lr){14-17} \cmidrule(lr){18-19}
 &  & GSM8K & MATH & AIME & HEval & MBPP & LCB & MT-B & ODoc & OCR2 & DVQA & olmO & CXiv & TVQA & MMMU & GQA & COCO & MMMT \\
\midrule
\multirow{9}{*}{Q3.5-4B} & \cellcolor{famlin}Linear & \cellcolor{famlin}3.71$\times$ & \cellcolor{famlin}4.62$\times$ & \cellcolor{famlin}4.55$\times$ & \cellcolor{famlin}3.80$\times$ & \cellcolor{famlin}3.48$\times$ & \cellcolor{famlin}3.22$\times$ & \cellcolor{famlin}2.67$\times$ & \cellcolor{famlin}2.09$\times$ & \cellcolor{famlin}2.61$\times$ & \cellcolor{famlin}2.59$\times$ & \cellcolor{famlin}3.33$\times$ & \cellcolor{famlin}3.57$\times$ & \cellcolor{famlin}2.47$\times$ & \cellcolor{famlin}2.93$\times$ & \cellcolor{famlin}2.55$\times$ & \cellcolor{famlin}2.48$\times$ & \cellcolor{famlin}2.94$\times$ \\
 & \cellcolor{famtrunc}Trunc.\ $N{=}5$ & \cellcolor{famtrunc}2.86$\times$ & \cellcolor{famtrunc}3.21$\times$ & \cellcolor{famtrunc}3.20$\times$ & \cellcolor{famtrunc}2.91$\times$ & \cellcolor{famtrunc}2.76$\times$ & \cellcolor{famtrunc}2.68$\times$ & \cellcolor{famtrunc}2.33$\times$ & \cellcolor{famtrunc}1.96$\times$ & \cellcolor{famtrunc}2.29$\times$ & \cellcolor{famtrunc}2.30$\times$ & \cellcolor{famtrunc}2.74$\times$ & \cellcolor{famtrunc}2.79$\times$ & \cellcolor{famtrunc}2.27$\times$ & \cellcolor{famtrunc}2.52$\times$ & \cellcolor{famtrunc}2.31$\times$ & \cellcolor{famtrunc}2.28$\times$ & \cellcolor{famtrunc}2.49$\times$ \\
\bottomrule
\end{tabular}

%% file: tables/table10.tex
\footnotesize
\definecolor{famlin}{RGB}{249,245,252}\definecolor{famtrunc}{RGB}{254,251,229}\definecolor{famtree}{RGB}{240,249,242}\definecolor{famuni}{RGB}{254,244,233}
\setlength{\tabcolsep}{2.2pt}
\renewcommand{\arraystretch}{1.169}
\setlength{\aboverulesep}{0.2ex}
\setlength{\belowrulesep}{0.3ex}
\begin{tabular}{llcccccccccccccc}
\toprule
\multirow{2}{*}[-2.7pt]{Model} & \multirow{2}{*}[-2.7pt]{Strategy} & \multicolumn{7}{c}{Text Benchmarks} & \multicolumn{7}{c}{Multimodal Benchmarks} \\
\cmidrule(lr){3-9} \cmidrule(lr){10-16}
 &  & $R{=}1$ & $R{=}4$ & $R{=}8$ & $R{=}16$ & $R{=}32$ & $R{=}64$ & Avg.\ $\tau$ & $R{=}1$ & $R{=}4$ & $R{=}8$ & $R{=}16$ & $R{=}32$ & $R{=}64$ & Avg.\ $\tau$ \\
\midrule
\multirow{9}{*}{Q3.5-4B} & \cellcolor{famlin}Linear & \cellcolor{famlin}3.72$\times$ & \cellcolor{famlin}3.30$\times$ & \cellcolor{famlin}2.70$\times$ & \cellcolor{famlin}2.26$\times$ & \cellcolor{famlin}1.34$\times$ & \cellcolor{famlin}1.01$\times$ & \cellcolor{famlin}5.36 & \cellcolor{famlin}2.76$\times$ & \cellcolor{famlin}2.34$\times$ & \cellcolor{famlin}1.90$\times$ & \cellcolor{famlin}1.37$\times$ & \cellcolor{famlin}1.08$\times$ & \cellcolor{famlin}0.78$\times$ & \cellcolor{famlin}4.05 \\
 & \cellcolor{famtrunc}Trunc.\ $N{=}5$ & \cellcolor{famtrunc}2.85$\times$ & \cellcolor{famtrunc}2.74$\times$ & \cellcolor{famtrunc}2.59$\times$ & \cellcolor{famtrunc}2.53$\times$ & \cellcolor{famtrunc}1.67$\times$ & \cellcolor{famtrunc}1.36$\times$ & \cellcolor{famtrunc}4.02 & \cellcolor{famtrunc}2.40$\times$ & \cellcolor{famtrunc}2.22$\times$ & \cellcolor{famtrunc}2.04$\times$ & \cellcolor{famtrunc}1.65$\times$ & \cellcolor{famtrunc}1.41$\times$ & \cellcolor{famtrunc}1.10$\times$ & \cellcolor{famtrunc}3.42 \\
 & \cellcolor{famtrunc}Trunc.\ $N{=}7$ & \cellcolor{famtrunc}3.21$\times$ & \cellcolor{famtrunc}3.06$\times$ & \cellcolor{famtrunc}2.80$\times$ & \cellcolor{famtrunc}2.71$\times$ & \cellcolor{famtrunc}1.70$\times$ & \cellcolor{famtrunc}1.34$\times$ & \cellcolor{famtrunc}4.56 & \cellcolor{famtrunc}2.55$\times$ & \cellcolor{famtrunc}2.35$\times$ & \cellcolor{famtrunc}2.09$\times$ & \cellcolor{famtrunc}1.67$\times$ & \cellcolor{famtrunc}1.38$\times$ & \cellcolor{famtrunc}1.06$\times$ & \cellcolor{famtrunc}3.71 \\
\bottomrule
\end{tabular}

%% file: tables/table11.tex
\footnotesize
\definecolor{famlin}{RGB}{249,245,252}\definecolor{famtrunc}{RGB}{254,251,229}\definecolor{famtree}{RGB}{240,249,242}\definecolor{famuni}{RGB}{254,244,233}
\setlength{\tabcolsep}{2.2pt}
\renewcommand{\arraystretch}{1.33}
\setlength{\aboverulesep}{0.2ex}
\setlength{\belowrulesep}{0.3ex}
\begin{tabular}{lllcccccccccccccc}
\toprule
\multirow{2}{*}[-2.7pt]{Device} & \multirow{2}{*}[-2.7pt]{Model} & \multirow{2}{*}[-2.7pt]{Strategy} & \multicolumn{7}{c}{Text Benchmarks} & \multicolumn{7}{c}{Multimodal Benchmarks} \\
\cmidrule(lr){4-10} \cmidrule(lr){11-17}
 &  &  & $R{=}1$ & $R{=}4$ & $R{=}8$ & $R{=}16$ & $R{=}32$ & $R{=}64$ & Avg.\ $\tau$ & $R{=}1$ & $R{=}4$ & $R{=}8$ & $R{=}16$ & $R{=}32$ & $R{=}64$ & Avg.\ $\tau$ \\
\midrule
\multirow{16}{*}{RTX 5090} & \multirow{8}{*}{Q3.5-4B} & \cellcolor{famlin}Linear & \cellcolor{famlin}3.05$\times$ & \cellcolor{famlin}3.24$\times$ & \cellcolor{famlin}2.90$\times$ & \cellcolor{famlin}2.08$\times$ & \cellcolor{famlin}1.35$\times$ & \cellcolor{famlin}0.90$\times$ & \cellcolor{famlin}5.35 & \cellcolor{famlin}2.32$\times$ & \cellcolor{famlin}2.35$\times$ & \cellcolor{famlin}2.13$\times$ & \cellcolor{famlin}1.51$\times$ & \cellcolor{famlin}1.03$\times$ & \cellcolor{famlin}0.77$\times$ & \cellcolor{famlin}4.06 \\
 &  & \cellcolor{famtrunc}Trunc.\ $N{=}5$ & \cellcolor{famtrunc}2.41$\times$ & \cellcolor{famtrunc}2.61$\times$ & \cellcolor{famtrunc}2.65$\times$ & \cellcolor{famtrunc}2.23$\times$ & \cellcolor{famtrunc}1.68$\times$ & \cellcolor{famtrunc}1.21$\times$ & \cellcolor{famtrunc}4.01 & \cellcolor{famtrunc}2.05$\times$ & \cellcolor{famtrunc}2.13$\times$ & \cellcolor{famtrunc}2.17$\times$ & \cellcolor{famtrunc}1.74$\times$ & \cellcolor{famtrunc}1.34$\times$ & \cellcolor{famtrunc}1.06$\times$ & \cellcolor{famtrunc}3.42 \\
\bottomrule
\end{tabular}

%% file: tables/table11_h100.tex
\footnotesize
\definecolor{famlin}{RGB}{249,245,252}\definecolor{famtrunc}{RGB}{254,251,229}\definecolor{famtree}{RGB}{240,249,242}\definecolor{famuni}{RGB}{254,244,233}
\setlength{\tabcolsep}{2.2pt}
\renewcommand{\arraystretch}{1.33}
\setlength{\aboverulesep}{0.2ex}
\setlength{\belowrulesep}{0.3ex}
\begin{tabular}{lllcccccccccccccc}
\toprule
\multirow{2}{*}[-2.7pt]{Device} & \multirow{2}{*}[-2.7pt]{Model} & \multirow{2}{*}[-2.7pt]{Strategy} & \multicolumn{7}{c}{Text Benchmarks} & \multicolumn{7}{c}{Multimodal Benchmarks} \\
\cmidrule(lr){4-10} \cmidrule(lr){11-17}
 &  &  & $R{=}1$ & $R{=}4$ & $R{=}8$ & $R{=}16$ & $R{=}32$ & $R{=}64$ & Avg.\ $\tau$ & $R{=}1$ & $R{=}4$ & $R{=}8$ & $R{=}16$ & $R{=}32$ & $R{=}64$ & Avg.\ $\tau$ \\
\midrule
\multirow{32}{*}{H100} & \multirow{8}{*}{Q3.5-4B} & \cellcolor{famlin}Linear & \cellcolor{famlin}3.75$\times$ & \cellcolor{famlin}3.29$\times$ & \cellcolor{famlin}2.91$\times$ & \cellcolor{famlin}2.31$\times$ & \cellcolor{famlin}1.74$\times$ & \cellcolor{famlin}1.33$\times$ & \cellcolor{famlin}5.38 & \cellcolor{famlin}2.75$\times$ & \cellcolor{famlin}2.43$\times$ & \cellcolor{famlin}2.15$\times$ & \cellcolor{famlin}1.72$\times$ & \cellcolor{famlin}1.32$\times$ & \cellcolor{famlin}1.04$\times$ & \cellcolor{famlin}4.06 \\
 &  & \cellcolor{famtrunc}Trunc.\ $N{=}5$ & \cellcolor{famtrunc}2.94$\times$ & \cellcolor{famtrunc}2.77$\times$ & \cellcolor{famtrunc}2.51$\times$ & \cellcolor{famtrunc}2.26$\times$ & \cellcolor{famtrunc}1.95$\times$ & \cellcolor{famtrunc}1.64$\times$ & \cellcolor{famtrunc}4.01 & \cellcolor{famtrunc}2.43$\times$ & \cellcolor{famtrunc}2.28$\times$ & \cellcolor{famtrunc}2.10$\times$ & \cellcolor{famtrunc}1.80$\times$ & \cellcolor{famtrunc}1.52$\times$ & \cellcolor{famtrunc}\textbf{1.36}$\times$ & \cellcolor{famtrunc}3.42 \\
 &  & \cellcolor{famtrunc}Trunc.\ $N{=}7$ & \cellcolor{famtrunc}3.31$\times$ & \cellcolor{famtrunc}3.05$\times$ & \cellcolor{famtrunc}2.85$\times$ & \cellcolor{famtrunc}2.45$\times$ & \cellcolor{famtrunc}2.02$\times$ & \cellcolor{famtrunc}1.63$\times$ & \cellcolor{famtrunc}4.55 & \cellcolor{famtrunc}2.60$\times$ & \cellcolor{famtrunc}2.41$\times$ & \cellcolor{famtrunc}2.24$\times$ & \cellcolor{famtrunc}\textbf{1.94}$\times$ & \cellcolor{famtrunc}\textbf{1.57}$\times$ & \cellcolor{famtrunc}1.31$\times$ & \cellcolor{famtrunc}3.71 \\
 &  & \cellcolor{famtrunc}Dyn.\ trunc. & \cellcolor{famtrunc}3.55$\times$ & \cellcolor{famtrunc}3.22$\times$ & \cellcolor{famtrunc}2.94$\times$ & \cellcolor{famtrunc}2.51$\times$ & \cellcolor{famtrunc}2.07$\times$ & \cellcolor{famtrunc}1.65$\times$ & \cellcolor{famtrunc}4.93 & \cellcolor{famtrunc}2.53$\times$ & \cellcolor{famtrunc}2.33$\times$ & \cellcolor{famtrunc}2.16$\times$ & \cellcolor{famtrunc}1.91$\times$ & \cellcolor{famtrunc}\textbf{1.57}$\times$ & \cellcolor{famtrunc}1.33$\times$ & \cellcolor{famtrunc}3.65 \\
 &  & \cellcolor{famtree}Tree $B{=}15$ & \cellcolor{famtree}3.80$\times$ & \cellcolor{famtree}3.24$\times$ & \cellcolor{famtree}1.99$\times$ & \cellcolor{famtree}1.72$\times$ & \cellcolor{famtree}1.19$\times$ & \cellcolor{famtree}0.86$\times$ & \cellcolor{famtree}6.15 & \cellcolor{famtree}2.82$\times$ & \cellcolor{famtree}2.42$\times$ & \cellcolor{famtree}1.49$\times$ & \cellcolor{famtree}1.29$\times$ & \cellcolor{famtree}0.95$\times$ & \cellcolor{famtree}0.73$\times$ & \cellcolor{famtree}4.74 \\
 &  & \cellcolor{famtree}Tree $B{=}31$ & \cellcolor{famtree}3.67$\times$ & \cellcolor{famtree}2.24$\times$ & \cellcolor{famtree}1.93$\times$ & \cellcolor{famtree}1.20$\times$ & \cellcolor{famtree}0.77$\times$ & \cellcolor{famtree}0.54$\times$ & \cellcolor{famtree}6.64 & \cellcolor{famtree}2.74$\times$ & \cellcolor{famtree}1.66$\times$ & \cellcolor{famtree}1.44$\times$ & \cellcolor{famtree}0.96$\times$ & \cellcolor{famtree}0.63$\times$ & \cellcolor{famtree}0.48$\times$ & \cellcolor{famtree}5.09 \\
 &  & \cellcolor{famtree}Tree $B{=}63$ & \cellcolor{famtree}3.07$\times$ & \cellcolor{famtree}1.47$\times$ & \cellcolor{famtree}0.89$\times$ & \cellcolor{famtree}0.49$\times$ & \cellcolor{famtree}0.30$\times$ & \cellcolor{famtree}0.21$\times$ & \cellcolor{famtree}\textbf{7.01} & \cellcolor{famtree}2.27$\times$ & \cellcolor{famtree}1.14$\times$ & \cellcolor{famtree}0.70$\times$ & \cellcolor{famtree}0.41$\times$ & \cellcolor{famtree}0.26$\times$ & \cellcolor{famtree}0.19$\times$ & \cellcolor{famtree}\textbf{5.39} \\
 &  & \cellcolor{famuni}\textbf{Uni-LaSpec} & \cellcolor{famuni}\textbf{4.47}$\times$ & \cellcolor{famuni}\textbf{3.68}$\times$ & \cellcolor{famuni}\textbf{3.08}$\times$ & \cellcolor{famuni}\textbf{2.53}$\times$ & \cellcolor{famuni}\textbf{2.09}$\times$ & \cellcolor{famuni}\textbf{1.70}$\times$ & \cellcolor{famuni}5.32 & \cellcolor{famuni}\textbf{3.19}$\times$ & \cellcolor{famuni}\textbf{2.64}$\times$ & \cellcolor{famuni}\textbf{2.31}$\times$ & \cellcolor{famuni}1.87$\times$ & \cellcolor{famuni}1.53$\times$ & \cellcolor{famuni}1.35$\times$ & \cellcolor{famuni}4.02 \\
\cmidrule(l){2-17}
 & \multirow{8}{*}{Q3.5-9B} & \cellcolor{famlin}Linear & \cellcolor{famlin}3.27$\times$ & \cellcolor{famlin}2.99$\times$ & \cellcolor{famlin}2.75$\times$ & \cellcolor{famlin}2.31$\times$ & \cellcolor{famlin}1.75$\times$ & \cellcolor{famlin}1.27$\times$ & \cellcolor{famlin}5.23 & \cellcolor{famlin}2.40$\times$ & \cellcolor{famlin}2.20$\times$ & \cellcolor{famlin}2.04$\times$ & \cellcolor{famlin}1.69$\times$ & \cellcolor{famlin}1.26$\times$ & \cellcolor{famlin}0.95$\times$ & \cellcolor{famlin}3.93 \\
 &  & \cellcolor{famtrunc}Trunc.\ $N{=}5$ & \cellcolor{famtrunc}2.58$\times$ & \cellcolor{famtrunc}2.53$\times$ & \cellcolor{famtrunc}2.45$\times$ & \cellcolor{famtrunc}2.30$\times$ & \cellcolor{famtrunc}2.04$\times$ & \cellcolor{famtrunc}1.64$\times$ & \cellcolor{famtrunc}3.96 & \cellcolor{famtrunc}2.15$\times$ & \cellcolor{famtrunc}2.08$\times$ & \cellcolor{famtrunc}2.01$\times$ & \cellcolor{famtrunc}1.85$\times$ & \cellcolor{famtrunc}1.56$\times$ & \cellcolor{famtrunc}1.28$\times$ & \cellcolor{famtrunc}3.35 \\
 &  & \cellcolor{famtrunc}Trunc.\ $N{=}7$ & \cellcolor{famtrunc}2.87$\times$ & \cellcolor{famtrunc}2.77$\times$ & \cellcolor{famtrunc}2.67$\times$ & \cellcolor{famtrunc}2.45$\times$ & \cellcolor{famtrunc}2.12$\times$ & \cellcolor{famtrunc}1.61$\times$ & \cellcolor{famtrunc}4.47 & \cellcolor{famtrunc}2.29$\times$ & \cellcolor{famtrunc}2.19$\times$ & \cellcolor{famtrunc}2.08$\times$ & \cellcolor{famtrunc}\textbf{1.90}$\times$ & \cellcolor{famtrunc}1.53$\times$ & \cellcolor{famtrunc}1.24$\times$ & \cellcolor{famtrunc}3.62 \\
 &  & \cellcolor{famtrunc}Dyn.\ trunc. & \cellcolor{famtrunc}2.93$\times$ & \cellcolor{famtrunc}2.78$\times$ & \cellcolor{famtrunc}2.62$\times$ & \cellcolor{famtrunc}2.45$\times$ & \cellcolor{famtrunc}\textbf{2.13}$\times$ & \cellcolor{famtrunc}1.64$\times$ & \cellcolor{famtrunc}4.72 & \cellcolor{famtrunc}2.15$\times$ & \cellcolor{famtrunc}2.07$\times$ & \cellcolor{famtrunc}2.00$\times$ & \cellcolor{famtrunc}1.82$\times$ & \cellcolor{famtrunc}1.57$\times$ & \cellcolor{famtrunc}1.27$\times$ & \cellcolor{famtrunc}3.53 \\
 &  & \cellcolor{famtree}Tree $B{=}15$ & \cellcolor{famtree}3.44$\times$ & \cellcolor{famtree}3.06$\times$ & \cellcolor{famtree}2.38$\times$ & \cellcolor{famtree}1.85$\times$ & \cellcolor{famtree}1.32$\times$ & \cellcolor{famtree}0.95$\times$ & \cellcolor{famtree}6.04 & \cellcolor{famtree}2.56$\times$ & \cellcolor{famtree}2.27$\times$ & \cellcolor{famtree}1.69$\times$ & \cellcolor{famtree}1.39$\times$ & \cellcolor{famtree}1.00$\times$ & \cellcolor{famtree}0.73$\times$ & \cellcolor{famtree}4.61 \\
 &  & \cellcolor{famtree}Tree $B{=}31$ & \cellcolor{famtree}3.42$\times$ & \cellcolor{famtree}2.56$\times$ & \cellcolor{famtree}1.97$\times$ & \cellcolor{famtree}1.30$\times$ & \cellcolor{famtree}0.84$\times$ & \cellcolor{famtree}0.59$\times$ & \cellcolor{famtree}6.57 & \cellcolor{famtree}2.52$\times$ & \cellcolor{famtree}1.81$\times$ & \cellcolor{famtree}1.45$\times$ & \cellcolor{famtree}1.00$\times$ & \cellcolor{famtree}0.67$\times$ & \cellcolor{famtree}0.47$\times$ & \cellcolor{famtree}4.98 \\
 &  & \cellcolor{famtree}Tree $B{=}63$ & \cellcolor{famtree}2.97$\times$ & \cellcolor{famtree}1.54$\times$ & \cellcolor{famtree}0.94$\times$ & \cellcolor{famtree}0.55$\times$ & \cellcolor{famtree}0.34$\times$ & \cellcolor{famtree}0.23$\times$ & \cellcolor{famtree}\textbf{6.95} & \cellcolor{famtree}2.19$\times$ & \cellcolor{famtree}1.15$\times$ & \cellcolor{famtree}0.73$\times$ & \cellcolor{famtree}0.44$\times$ & \cellcolor{famtree}0.28$\times$ & \cellcolor{famtree}0.20$\times$ & \cellcolor{famtree}\textbf{5.26} \\
 &  & \cellcolor{famuni}\textbf{Uni-LaSpec} & \cellcolor{famuni}\textbf{3.74}$\times$ & \cellcolor{famuni}\textbf{3.26}$\times$ & \cellcolor{famuni}\textbf{2.88}$\times$ & \cellcolor{famuni}\textbf{2.51}$\times$ & \cellcolor{famuni}2.12$\times$ & \cellcolor{famuni}\textbf{1.71}$\times$ & \cellcolor{famuni}5.15 & \cellcolor{famuni}\textbf{2.73}$\times$ & \cellcolor{famuni}\textbf{2.39}$\times$ & \cellcolor{famuni}\textbf{2.15}$\times$ & \cellcolor{famuni}\textbf{1.90}$\times$ & \cellcolor{famuni}\textbf{1.58}$\times$ & \cellcolor{famuni}\textbf{1.29}$\times$ & \cellcolor{famuni}3.91 \\
\cmidrule(l){2-17}
 & \multirow{8}{*}{IVL3.5-4B} & \cellcolor{famlin}Linear & \cellcolor{famlin}3.21$\times$ & \cellcolor{famlin}2.66$\times$ & \cellcolor{famlin}2.31$\times$ & \cellcolor{famlin}1.99$\times$ & \cellcolor{famlin}1.71$\times$ & \cellcolor{famlin}1.30$\times$ & \cellcolor{famlin}4.68 & \cellcolor{famlin}\textbf{2.25}$\times$ & \cellcolor{famlin}1.84$\times$ & \cellcolor{famlin}1.64$\times$ & \cellcolor{famlin}1.39$\times$ & \cellcolor{famlin}1.27$\times$ & \cellcolor{famlin}1.20$\times$ & \cellcolor{famlin}3.69 \\
 &  & \cellcolor{famtrunc}Trunc.\ $N{=}5$ & \cellcolor{famtrunc}2.65$\times$ & \cellcolor{famtrunc}2.41$\times$ & \cellcolor{famtrunc}2.25$\times$ & \cellcolor{famtrunc}2.07$\times$ & \cellcolor{famtrunc}1.74$\times$ & \cellcolor{famtrunc}1.56$\times$ & \cellcolor{famtrunc}3.69 & \cellcolor{famtrunc}1.95$\times$ & \cellcolor{famtrunc}1.73$\times$ & \cellcolor{famtrunc}1.65$\times$ & \cellcolor{famtrunc}1.47$\times$ & \cellcolor{famtrunc}1.35$\times$ & \cellcolor{famtrunc}1.28$\times$ & \cellcolor{famtrunc}3.13 \\
 &  & \cellcolor{famtrunc}Trunc.\ $N{=}7$ & \cellcolor{famtrunc}2.88$\times$ & \cellcolor{famtrunc}2.61$\times$ & \cellcolor{famtrunc}2.42$\times$ & \cellcolor{famtrunc}\textbf{2.24}$\times$ & \cellcolor{famtrunc}1.81$\times$ & \cellcolor{famtrunc}\textbf{1.58}$\times$ & \cellcolor{famtrunc}4.09 & \cellcolor{famtrunc}2.08$\times$ & \cellcolor{famtrunc}1.83$\times$ & \cellcolor{famtrunc}\textbf{1.73}$\times$ & \cellcolor{famtrunc}\textbf{1.52}$\times$ & \cellcolor{famtrunc}\textbf{1.37}$\times$ & \cellcolor{famtrunc}\textbf{1.32}$\times$ & \cellcolor{famtrunc}3.38 \\
 &  & \cellcolor{famtrunc}Dyn.\ trunc. & \cellcolor{famtrunc}3.02$\times$ & \cellcolor{famtrunc}2.73$\times$ & \cellcolor{famtrunc}2.36$\times$ & \cellcolor{famtrunc}2.14$\times$ & \cellcolor{famtrunc}1.78$\times$ & \cellcolor{famtrunc}1.39$\times$ & \cellcolor{famtrunc}4.33 & \cellcolor{famtrunc}1.99$\times$ & \cellcolor{famtrunc}1.72$\times$ & \cellcolor{famtrunc}1.65$\times$ & \cellcolor{famtrunc}1.40$\times$ & \cellcolor{famtrunc}1.29$\times$ & \cellcolor{famtrunc}1.27$\times$ & \cellcolor{famtrunc}3.27 \\
 &  & \cellcolor{famtree}Tree $B{=}15$ & \cellcolor{famtree}2.80$\times$ & \cellcolor{famtree}2.51$\times$ & \cellcolor{famtree}2.12$\times$ & \cellcolor{famtree}1.67$\times$ & \cellcolor{famtree}1.18$\times$ & \cellcolor{famtree}0.86$\times$ & \cellcolor{famtree}5.45 & \cellcolor{famtree}1.94$\times$ & \cellcolor{famtree}1.44$\times$ & \cellcolor{famtree}1.22$\times$ & \cellcolor{famtree}0.90$\times$ & \cellcolor{famtree}0.72$\times$ & \cellcolor{famtree}0.67$\times$ & \cellcolor{famtree}4.38 \\
 &  & \cellcolor{famtree}Tree $B{=}31$ & \cellcolor{famtree}3.00$\times$ & \cellcolor{famtree}2.50$\times$ & \cellcolor{famtree}1.94$\times$ & \cellcolor{famtree}1.58$\times$ & \cellcolor{famtree}0.97$\times$ & \cellcolor{famtree}0.69$\times$ & \cellcolor{famtree}5.91 & \cellcolor{famtree}2.05$\times$ & \cellcolor{famtree}1.42$\times$ & \cellcolor{famtree}1.15$\times$ & \cellcolor{famtree}0.87$\times$ & \cellcolor{famtree}0.60$\times$ & \cellcolor{famtree}0.57$\times$ & \cellcolor{famtree}4.68 \\
 &  & \cellcolor{famtree}Tree $B{=}63$ & \cellcolor{famtree}3.03$\times$ & \cellcolor{famtree}2.16$\times$ & \cellcolor{famtree}1.75$\times$ & \cellcolor{famtree}1.19$\times$ & \cellcolor{famtree}0.73$\times$ & \cellcolor{famtree}0.52$\times$ & \cellcolor{famtree}\textbf{6.22} & \cellcolor{famtree}2.06$\times$ & \cellcolor{famtree}1.26$\times$ & \cellcolor{famtree}1.05$\times$ & \cellcolor{famtree}0.69$\times$ & \cellcolor{famtree}0.48$\times$ & \cellcolor{famtree}0.44$\times$ & \cellcolor{famtree}\textbf{4.92} \\
 &  & \cellcolor{famuni}\textbf{Uni-LaSpec} & \cellcolor{famuni}\textbf{3.24}$\times$ & \cellcolor{famuni}\textbf{2.93}$\times$ & \cellcolor{famuni}\textbf{2.43}$\times$ & \cellcolor{famuni}2.23$\times$ & \cellcolor{famuni}\textbf{1.83}$\times$ & \cellcolor{famuni}1.42$\times$ & \cellcolor{famuni}4.71 & \cellcolor{famuni}2.23$\times$ & \cellcolor{famuni}\textbf{1.88}$\times$ & \cellcolor{famuni}1.69$\times$ & \cellcolor{famuni}1.46$\times$ & \cellcolor{famuni}1.33$\times$ & \cellcolor{famuni}1.21$\times$ & \cellcolor{famuni}3.60 \\
\cmidrule(l){2-17}
 & \multirow{8}{*}{IVL3.5-8B} & \cellcolor{famlin}Linear & \cellcolor{famlin}3.93$\times$ & \cellcolor{famlin}3.42$\times$ & \cellcolor{famlin}3.09$\times$ & \cellcolor{famlin}2.48$\times$ & \cellcolor{famlin}2.02$\times$ & \cellcolor{famlin}1.55$\times$ & \cellcolor{famlin}5.38 & \cellcolor{famlin}2.43$\times$ & \cellcolor{famlin}1.98$\times$ & \cellcolor{famlin}1.76$\times$ & \cellcolor{famlin}1.56$\times$ & \cellcolor{famlin}1.31$\times$ & \cellcolor{famlin}1.15$\times$ & \cellcolor{famlin}3.68 \\
 &  & \cellcolor{famtrunc}Trunc.\ $N{=}5$ & \cellcolor{famtrunc}2.98$\times$ & \cellcolor{famtrunc}2.83$\times$ & \cellcolor{famtrunc}2.65$\times$ & \cellcolor{famtrunc}2.40$\times$ & \cellcolor{famtrunc}2.06$\times$ & \cellcolor{famtrunc}1.72$\times$ & \cellcolor{famtrunc}4.02 & \cellcolor{famtrunc}2.12$\times$ & \cellcolor{famtrunc}1.89$\times$ & \cellcolor{famtrunc}1.72$\times$ & \cellcolor{famtrunc}1.61$\times$ & \cellcolor{famtrunc}1.45$\times$ & \cellcolor{famtrunc}\textbf{1.30}$\times$ & \cellcolor{famtrunc}3.14 \\
 &  & \cellcolor{famtrunc}Trunc.\ $N{=}7$ & \cellcolor{famtrunc}3.36$\times$ & \cellcolor{famtrunc}3.18$\times$ & \cellcolor{famtrunc}2.84$\times$ & \cellcolor{famtrunc}2.58$\times$ & \cellcolor{famtrunc}2.16$\times$ & \cellcolor{famtrunc}\textbf{1.79}$\times$ & \cellcolor{famtrunc}4.54 & \cellcolor{famtrunc}2.25$\times$ & \cellcolor{famtrunc}2.03$\times$ & \cellcolor{famtrunc}\textbf{1.80}$\times$ & \cellcolor{famtrunc}\textbf{1.69}$\times$ & \cellcolor{famtrunc}1.45$\times$ & \cellcolor{famtrunc}1.27$\times$ & \cellcolor{famtrunc}3.38 \\
 &  & \cellcolor{famtrunc}Dyn.\ trunc. & \cellcolor{famtrunc}3.72$\times$ & \cellcolor{famtrunc}3.36$\times$ & \cellcolor{famtrunc}3.08$\times$ & \cellcolor{famtrunc}2.66$\times$ & \cellcolor{famtrunc}2.22$\times$ & \cellcolor{famtrunc}1.72$\times$ & \cellcolor{famtrunc}5.13 & \cellcolor{famtrunc}2.15$\times$ & \cellcolor{famtrunc}1.90$\times$ & \cellcolor{famtrunc}1.74$\times$ & \cellcolor{famtrunc}1.57$\times$ & \cellcolor{famtrunc}1.38$\times$ & \cellcolor{famtrunc}1.22$\times$ & \cellcolor{famtrunc}3.34 \\
 &  & \cellcolor{famtree}Tree $B{=}15$ & \cellcolor{famtree}3.78$\times$ & \cellcolor{famtree}3.28$\times$ & \cellcolor{famtree}2.86$\times$ & \cellcolor{famtree}2.13$\times$ & \cellcolor{famtree}1.46$\times$ & \cellcolor{famtree}1.04$\times$ & \cellcolor{famtree}6.10 & \cellcolor{famtree}2.38$\times$ & \cellcolor{famtree}1.78$\times$ & \cellcolor{famtree}1.46$\times$ & \cellcolor{famtree}1.14$\times$ & \cellcolor{famtree}0.86$\times$ & \cellcolor{famtree}0.68$\times$ & \cellcolor{famtree}4.35 \\
 &  & \cellcolor{famtree}Tree $B{=}31$ & \cellcolor{famtree}4.04$\times$ & \cellcolor{famtree}3.26$\times$ & \cellcolor{famtree}2.65$\times$ & \cellcolor{famtree}1.88$\times$ & \cellcolor{famtree}1.13$\times$ & \cellcolor{famtree}0.83$\times$ & \cellcolor{famtree}6.62 & \cellcolor{famtree}2.47$\times$ & \cellcolor{famtree}1.75$\times$ & \cellcolor{famtree}1.34$\times$ & \cellcolor{famtree}1.03$\times$ & \cellcolor{famtree}0.69$\times$ & \cellcolor{famtree}0.58$\times$ & \cellcolor{famtree}4.66 \\
 &  & \cellcolor{famtree}Tree $B{=}63$ & \cellcolor{famtree}3.93$\times$ & \cellcolor{famtree}2.84$\times$ & \cellcolor{famtree}2.16$\times$ & \cellcolor{famtree}1.33$\times$ & \cellcolor{famtree}0.82$\times$ & \cellcolor{famtree}0.59$\times$ & \cellcolor{famtree}\textbf{6.94} & \cellcolor{famtree}2.40$\times$ & \cellcolor{famtree}1.57$\times$ & \cellcolor{famtree}1.16$\times$ & \cellcolor{famtree}0.77$\times$ & \cellcolor{famtree}0.54$\times$ & \cellcolor{famtree}0.44$\times$ & \cellcolor{famtree}\textbf{4.93} \\
 &  & \cellcolor{famuni}\textbf{Uni-LaSpec} & \cellcolor{famuni}\textbf{4.14}$\times$ & \cellcolor{famuni}\textbf{3.78}$\times$ & \cellcolor{famuni}\textbf{3.16}$\times$ & \cellcolor{famuni}\textbf{2.74}$\times$ & \cellcolor{famuni}\textbf{2.25}$\times$ & \cellcolor{famuni}1.76$\times$ & \cellcolor{famuni}5.41 & \cellcolor{famuni}\textbf{2.61}$\times$ & \cellcolor{famuni}\textbf{2.13}$\times$ & \cellcolor{famuni}\textbf{1.80}$\times$ & \cellcolor{famuni}1.66$\times$ & \cellcolor{famuni}\textbf{1.46}$\times$ & \cellcolor{famuni}1.28$\times$ & \cellcolor{famuni}3.62 \\
\bottomrule
\end{tabular}

%% file: references.bib
@IEEEtranBSTCTL{IEEEtran:BSTcontrol,
  CTLdash_repeated_names  = "no",
  CTLuse_forced_etal      = "yes",
  CTLmax_names_forced_etal = "6",
  CTLnames_show_etal      = "6"
}

@inproceedings{gagrani2024speculative,
  author    = {Mukul Gagrani and Raghavv Goel and Wonseok Jeon and Junyoung Park and Mingu Lee and Christopher Lott},
  title     = {On Speculative Decoding for Multimodal Large Language Models},
  booktitle = {Proceedings of the IEEE/CVF Conference on Computer Vision and Pattern Recognition Workshops},
  pages     = {8285--8289},
  year      = {2024}
}

@inproceedings{ganesan2025massv,
  author    = {Mugilan Ganesan and Shane Segal and Ankur Aggarwal and Nish Sinnadurai and Sean Lie and Vithursan Thangarasa},
  title     = {{MASSV}: Multimodal Adaptation and Self-Data Distillation for Speculative Decoding of Vision-Language Models},
  booktitle = {Findings of the Association for Computational Linguistics: EMNLP 2025},
  pages     = {12265--12276},
  year      = {2025},
  doi       = {10.18653/v1/2025.findings-emnlp.656}
}

@article{lin2025msd,
  author  = {Luxi Lin and Zhihang Lin and Zhanpeng Zeng and Rongrong Ji},
  title   = {Speculative Decoding Reimagined for Multimodal Large Language Models},
  journal = {arXiv preprint arXiv:2505.14260},
  year    = {2025},
  doi     = {10.48550/arXiv.2505.14260}
}

@inproceedings{hu2025dream,
  author    = {Yunhai Hu and Tianhua Xia and Zining Liu and Rahul Raman and Xingyu Liu and Bo Bao and Eric Sather and Vithursan Thangarasa and Sai Qian Zhang},
  title     = {{DREAM}: Drafting with Refined Target Features and Entropy-Adaptive Cross-Attention Fusion for Multimodal Speculative Decoding},
  booktitle = {Advances in Neural Information Processing Systems},
  volume    = {38},
  year      = {2025},
  doi       = {10.52202/085713-5586}
}

@article{huang2025specvlm,
  author  = {Haiduo Huang and Fuwei Yang and Zhenhua Liu and Xuanwu Yin and Dong Li and Pengju Ren and Emad Barsoum},
  title   = {{SpecVLM}: Fast Speculative Decoding in Vision-Language Models},
  journal = {arXiv preprint arXiv:2509.11815},
  year    = {2025},
  doi     = {10.48550/arXiv.2509.11815}
}

@inproceedings{wang2025specflash,
  author    = {Zihua Wang and Ruibo Li and Haozhe Du and Joey Tianyi Zhou and Yu Zhang and Xu Yang},
  title     = {{SpecFLASH}: A Latent-Guided Semi-Autoregressive Speculative Decoding Framework for Efficient Multimodal Generation},
  booktitle = {Proceedings of the 34th ACM International Conference on Multimedia},
  year      = {2026}
}

@inproceedings{xie2025hivis,
  author    = {Zhinan Xie and Peisong Wang and Shuang Qiu and Jian Cheng},
  title     = {{HiViS}: Hiding Visual Tokens from the Drafter for Speculative Decoding in Vision-Language Models},
  booktitle = {Proceedings of the IEEE/CVF Conference on Computer Vision and Pattern Recognition Findings},
  pages     = {8952--8961},
  year      = {2026}
}

@article{shen2026mmspec,
  author  = {Hui Shen and Xin Wang and Ping Zhang and Yunta Hsieh and Qi Han and Zhongwei Wan and Ziheng Zhang and Jingxuan Zhang and Jing Xiong and Ziyuan Liu and Yifan Zhang and Hangrui Cao and Chenyang Zhao and Mi Zhang},
  title   = {{MMSpec}: Benchmarking Speculative Decoding for Vision-Language Models},
  journal = {arXiv preprint arXiv:2603.14989},
  year    = {2026},
  doi     = {10.48550/arXiv.2603.14989}
}

@inproceedings{sun2024blockverification,
  author    = {Ziteng Sun and Uri Mendlovic and Yaniv Leviathan and Asaf Aharoni and Jae Hun Ro and Ahmad Beirami and Ananda Theertha Suresh},
  title     = {Block Verification Accelerates Speculative Decoding},
  booktitle = {International Conference on Learning Representations},
  year      = {2025}
}

@inproceedings{weng2025traversal,
  author    = {Yepeng Weng and Qiao Hu and Xujie Chen and Li Liu and Dianwen Mei and Huishi Qiu and Jiang Tian and Zhongchao Shi},
  title     = {Traversal Verification for Speculative Tree Decoding},
  booktitle = {Advances in Neural Information Processing Systems},
  volume    = {38},
  year      = {2025},
  doi       = {10.52202/085713-2611}
}

@inproceedings{liu2025pearl,
  author    = {Tianyu Liu and Yun Li and Qitan Lv and Kai Liu and Jianchen Zhu and Winston Hu and Xiao Sun},
  title     = {{PEARL}: Parallel Speculative Decoding with Adaptive Draft Length},
  booktitle = {International Conference on Learning Representations},
  year      = {2025}
}

@inproceedings{shen2026specbranch,
  author    = {Yuhao Shen and Junyi Shen and Quan Kong and Tianyu Liu and Yao Lu and Cong Wang},
  title     = {{SpecBranch}: Speculative Decoding via Hybrid Drafting and Rollback-Aware Branch Parallelism},
  booktitle = {International Conference on Learning Representations},
  year      = {2026}
}

@article{zhang2026flexdraft,
  author  = {Yaojie Zhang and Jianuo Huang and Junlong Ke and Yuhang Han and Yongji Long and Tianchen Zhao and Biqing Qi and Linfeng Zhang},
  title   = {{FlexDraft}: Flexible Speculative Decoding via Attention Tuning and Bonus-Guided Calibration},
  journal = {arXiv preprint arXiv:2605.20022},
  year    = {2026},
  doi     = {10.48550/arXiv.2605.20022}
}

@inproceedings{kong2026parallelvlm,
  author    = {Quan Kong and Yuhao Shen and Yicheng Ji and Huan Li and Cong Wang},
  title     = {{ParallelVLM}: Lossless Video-{LLM} Acceleration with Visual Alignment Aware Parallel Speculative Decoding},
  booktitle = {Proceedings of the IEEE/CVF Conference on Computer Vision and Pattern Recognition},
  pages     = {11392--11402},
  year      = {2026}
}

@inproceedings{qian2024bass,
  author    = {Haifeng Qian and Sujan Kumar Gonugondla and Sungsoo Ha and Mingyue Shang and Sanjay Krishna Gouda and Ramesh Nallapati and Sudipta Sengupta and Xiaofei Ma and Anoop Deoras},
  title     = {{BASS}: Batched Attention-Optimized Speculative Sampling},
  booktitle = {Findings of the Association for Computational Linguistics: ACL 2024},
  pages     = {8214--8224},
  year      = {2024},
  doi       = {10.18653/v1/2024.findings-acl.489}
}

@inproceedings{sadhukhan2024magicdec,
  author    = {Ranajoy Sadhukhan and Jian Chen and Zhuoming Chen and Vashisth Tiwari and Ruihang Lai and Jinyuan Shi and Ian En-Hsu Yen and Avner May and Tianqi Chen and Beidi Chen},
  title     = {{MagicDec}: Breaking the Latency-Throughput Tradeoff for Long Context Generation with Speculative Decoding},
  booktitle = {International Conference on Learning Representations},
  year      = {2025}
}

@inproceedings{wu2025tetris,
  author    = {Zhaoxuan Wu and Zijian Zhou and Arun Verma and Alok Prakash and Daniela Rus and Bryan Kian Hsiang Low},
  title     = {{TETRIS}: Optimal Draft Token Selection for Batch Speculative Decoding},
  booktitle = {Proceedings of the 63rd Annual Meeting of the Association for Computational Linguistics},
  pages     = {33329--33345},
  year      = {2025},
  doi       = {10.18653/v1/2025.acl-long.1598}
}

@inproceedings{huang2025adaspec,
  author    = {Kaiyu Huang and Hao Wu and Zhubo Shi and Han Zou and Minchen Yu and Qingjiang Shi},
  title     = {{AdaSpec}: Adaptive Speculative Decoding for Fast, {SLO}-Aware Large Language Model Serving},
  booktitle = {Proceedings of the ACM Symposium on Cloud Computing},
  year      = {2025},
  doi       = {10.1145/3772052.3772239}
}

@article{zhang2025batchspec,
  author  = {Ranran Haoran Zhang and Soumik Dey and Ashirbad Mishra and Hansi Wu and Binbin Li and Rui Zhang},
  title   = {Batch Speculative Decoding Done Right},
  journal = {arXiv preprint arXiv:2510.22876},
  year    = {2025},
  doi     = {10.48550/arXiv.2510.22876}
}

@article{xie2026spectre,
  author  = {Jincheng Xie and Yawen Ling and Qi Xiao and Feiyu Zhang and Zhongyi Huang and Wen Hu and Yu Zheng},
  title   = {{SPECTRE}: Hybrid Ordinary-Parallel Speculative Serving for Resource-Efficient {LLM} Inference},
  journal = {arXiv preprint arXiv:2605.08151},
  year    = {2026},
  doi     = {10.48550/arXiv.2605.08151}
}

@inproceedings{leviathan2023fast,
  author    = {Yaniv Leviathan and Matan Kalman and Yossi Matias},
  title     = {Fast Inference from Transformers via Speculative Decoding},
  booktitle = {Proceedings of the 40th International Conference on Machine Learning},
  series    = {Proceedings of Machine Learning Research},
  volume    = {202},
  pages     = {19274--19286},
  year      = {2023}
}

@article{chen2023accelerating,
  author  = {Charlie Chen and Sebastian Borgeaud and Geoffrey Irving and Jean-Baptiste Lespiau and Laurent Sifre and John Jumper},
  title   = {Accelerating Large Language Model Decoding with Speculative Sampling},
  journal = {arXiv preprint arXiv:2302.01318},
  year    = {2023},
  doi     = {10.48550/arXiv.2302.01318}
}

@inproceedings{li2024eagle,
  author    = {Yuhui Li and Fangyun Wei and Chao Zhang and Hongyang Zhang},
  title     = {{EAGLE}: Speculative Sampling Requires Rethinking Feature Uncertainty},
  booktitle = {Proceedings of the 41st International Conference on Machine Learning},
  series    = {Proceedings of Machine Learning Research},
  volume    = {235},
  pages     = {28935--28948},
  year      = {2024}
}

@inproceedings{li2025eagle3,
  author    = {Yuhui Li and Fangyun Wei and Chao Zhang and Hongyang Zhang},
  title     = {{EAGLE-3}: Scaling up Inference Acceleration of Large Language Models via Training-Time Test},
  booktitle = {Advances in Neural Information Processing Systems},
  volume    = {38},
  year      = {2025},
  doi       = {10.52202/085713-4562}
}

@inproceedings{cai2024medusa,
  author    = {Tianle Cai and Yuhong Li and Zhengyang Geng and Hongwu Peng and Jason D. Lee and Deming Chen and Tri Dao},
  title     = {{Medusa}: Simple {LLM} Inference Acceleration Framework with Multiple Decoding Heads},
  booktitle = {Proceedings of the 41st International Conference on Machine Learning},
  series    = {Proceedings of Machine Learning Research},
  volume    = {235},
  pages     = {5209--5235},
  year      = {2024}
}

@inproceedings{chen2026dflash,
  author    = {Jian Chen and Yesheng Liang and Zhijian Liu},
  title     = {{DFlash}: Block Diffusion for Flash Speculative Decoding},
  booktitle = {International Conference on Machine Learning},
  year      = {2026}
}

@article{zhang2026caddtree,
  author  = {Shuai Zhang and Huachuan Qiu and Hongliang He and Yong Dai},
  title   = {Cost-Aware Diffusion Draft Trees for Speculative Decoding},
  journal = {arXiv preprint arXiv:2606.01813},
  year    = {2026},
  doi     = {10.48550/arXiv.2606.01813}
}

@article{ringel2026ddtree,
  author  = {Liran Ringel and Yaniv Romano},
  title   = {Accelerating Speculative Decoding with Block Diffusion Draft Trees},
  journal = {arXiv preprint arXiv:2604.12989},
  year    = {2026},
  doi     = {10.48550/arXiv.2604.12989}
}

@article{li2026mmsdready,
  author  = {Yantao Li and Huanlin Gao and Fang Zhao and Chao Tan and Qiang Hui and Shuting Liu and Fuyuan Shi and Ting Lu and Shaoan Zhao and Xueqiang Guo and Xinpei Su and Jianbing Zhang and Xinyu Dai and Kai Wang and Shiguo Lian},
  title   = {Is Multimodal Speculative Decoding Ready for Diffusion-Based Parallel Drafting? {A} Survey and Empirical Diagnosis},
  journal = {arXiv preprint arXiv:2608.20743},
  year    = {2026},
  doi     = {10.48550/arXiv.2608.20743}
}

@inproceedings{kang2025vispec,
  author    = {Jialiang Kang and Han Shu and Wenshuo Li and Yingjie Zhai and Xinghao Chen},
  title     = {{ViSpec}: Accelerating Vision-Language Models with Vision-Aware Speculative Decoding},
  booktitle = {Advances in Neural Information Processing Systems},
  volume    = {38},
  year      = {2025},
  doi       = {10.52202/085713-3852}
}

@article{lee2026glance,
  author  = {Jungseob Lee and Seongtae Hong and Dongyub Jude Lee and Chanjun Park and Jaehyung Seo and Sugyeong Eo and Heuiseok Lim},
  title   = {Vision Is Not Overhead: One-Pass Block Drafting for Lossless Speculative Decoding in Vision-Language Models},
  journal = {arXiv preprint arXiv:2609.00355},
  year    = {2026},
  doi     = {10.48550/arXiv.2609.00355}
}

@inproceedings{stern2018blockwise,
  author    = {Mitchell Stern and Noam Shazeer and Jakob Uszkoreit},
  title     = {Blockwise Parallel Decoding for Deep Autoregressive Models},
  booktitle = {Advances in Neural Information Processing Systems},
  volume    = {31},
  pages     = {10107--10116},
  year      = {2018}
}

@inproceedings{dao2022flashattention,
  author    = {Tri Dao and Daniel Y. Fu and Stefano Ermon and Atri Rudra and Christopher Ré},
  title     = {{FlashAttention}: Fast and Memory-Efficient Exact Attention with IO-Awareness},
  booktitle = {Advances in Neural Information Processing Systems},
  volume    = {35},
  year      = {2022},
  doi       = {10.52202/068431-1189}
}

@inproceedings{zhang2023draftverify,
  author    = {Jun Zhang and Jue Wang and Huan Li and Lidan Shou and Ke Chen and Gang Chen and Sharad Mehrotra},
  title     = {Draft \& Verify: Lossless Large Language Model Acceleration via Self-Speculative Decoding},
  booktitle = {Proceedings of the 62nd Annual Meeting of the Association for Computational Linguistics},
  pages     = {11263--11282},
  year      = {2024},
  doi       = {10.18653/v1/2024.acl-long.607}
}

@inproceedings{kim2023bild,
  author    = {Sehoon Kim and Karttikeya Mangalam and Suhong Moon and Jitendra Malik and Michael W. Mahoney and Amir Gholami and Kurt Keutzer},
  title     = {Speculative Decoding with Big Little Decoder},
  booktitle = {Advances in Neural Information Processing Systems},
  volume    = {36},
  year      = {2023},
  doi       = {10.52202/075280-1705}
}

@inproceedings{dao2023flashattention2,
  author    = {Tri Dao},
  title     = {{FlashAttention-2}: Faster Attention with Better Parallelism and Work Partitioning},
  booktitle = {International Conference on Learning Representations},
  year      = {2024}
}

@inproceedings{kwon2023pagedattention,
  author    = {Woosuk Kwon and Zhuohan Li and Siyuan Zhuang and Ying Sheng and Lianmin Zheng and Cody Hao Yu and Joseph E. Gonzalez and Hao Zhang and Ion Stoica},
  title     = {Efficient Memory Management for Large Language Model Serving with {PagedAttention}},
  booktitle = {Proceedings of the 29th ACM Symposium on Operating Systems Principles},
  pages     = {611--626},
  year      = {2023},
  doi       = {10.1145/3600006.3613165}
}

@inproceedings{miao2023specinfer,
  author    = {Xupeng Miao and Gabriele Oliaro and Zhihao Zhang and Xinhao Cheng and Zeyu Wang and Zhengxin Zhang and Rae Ying Yee Wong and Alan Zhu and Lijie Yang and Xiaoxiang Shi and Chunan Shi and Zhuoming Chen and Daiyaan Arfeen and Reyna Abhyankar and Zhihao Jia},
  title     = {{SpecInfer}: Accelerating Generative Large Language Model Serving with Tree-based Speculative Inference and Verification},
  booktitle = {Proceedings of the 29th ACM International Conference on Architectural Support for Programming Languages and Operating Systems, Volume 3},
  pages     = {932--949},
  year      = {2024},
  doi       = {10.1145/3620666.3651335}
}

@inproceedings{he2023rest,
  author    = {Zhenyu He and Zexuan Zhong and Tianle Cai and Jason D. Lee and Di He},
  title     = {{REST}: Retrieval-Based Speculative Decoding},
  booktitle = {Proceedings of the 2024 Conference of the North American Chapter of the Association for Computational Linguistics: Human Language Technologies},
  pages     = {1582--1595},
  year      = {2024},
  doi       = {10.18653/v1/2024.naacl-long.88}
}

@inproceedings{liu2024kangaroo,
  author    = {Fangcheng Liu and Yehui Tang and Zhenhua Liu and Yunsheng Ni and Duyu Tang and Kai Han and Yunhe Wang},
  title     = {{Kangaroo}: Lossless Self-Speculative Decoding for Accelerating {LLMs} via Double Early Exiting},
  booktitle = {Advances in Neural Information Processing Systems},
  volume    = {37},
  year      = {2024},
  doi       = {10.52202/079017-0381}
}

@inproceedings{sun2024triforce,
  author    = {Hanshi Sun and Zhuoming Chen and Xinyu Yang and Yuandong Tian and Beidi Chen},
  title     = {{TriForce}: Lossless Acceleration of Long Sequence Generation with Hierarchical Speculative Decoding},
  booktitle = {Conference on Language Modeling},
  year      = {2024}
}

@inproceedings{xia2024survey,
  author    = {Heming Xia and Zhe Yang and Qingxiu Dong and Peiyi Wang and Yongqi Li and Tao Ge and Tianyu Liu and Wenjie Li and Zhifang Sui},
  title     = {Unlocking Efficiency in Large Language Model Inference: A Comprehensive Survey of Speculative Decoding},
  booktitle = {Findings of the Association for Computational Linguistics: ACL 2024},
  pages     = {7655--7671},
  year      = {2024},
  doi       = {10.18653/v1/2024.findings-acl.456}
}

@inproceedings{svirschevski2024specexec,
  author    = {Ruslan Svirschevski and Avner May and Zhuoming Chen and Beidi Chen and Zhihao Jia and Max Ryabinin},
  title     = {{SpecExec}: Massively Parallel Speculative Decoding for Interactive {LLM} Inference on Consumer Devices},
  booktitle = {Advances in Neural Information Processing Systems},
  volume    = {37},
  year      = {2024},
  doi       = {10.52202/079017-0522}
}

@inproceedings{fu2024lookahead,
  author    = {Yichao Fu and Peter Bailis and Ion Stoica and Hao Zhang},
  title     = {Break the Sequential Dependency of {LLM} Inference Using Lookahead Decoding},
  booktitle = {Proceedings of the 41st International Conference on Machine Learning},
  series    = {Proceedings of Machine Learning Research},
  volume    = {235},
  pages     = {14060--14079},
  year      = {2024}
}

@inproceedings{li2024eagle2,
  author    = {Yuhui Li and Fangyun Wei and Chao Zhang and Hongyang Zhang},
  title     = {{EAGLE-2}: Faster Inference of Language Models with Dynamic Draft Trees},
  booktitle = {Proceedings of the 2024 Conference on Empirical Methods in Natural Language Processing},
  pages     = {7421--7432},
  year      = {2024},
  doi       = {10.18653/v1/2024.emnlp-main.422}
}

@article{cheng2024redrafter,
  author  = {Yunfei Cheng and Aonan Zhang and Xuanyu Zhang and Chong Wang and Yi Wang},
  title   = {Recurrent Drafter for Fast Speculative Decoding in Large Language Models},
  journal = {arXiv preprint arXiv:2403.09919},
  year    = {2024},
  doi     = {10.48550/arXiv.2403.09919}
}

@inproceedings{ankner2024hydra,
  author    = {Zachary Ankner and Rishab Parthasarathy and Aniruddha Nrusimha and Christopher Rinard and Jonathan Ragan-Kelley and William Brandon},
  title     = {{Hydra}: Sequentially-Dependent Draft Heads for Medusa Decoding},
  booktitle = {Conference on Language Modeling},
  year      = {2024}
}

@article{chen2024sequoia,
  author  = {Zhuoming Chen and Avner May and Ruslan Svirschevski and Yuhsun Huang and Max Ryabinin and Zhihao Jia and Beidi Chen},
  title   = {{Sequoia}: Scalable, Robust, and Hardware-aware Speculative Decoding},
  journal = {arXiv preprint arXiv:2402.12374},
  year    = {2024},
  doi     = {10.48550/arXiv.2402.12374}
}

@inproceedings{liu2024kivi,
  author    = {Zirui Liu and Jiayi Yuan and Hongye Jin and Shaochen Zhong and Zhaozhuo Xu and Vladimir Braverman and Beidi Chen and Xia Hu},
  title     = {{KIVI}: A Tuning-Free Asymmetric 2bit Quantization for {KV} Cache},
  booktitle = {Proceedings of the 41st International Conference on Machine Learning},
  series    = {Proceedings of Machine Learning Research},
  volume    = {235},
  pages     = {32332--32344},
  year      = {2024}
}

@article{chen2015cococaptions,
  author  = {Xinlei Chen and Hao Fang and Tsung-Yi Lin and Ramakrishna Vedantam and Saurabh Gupta and Piotr Dollar and C. Lawrence Zitnick},
  title   = {Microsoft {COCO} Captions: Data Collection and Evaluation Server},
  journal = {arXiv preprint arXiv:1504.00325},
  year    = {2015},
  doi     = {10.48550/arXiv.1504.00325}
}

@inproceedings{singh2019textvqa,
  author    = {Amanpreet Singh and Vivek Natarajan and Meet Shah and Yu Jiang and Xinlei Chen and Dhruv Batra and Devi Parikh and Marcus Rohrbach},
  title     = {Towards {VQA} Models That Can Read},
  booktitle = {Proceedings of the IEEE/CVF Conference on Computer Vision and Pattern Recognition},
  pages     = {8317--8326},
  year      = {2019}
}

@inproceedings{hudson2019gqa,
  author    = {Drew A. Hudson and Christopher D. Manning},
  title     = {{GQA}: A New Dataset for Real-World Visual Reasoning and Compositional Question Answering},
  booktitle = {Proceedings of the IEEE/CVF Conference on Computer Vision and Pattern Recognition},
  pages     = {6700--6709},
  year      = {2019},
  doi       = {10.1109/CVPR.2019.00686}
}

@inproceedings{mathew2020docvqa,
  author    = {Minesh Mathew and Dimosthenis Karatzas and C. V. Jawahar},
  title     = {{DocVQA}: A Dataset for {VQA} on Document Images},
  booktitle = {Proceedings of the IEEE/CVF Winter Conference on Applications of Computer Vision},
  pages     = {2200--2209},
  year      = {2021},
  doi       = {10.1109/WACV48630.2021.00225}
}

@article{cobbe2021gsm8k,
  author  = {Karl Cobbe and Vineet Kosaraju and Mohammad Bavarian and Mark Chen and Heewoo Jun and Lukasz Kaiser and Matthias Plappert and Jerry Tworek and Jacob Hilton and Reiichiro Nakano and Christopher Hesse and John Schulman},
  title   = {Training Verifiers to Solve Math Word Problems},
  journal = {arXiv preprint arXiv:2110.14168},
  year    = {2021},
  doi     = {10.48550/arXiv.2110.14168}
}

@inproceedings{hendrycks2021math,
  author    = {Dan Hendrycks and Collin Burns and Saurav Kadavath and Akul Arora and Steven Basart and Eric Tang and Dawn Song and Jacob Steinhardt},
  title     = {Measuring Mathematical Problem Solving With the {MATH} Dataset},
  booktitle = {Advances in Neural Information Processing Systems Datasets and Benchmarks Track},
  volume    = {34},
  year      = {2021}
}

@article{chen2021humaneval,
  author  = {Mark Chen and Jerry Tworek and Heewoo Jun and Qiming Yuan and Henrique Ponde de Oliveira Pinto and Jared Kaplan and Harri Edwards and Yuri Burda and Nicholas Joseph and Greg Brockman and Alex Ray and Raul Puri and Gretchen Krueger and Michael Petrov and Heidy Khlaaf and Girish Sastry and Pamela Mishkin and Brooke Chan and Scott Gray and Nick Ryder and Mikhail Pavlov and Alethea Power and Lukasz Kaiser and Mohammad Bavarian and Clemens Winter and Philippe Tillet and Felipe Petroski Such and Dave Cummings and Matthias Plappert and Fotios Chantzis and Elizabeth Barnes and Ariel Herbert-Voss and William Hebgen Guss and Alex Nichol and Alex Paino and Nikolas Tezak and Jie Tang and Igor Babuschkin and Suchir Balaji and Shantanu Jain and William Saunders and Christopher Hesse and Andrew N. Carr and Jan Leike and Josh Achiam and Vedant Misra and Evan Morikawa and Alec Radford and Matthew Knight and Miles Brundage and Mira Murati and Katie Mayer and Peter Welinder and Bob McGrew and Dario Amodei and Sam McCandlish and Ilya Sutskever and Wojciech Zaremba},
  title   = {Evaluating Large Language Models Trained on Code},
  journal = {arXiv preprint arXiv:2107.03374},
  year    = {2021},
  doi     = {10.48550/arXiv.2107.03374}
}

@article{austin2021mbpp,
  author  = {Jacob Austin and Augustus Odena and Maxwell Nye and Maarten Bosma and Henryk Michalewski and David Dohan and Ellen Jiang and Carrie Cai and Michael Terry and Quoc Le and Charles Sutton},
  title   = {Program Synthesis with Large Language Models},
  journal = {arXiv preprint arXiv:2108.07732},
  year    = {2021},
  doi     = {10.48550/arXiv.2108.07732}
}

@inproceedings{lightman2023math500,
  author    = {Hunter Lightman and Vineet Kosaraju and Yura Burda and Harri Edwards and Bowen Baker and Teddy Lee and Jan Leike and John Schulman and Ilya Sutskever and Karl Cobbe},
  title     = {Let's Verify Step by Step},
  booktitle = {International Conference on Learning Representations},
  year      = {2024}
}

@inproceedings{zheng2023mtbench,
  author    = {Lianmin Zheng and Wei-Lin Chiang and Ying Sheng and Siyuan Zhuang and Zhanghao Wu and Yonghao Zhuang and Zi Lin and Zhuohan Li and Dacheng Li and Eric P. Xing and Hao Zhang and Joseph E. Gonzalez and Ion Stoica},
  title     = {Judging {LLM-as-a-Judge} with {MT-Bench} and Chatbot Arena},
  booktitle = {Advances in Neural Information Processing Systems Datasets and Benchmarks Track},
  volume    = {36},
  year      = {2023},
  doi       = {10.52202/075280-2020}
}

@inproceedings{ouyang2024omnidocbench,
  author    = {Linke Ouyang and Yuan Qu and Hongbin Zhou and Jiawei Zhu and Rui Zhang and Qunshu Lin and Bin Wang and Zhiyuan Zhao and Man Jiang and Xiaomeng Zhao and Jin Shi and Fan Wu and Pei Chu and Minghao Liu and Zhenxiang Li and Chao Xu and Bo Zhang and Botian Shi and Zhongying Tu and Conghui He},
  title     = {{OmniDocBench}: Benchmarking Diverse PDF Document Parsing with Comprehensive Annotations},
  booktitle = {Proceedings of the IEEE/CVF Conference on Computer Vision and Pattern Recognition},
  pages     = {24838--24848},
  year      = {2025}
}

@inproceedings{fu2024ocrbenchv2,
  author    = {Ling Fu and Zhebin Kuang and Jiajun Song and Mingxin Huang and Biao Yang and Yuzhe Li and Linghao Zhu and Qidi Luo and Xinyu Wang and Hao Lu and Zhang Li and Guozhi Tang and Bin Shan and Chunhui Lin and Qi Liu and Binghong Wu and Hao Feng and Hao Liu and Can Huang and Jingqun Tang and Wei Chen and Lianwen Jin and Yuliang Liu and Xiang Bai},
  title     = {{OCRBench} v2: An Improved Benchmark for Evaluating Large Multimodal Models on Visual Text Localization and Reasoning},
  booktitle = {Advances in Neural Information Processing Systems Datasets and Benchmarks Track},
  volume    = {38},
  year      = {2025}
}

@inproceedings{wang2024charxiv,
  author    = {Zirui Wang and Mengzhou Xia and Luxi He and Howard Chen and Yitao Liu and Richard Zhu and Kaiqu Liang and Xindi Wu and Haotian Liu and Sadhika Malladi and Alexis Chevalier and Sanjeev Arora and Danqi Chen},
  title     = {{CharXiv}: Charting Gaps in Realistic Chart Understanding in Multimodal {LLMs}},
  booktitle = {Advances in Neural Information Processing Systems Datasets and Benchmarks Track},
  volume    = {37},
  year      = {2024},
  doi       = {10.52202/079017-3609}
}

@inproceedings{yue2024mmmupro,
  author    = {Xiang Yue and Tianyu Zheng and Yuansheng Ni and Yubo Wang and Kai Zhang and Shengbang Tong and Yuxuan Sun and Botao Yu and Ge Zhang and Huan Sun and Yu Su and Wenhu Chen and Graham Neubig},
  title     = {{MMMU-Pro}: A More Robust Multi-discipline Multimodal Understanding Benchmark},
  booktitle = {Proceedings of the 63rd Annual Meeting of the Association for Computational Linguistics},
  pages     = {15134--15186},
  year      = {2025},
  doi       = {10.18653/v1/2025.acl-long.736}
}

@article{agrawal2024pixtral,
  author  = {Pravesh Agrawal and Szymon Antoniak and Emma Bou Hanna and Baptiste Bout and Devendra Chaplot and Jessica Chudnovsky and Diogo Costa and Baudouin De Monicault and Saurabh Garg and Theophile Gervet and Soham Ghosh and Amélie Héliou and Paul Jacob and Albert Q. Jiang and Kartik Khandelwal and Timothée Lacroix and Guillaume Lample and Diego Las Casas and Thibaut Lavril and Teven Le Scao and Andy Lo and William Marshall and Louis Martin and Arthur Mensch and Pavankumar Muddireddy and Valera Nemychnikova and Marie Pellat and Patrick Von Platen and Nikhil Raghuraman and Baptiste Rozière and Alexandre Sablayrolles and Lucile Saulnier and Romain Sauvestre and Wendy Shang and Roman Soletskyi and Lawrence Stewart and Pierre Stock and Joachim Studnia and Sandeep Subramanian and Sagar Vaze and Thomas Wang and Sophia Yang},
  title   = {Pixtral 12B},
  journal = {arXiv preprint arXiv:2410.07073},
  year    = {2024},
  doi     = {10.48550/arXiv.2410.07073}
}

@inproceedings{jain2024livecodebench,
  author    = {Naman Jain and King Han and Alex Gu and Wen-Ding Li and Fanjia Yan and Tianjun Zhang and Sida Wang and Armando Solar-Lezama and Koushik Sen and Ion Stoica},
  title     = {{LiveCodeBench}: Holistic and Contamination Free Evaluation of Large Language Models for Code},
  booktitle = {International Conference on Learning Representations},
  year      = {2025}
}

@article{poznanski2025olmocr,
  author  = {Jake Poznanski and Aman Rangapur and Jon Borchardt and Jason Dunkelberger and Regan Huff and Daniel Lin and Aman Rangapur and Christopher Wilhelm and Kyle Lo and Luca Soldaini},
  title   = {{olmOCR}: Unlocking Trillions of Tokens in {PDFs} with Vision Language Models},
  journal = {arXiv preprint arXiv:2502.18443},
  year    = {2025},
  doi     = {10.48550/arXiv.2502.18443}
}

@article{patterson2004latency,
  author  = {David A. Patterson},
  title   = {Latency Lags Bandwith},
  journal = {Communications of the {ACM}},
  volume  = {47},
  number  = {10},
  pages   = {71--75},
  year    = {2004},
  doi     = {10.1145/1022594.1022596}
}

@misc{gante2023assisted,
  author       = {Joao Gante},
  title        = {Assisted Generation: A New Direction toward Low-Latency Text Generation},
  howpublished = {Hugging Face Blog, \url{https://huggingface.co/blog/assisted-generation}},
  year         = {2023}
}

@inproceedings{xia2022specdec,
  author    = {Heming Xia and Tao Ge and Peiyi Wang and Si-Qing Chen and Furu Wei and Zhifang Sui},
  title     = {Speculative Decoding: Exploiting Speculative Execution for Accelerating Seq2seq Generation},
  booktitle = {Findings of the Association for Computational Linguistics: EMNLP 2023},
  pages     = {3909--3925},
  year      = {2023},
  doi       = {10.18653/v1/2023.findings-emnlp.257}
}

@inproceedings{zhou2023distillspec,
  author    = {Yongchao Zhou and Kaifeng Lyu and Ankit Singh Rawat and Aditya Krishna Menon and Afshin Rostamizadeh and Sanjiv Kumar and Jean-François Kagy and Rishabh Agarwal},
  title     = {{DistillSpec}: Improving Speculative Decoding via Knowledge Distillation},
  booktitle = {International Conference on Learning Representations},
  year      = {2024}
}

@article{yang2023ppd,
  author  = {Seongjun Yang and Gibbeum Lee and Jaewoong Cho and Dimitris Papailiopoulos and Kangwook Lee},
  title   = {Predictive Pipelined Decoding: A Compute-Latency Trade-off for Exact {LLM} Decoding},
  journal = {Transactions on Machine Learning Research},
  year    = {2024}
}

@inproceedings{santilli2023parallel,
  author    = {Andrea Santilli and Silvio Severino and Emilian Postolache and Valentino Maiorca and Michele Mancusi and Riccardo Marin and Emanuele Rodolà},
  title     = {Accelerating Transformer Inference for Translation via Parallel Decoding},
  booktitle = {Proceedings of the 61st Annual Meeting of the Association for Computational Linguistics},
  pages     = {12336--12355},
  year      = {2023},
  doi       = {10.18653/v1/2023.acl-long.689}
}

@inproceedings{ainslie2023gqa,
  author    = {Joshua Ainslie and James Lee-Thorp and Michiel de Jong and Yury Zemlyanskiy and Federico Lebr{\'o}n and Sumit Sanghai},
  title     = {{GQA}: Training Generalized Multi-Query Transformer Models from Multi-Head Checkpoints},
  booktitle = {Proceedings of the 2023 Conference on Empirical Methods in Natural Language Processing},
  pages     = {4895--4901},
  year      = {2023},
  doi       = {10.18653/v1/2023.emnlp-main.298}
}

@article{shazeer2019mqa,
  author  = {Noam Shazeer},
  title   = {Fast Transformer Decoding: One Write-Head is All You Need},
  journal = {arXiv preprint arXiv:1911.02150},
  year    = {2019},
  doi     = {10.48550/arXiv.1911.02150}
}

@inproceedings{gu2017nat,
  author    = {Jiatao Gu and James Bradbury and Caiming Xiong and Victor O. K. Li and Richard Socher},
  title     = {Non-Autoregressive Neural Machine Translation},
  booktitle = {International Conference on Learning Representations},
  year      = {2018}
}

@inproceedings{qian2020glat,
  author    = {Lihua Qian and Hao Zhou and Yu Bao and Mingxuan Wang and Lin Qiu and Weinan Zhang and Yong Yu and Lei Li},
  title     = {Glancing Transformer for Non-Autoregressive Neural Machine Translation},
  booktitle = {Proceedings of the 59th Annual Meeting of the Association for Computational Linguistics},
  pages     = {1993--2003},
  year      = {2021},
  doi       = {10.18653/v1/2021.acl-long.155}
}

@article{an2025llavaonevision15,
  author  = {Xiang An and Yin Xie and Kaicheng Yang and Wenkang Zhang and Xiuwei Zhao and Zheng Cheng and Yirui Wang and Songcen Xu and Changrui Chen and Chunsheng Wu and Huajie Tan and Chunyuan Li and Jing Yang and Jie Yu and Xiyao Wang and Bin Qin and Yumeng Wang and Zizhen Yan and Ziyong Feng and Ziwei Liu and Bo Li and Jiankang Deng},
  title   = {{LLaVA-OneVision-1.5}: Fully Open Framework for Democratized Multimodal Training},
  journal = {arXiv preprint arXiv:2509.23661},
  year    = {2025},
  doi     = {10.48550/arXiv.2509.23661}
}

@misc{nathawani2025nemotron,
  author       = {Dhruv Nathawani and Shuoyang Ding and Vitaly Lavrukhin and Igor Gitman and Somshubra Majumdar and Evelina Bakhturina and Boris Ginsburg and Jane Polak Scowcroft},
  title        = {{Nemotron} Post-Training Dataset v2},
  year         = {2025},
  howpublished = {\url{https://huggingface.co/datasets/nvidia/Nemotron-Post-Training-Dataset-v2}},
  note         = {NVIDIA}
}

@misc{chaudhary2023codealpaca,
  author       = {Sahil Chaudhary},
  title        = {{Code Alpaca}: An Instruction-following {LLaMA} Model for Code Generation},
  year         = {2023},
  howpublished = {\url{https://github.com/sahil280114/codealpaca}}
}
